\documentclass[10pt,twocolumn,letterpaper]{article}

\usepackage{cvpr}
\usepackage{times}
\usepackage{epsfig}
\usepackage{graphicx}
\usepackage{amsmath}
\usepackage{amssymb}
\usepackage{array}

\usepackage{subfigure}

\usepackage{url}            
\usepackage{booktabs}       
\usepackage{amsfonts}       
\usepackage{nicefrac}       
\usepackage{microtype}      
\usepackage{color}

\usepackage{algorithm}
\usepackage{algorithmic}
\usepackage{amssymb}
\usepackage{amsthm}
\usepackage{amsmath,epsfig}
\usepackage{epstopdf}

\usepackage{enumitem}

\newtheorem{theorem}{Theorem}

\newcommand{\TODO}[1]{\textcolor{blue}{}\textcolor{blue}{#1}}
\newcommand{\SM}{Appendix}

\usepackage[pagebackref=true,breaklinks=true,letterpaper=true,colorlinks,bookmarks=false]{hyperref}

\cvprfinalcopy 

\def\cvprPaperID{2854} 

\begin{document}

\title{Controllable Orthogonalization in Training DNNs}
\author{Lei Huang$^{1}$ \quad  Li Liu$^{1}$ \quad Fan Zhu$^{1}$ \quad Diwen Wan$^{1,2}$ \quad Zehuan Yuan$^{3}$ \quad Bo Li$^{4}$ \quad Ling Shao$^{1}$\\
	$^{1}$Inception Institute of Artificial Intelligence (IIAI), Abu Dhabi, UAE\\
	$^{2}$University of Electronic Science and Technology of China, Chengdu, China\\
	$^{3}$ByteDance AI Lab, Beijing, China\\
	$^{4}$University of Illinois at Urbana-Champaign
	Illinois, USA\\
}
\maketitle

\begin{abstract}
Orthogonality is widely used for training deep neural networks (DNNs) due to its ability to maintain all singular values of the Jacobian close to 1 and reduce redundancy in representation. 
This paper proposes a computationally efficient and numerically stable orthogonalization method using Newton's iteration (ONI), to learn a layer-wise orthogonal weight matrix in DNNs.
ONI works by iteratively stretching the singular values of a weight matrix towards 1.
This property enables it to control the orthogonality of a weight matrix by its number of iterations. 
We show that our method improves the performance of  image classification networks by effectively  controlling the orthogonality to provide an optimal tradeoff between optimization benefits and representational capacity reduction.
We also show that ONI stabilizes the training of generative adversarial networks (GANs)  by  maintaining the Lipschitz continuity of a network, similar to spectral normalization (SN), and further outperforms SN by providing controllable orthogonality. 	
\end{abstract}

\vspace{-0.05in}
\section{Introduction}
\label{sec_intro}
\vspace{-0.05in}

Training deep neural networks (DNNs) is often difficult due to the occurrence of vanishing/exploding gradients \cite{1994_TNN_Bengio,2010_AISTATS_Glorot,2013_ICML_Pascanu}. Preliminary research \cite{1998_NN_Yann,2010_AISTATS_Glorot} has suggested that weight initialization techniques are essential for avoiding these issues. As such, various works have tried to tackle the problem by designing weight matrices that can provide nearly equal variance to activations from different layers \cite{2010_AISTATS_Glorot,2015_ICCV_He}. Such a property can be further amplified by orthogonal weight initialization \cite{2013_CoRR_Saxe,2016_ICLR_Mishkin,2018_Arxiv_Piotr}, which shows excellent theoretical results in convergence due to its ability to obtain a DNN's \textit{dynamical isometry} \cite{2013_CoRR_Saxe,2017_NIPS_Pennington_non,2019_ICLR_Yang}, \ie all singular values of the input-output Jacobian are concentrated near 1. The improved performance of orthogonal initialization  is  empirically observed in \cite{2013_CoRR_Saxe,2016_ICLR_Mishkin,2017_NIPS_Pennington_non,2018_ICML_Xiao} and it  makes training even 10,000-layer DNNs possible \cite{2018_ICML_Xiao}. However, the initial orthogonality can be broken down and is not necessarily sustained throughout training \cite{2017_CVPR_Xie}. 


Previous works have tried to maintain the orthogonal weight matrix  by imposing an additional orthogonality penalty on the objective function, which can be viewed as a `soft orthogonal constraint' \cite{2013_ICML_Pascanu,2017_ICML_Eugene,2017_CVPR_Xie,2018_NIPS_WANG,2019_Arxiv_Amjad}. These methods show improved performance  in image classification \cite{2017_CVPR_Xie,2018_NIPS_Zhang,2018_CVPR_Lezama,2018_NIPS_WANG},  resisting  attacks from adversarial examples \cite{2017_ICML_Cisse}, neural photo editing \cite{2017_ICLR_Brock} and training generative adversarial networks (GAN) \cite{2019_ICLR_Brock, 2018_ICLR_Miyato}.
However, the introduced penalty  works like a pure regularization, and whether or not the orthogonality is truly maintained or  training benefited is unclear.
Other methods have been developed to directly solve the `hard orthogonal constraint' \cite{2017_ICML_Eugene,2018_NIPS_WANG}, either by  Riemannian optimization \cite{2016_Corr_Ozay,2017_Corr_Harandi} or by orthogonal weight normalization \cite{2016_NIPS_Wisdom,2018_AAAI_Huang}. 
However, Riemannian optimization often suffers from training instability \cite{2017_Corr_Harandi,2018_AAAI_Huang}, while  
 orthogonal weight normalization \cite{2018_AAAI_Huang} requires computationally expensive eigen decomposition, and the necessary back-propagation  through this eigen decomposition may suffer from numerical instability, as shown in \cite{2015_ICCV_Ionescu,2017_BMVC_Lin}.

\begin{figure}[t]
		\vspace{-0.1in}
	\centering
		\begin{minipage}[c]{.9\linewidth}
			\centering
			\includegraphics[width=7.4cm]{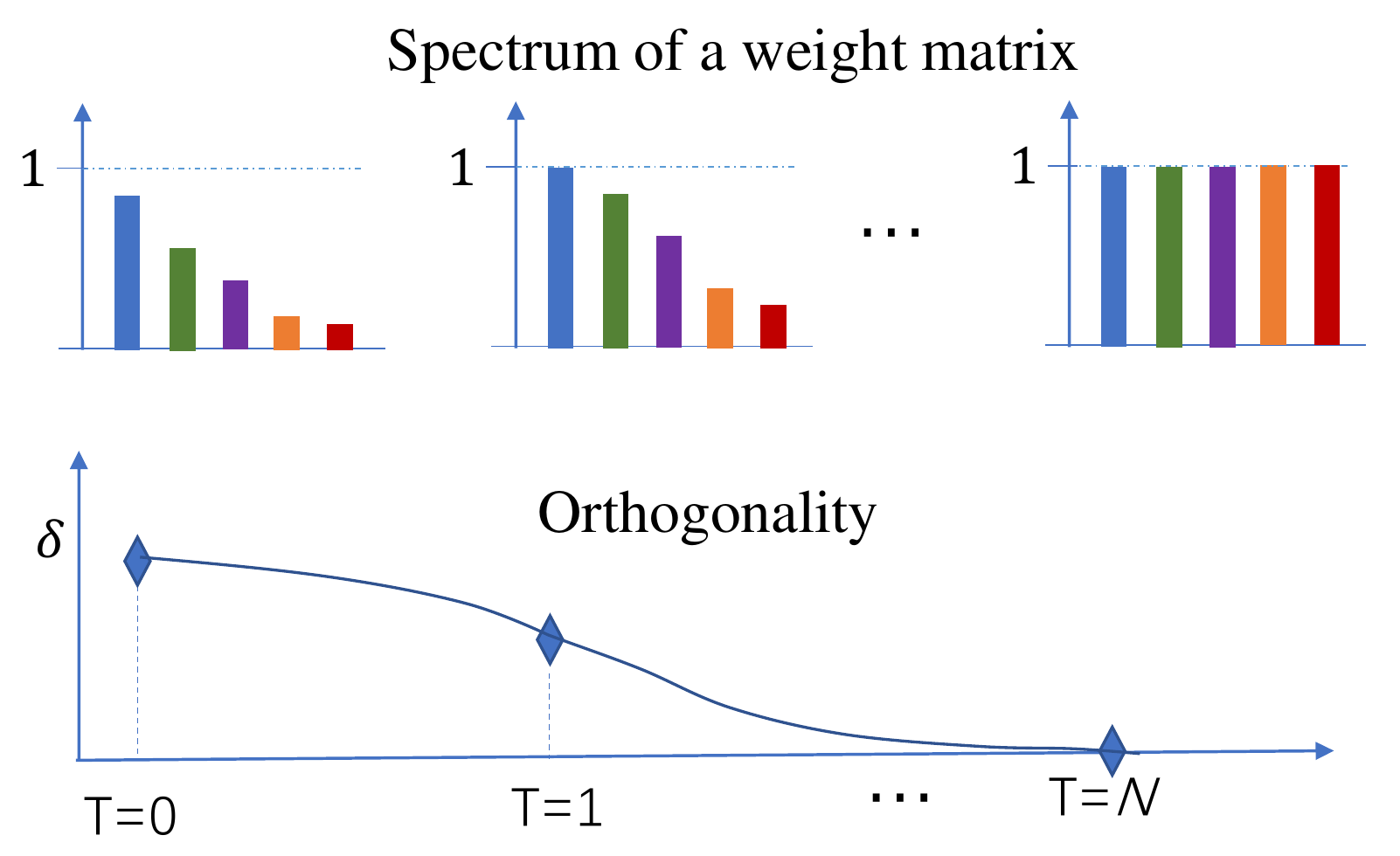}
		\end{minipage}
			\vspace{0.02in}
	\caption{ONI controls a weight matrix' magnitude of orthogonality (measured as $\delta= \|\mathbf{W} \mathbf{W}^T -\mathbf{I} \|_{F}$), by iteratively stretching its singular values towards 1.} 
	\label{fig:intro}
	\vspace{-0.22in}
\end{figure}
We  propose to perform orthogonalization by Newton's iteration (ONI) \cite{1950_ChemicalPhysics_Lwdin,2005_NumerialAlg,2019_CVPR_Huang} to  learn an exact orthogonal weight matrix, which is computationally efficient and numerically stable. To further speed up the convergence of Newton's iteration, we  propose two techniques: 1) we perform centering   to improve the conditioning of the proxy matrix; 2) we explore a more compact spectral bounding method to make  the initial singular value of the proxy matrix  closer to $1$.

We provide an insightful analysis and show that ONI works by iteratively stretching the singular values of the weight matrix towards $1$ (Figure \ref{fig:intro}). This property makes ONI  work well even if the weight matrix is singular (with multiple zero singular values), under which the eigen decomposition based method \cite{2018_AAAI_Huang} often suffers from numerical instability \cite{2015_ICCV_Ionescu,2017_BMVC_Lin}.  Moreover, we show that controlling orthogonality is necessary to balance the increase in optimization and reduction in representational capacity, and ONI can elegantly achieve this through its iteration number (Figure \ref{fig:intro}).  Besides,  ONI provides a unified solution for the row/column orthogonality, regardless of whether the weight matrix's output dimension  is smaller or larger than the input.

We also address practical strategies for effectively learning orthogonal weight matrices in DNNs. We introduce a constant of $\sqrt{2}$ to initially scale the orthonormal weight matrix  so that the  \textit{dynamical isometry} \cite{2013_CoRR_Saxe} can be well maintained for deep ReLU networks \cite{2010_ICML_Nair}. 
We conduct extensive experiments on multilayer perceptrons (MLPs) and convolutional neural networks (CNNs). Our proposed method benefits the training  and improves the test performance over multiple datasets, including ImageNet \cite{2015_ImageNet}. We also show that our method stabilizes the training of GANs and achieves improved performance on unsupervised image generation, compared to the widely used spectral normalization \cite{2018_ICLR_Miyato}.

\vspace{-0.08in}
\section{Related Work}
\vspace{-0.05in}
Orthogonal filters have been extensively explored in signal processing since they are capable of preserving  activation energy and reducing redundancy in  representation \cite{2006_TIP_Zhou}.
Saxe \etal \cite{2013_CoRR_Saxe} introduced an orthogonal weight  matrix for DNNs and showed that it achieves approximate \textit{dynamical isometry} \cite{2013_CoRR_Saxe} for deep linear neural networks, therefore significantly improving the optimization efficiency \cite{2016_ICLR_Mishkin,2018_Arxiv_Piotr}. Pennington \etal \cite{2017_NIPS_Pennington_non} further found that the nonlinear sigmoid network  can also obtain \textit{dynamical isometry} when combined with orthogonal weight initialization \cite{2017_NIPS_Pennington_non,2018_ICML_Xiao,2019_ICLR_Yang}.

Research has also been conducted into using \emph{orthogonal matrices}   to avoid the gradient vanishing/explosion problems in recurrent neural networks (RNNs). These methods mainly focus on constructing square orthogonal matrices/unitary matrices  for the hidden-to-hidden transformations in RNNs ~\cite{2016_ICML_Arjovsky,2016_NIPS_Wisdom,2016_CoRR_Dorobantu,2017_ICML_Eugene,2017_AAAI_Hyland,2017_GRU_Jing,2018_ICML_Kyle}. This is done by either constructing a decomposed unitary weight matrix with a restricted \cite{2016_ICML_Arjovsky} or  full representational capability \cite{2016_NIPS_Wisdom,2018_ICML_Kyle}, or by using soft constraints \cite{2017_ICML_Eugene}.
Different from  these methods requiring a square weight matrix and limited to hidden-to-hidden transformations in RNNs, our method is more general and can adapt to situations where the weight matrix is not square.

Our method is related to the methods that impose orthogonal penalties on the loss functions \cite{2013_ICML_Pascanu,2017_ICML_Eugene,2018_NIPS_WANG}. Most works propose to use soft orthogonality regularization under the standard Frobenius norm \cite{2013_ICML_Pascanu,2017_ICML_Eugene,2018_NIPS_WANG}, though other alternative orthogonal  penalties were explored in \cite{2018_NIPS_WANG}.  There are also methods that propose to bound the singular values with periodical projection \cite{2017_CVPR_Jia}. Our method targets at solving the `hard constraint' and providing controllable orthogonality.

One way to obtain exact orthogonality is through Riemannian optimization methods \cite{2016_Corr_Ozay,2017_Corr_Harandi}. These methods usually require a retract operation \cite{2012_SIAM_Absil} to project the updated weight back to the Stiefel manifold \cite{2016_Corr_Ozay,2017_Corr_Harandi}, which may result in training instability for DNNs \cite{2017_Corr_Harandi,2018_AAAI_Huang}.  
Our method avoids this by  employing re-parameterization to construct the orthogonal matrix \cite{2018_AAAI_Huang}.
Our work is closely related to  orthogonal weight normalization \cite{2018_AAAI_Huang}, which also uses re-parameterization to design an orthogonal transformation. However,  \cite{2018_AAAI_Huang} solves the problem by computationally expensive eigen decomposition and may result in numeric instability \cite{2015_ICCV_Ionescu,2017_BMVC_Lin}.
We use Newton's iteration \cite{1950_ChemicalPhysics_Lwdin, 2005_NumerialAlg}, which is more computationally efficient and numerically stable. 
 We further argue that fully orthogonalizing the weight matrix limits the network's learning capacity, which may result in degenerated performance \cite{2018_ICLR_Miyato,2019_ICLR_Brock}.
  Another related work is spectral normalization  \cite{2018_ICLR_Miyato}, which uses reparametrization to  bound only the maximum eigenvalue as 1. Our method can effectively interpolate between spectral normalization and full orthogonalization, by altering the iteration number.

Newton's iteration has also been employed in DNNs for constructing bilinear/second-order pooling \cite{2017_BMVC_Lin,2018_CVPR_Li}, or whitening the activations  \cite{2019_CVPR_Huang}. \cite{2017_BMVC_Lin} and \cite{2018_CVPR_Li} focused on calculating the square root of the covariance matrix, while our method computes the square root inverse of the covariance matrix, like the  work in \cite{2019_CVPR_Huang}. However, our work has several main differences from \cite{2019_CVPR_Huang}: 1) In \cite{2019_CVPR_Huang}, they aimed to whiten the activation \cite{2018_CVPR_Huang} over batch data using Newton's iteration, while our work seeks to learn the orthogonal weight matrix, which is an entirely different research problem \cite{2015_ICML_Ioffe,2016_NIPS_Salimans,2018_CVPR_Huang}; 2) We further improve the convergence speed compared to the Newton's iteration proposed in \cite{2019_CVPR_Huang} by providing more compact bounds; 
3) Our method can maintain the Lipschitz continuity of the network and thus has potential in stabilizing the training of GANs \cite{2018_ICLR_Miyato,2019_ICLR_Brock}. It is unclear whether or not the work in \cite{2019_CVPR_Huang} has such a property, since it is data-dependent normalization \cite{2015_ICML_Ioffe,2018_ICLR_Miyato,2019_ICLR_Brock}.  

\vspace{-0.06in}
\section{Proposed Method}
\vspace{-0.06in}
\label{sec:solve_optimization}
Given the dataset $D=\{(\mathbf{x}_i, \mathbf{y}_i)\}_{i=1}^{M}$ composed of an input $\mathbf{x}_i \in \mathbb{R}^d$ and its corresponding labels $\mathbf{y}_i \in \mathbb{R}^c$, we represent a standard feed-forward neural network as a function $f(\mathbf{x}; \theta)$ parameterized by $\theta$.  $f(\mathbf{x}; \theta)$ is a composition of $L$ simple nonlinear functions. Each of these consists of a linear transformation $\mathbf{\hat{h}}^l= \mathbf{W}^{l} \mathbf{h}^{l-1}+ \mathbf{b}^l$ with learnable weights  $\mathbf{W}^l \in \mathbb{R}^{n_l \times d_l}$ and biases $\mathbf{b}^l \in \mathbb{R}^{{n}_l}$, followed by an element-wise nonlinearity: $\mathbf{h}^l=\varphi(\mathbf{\hat{h}}^l)$. Here $l \in \{1,2,...,L\}$ indexes the layers.  
We denote the learnable parameters as $\theta=\{ {\mathbf{W}^l}, \mathbf{b}^l| l=1,2,\ldots,L \}$.
Training  neural networks involves minimizing the discrepancy between the desired output $\mathbf{y}$ and the predicted output $f(\mathbf{x}; \theta)$, described by a loss function $\mathcal{L}(\mathbf{y}, f(\mathbf{x}; \theta))$. Thus, the optimization objective is: $\theta^* =\arg \min_{\theta} \mathbb{E}_{(\mathbf{x},\mathbf{y})\in D} [\mathcal{L}(\mathbf{y}, f(\mathbf{x}; \theta))]$.

\begin{algorithm}[t]
	\algsetup{linenosize=\footnotesize}
	\small
	\caption{\small Orthogonalization by Newton's Iteration (ONI).}
	\label{alg_forward}
	\begin{algorithmic}[1]
		\STATE \textbf{Input}: proxy parameters $\mathbf{Z} \in \mathbb{R}^{n \times d} $ and iteration numbers $T$.
		\STATE  Bounding $\mathbf{Z}$'s singular values:  $\mathbf{V}= \frac{\mathbf{Z}}{\| \mathbf{Z} \|_F}$.
		\STATE  Calculate covariance matrix: $\mathbf{S}= \mathbf{V} \mathbf{V}^T$.
		\STATE $\mathbf{B}_0=\mathbf{I}$.
		\FOR {$t = 1$ to T}
		\STATE $\mathbf{B}_{t}=\frac{3}{2} \mathbf{B}_{t-1}-\frac{1}{2}\mathbf{B}_{t-1}^3  \mathbf{S} $.
		\ENDFOR		
		\STATE $\mathbf{W}=\mathbf{B}_{T} \mathbf{V}$
		\STATE \textbf{Output}:  orthogonalized weight matrix: $\mathbf{W} \in \mathbb{R}^{n \times d}$.
	\end{algorithmic}
\end{algorithm}

\begin{tiny}
	\begin{figure}[t]
		\vspace{-0.1in}
		\centering
		\subfigure[]{
			\begin{minipage}[c]{.46\linewidth}
				\centering
				\includegraphics[width=3.8cm]{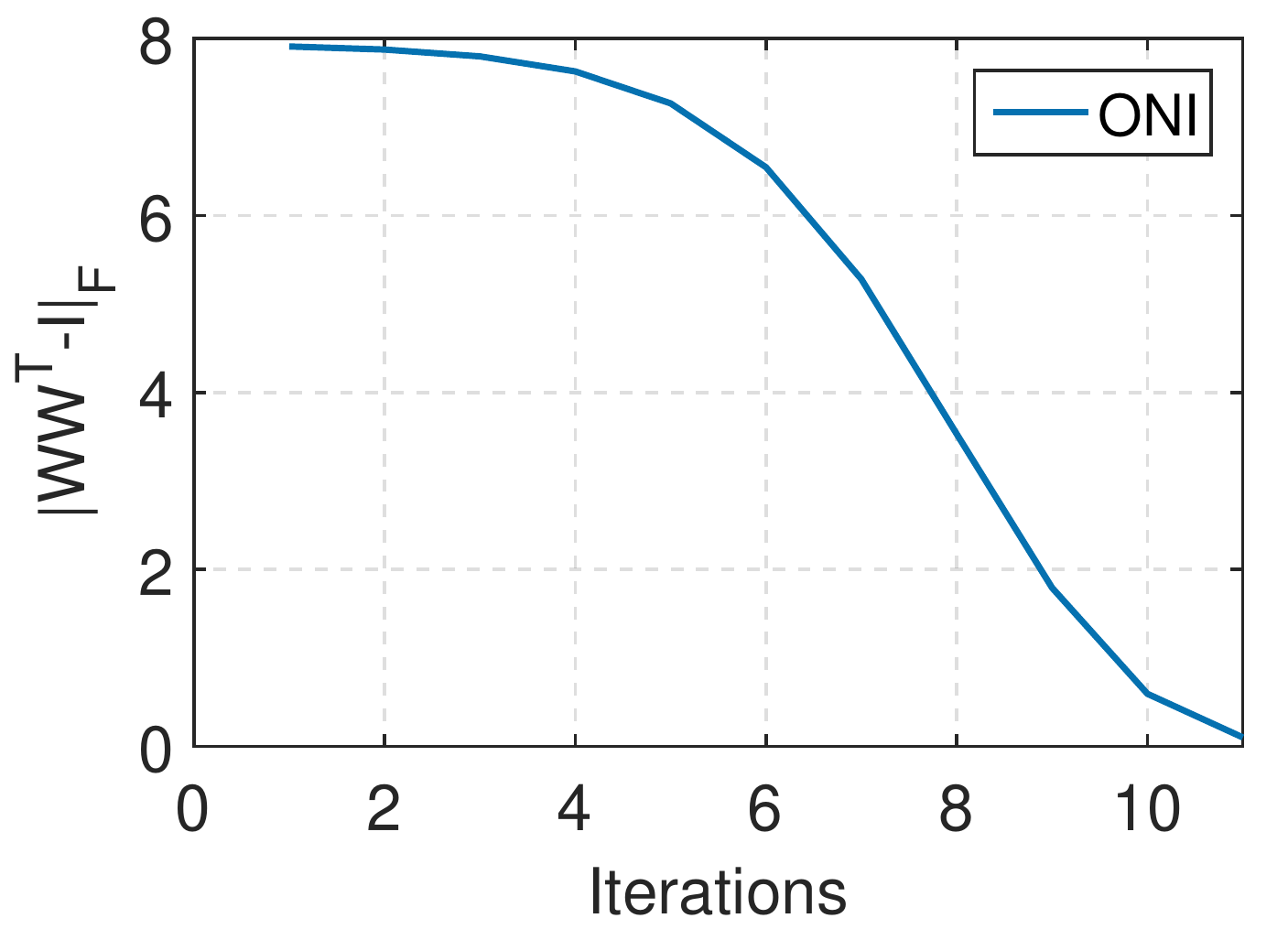}
			\end{minipage}
		}
		\subfigure[]{
			\begin{minipage}[c]{.46\linewidth}
				\centering
				\includegraphics[width=3.8cm]{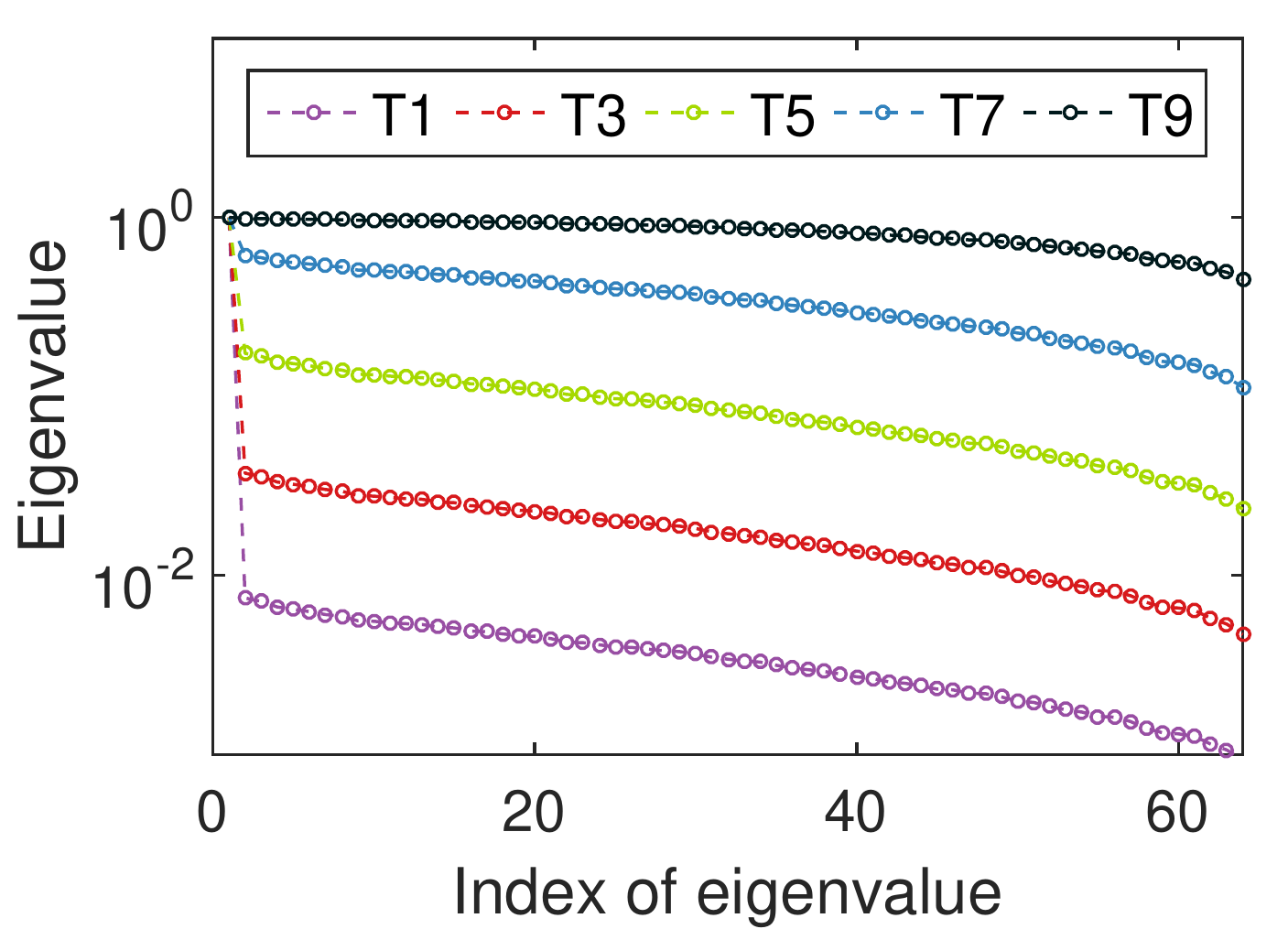}
			\end{minipage}
		}
		\vspace{-0.05in}
		\caption{Convergence behaviors of the proposed Orthogonalization by Newton's Iteration. The entries of proxy matrix $\mathbf{Z} \in \mathbb{R}^{64 \times 256}$ are sampled from the Gaussian distribution $N(3, 1)$.  We show (a) the magnitude of the orthogonality, measured as $\delta= \|\mathbf{W} \mathbf{W}^T -\mathbf{I} \|_{F}$, with respect to the iterations and (b) the distribution (log scale) of the eigenvalues of $\mathbf{W} \mathbf{W}^T$ with different iterations.}
		\label{fig:NI}
		\vspace{-0.1in}
	\end{figure}
\end{tiny}

\vspace{-0.1in}
\subsection{Preliminaries}
\label{sec:prelim}
This paper starts with learning orthogonal filter banks (row orthogonalization of a weight matrix) for deep neural networks (DNNs). We assume $n\leq d$ for simplicity, and will discuss the situation where $n > d$ in Section \ref{sec:RowAndColumn}. This problem is formulated in \cite{2018_AAAI_Huang} as an optimization with layer-wise orthogonal constraints, as follows:
\begin{small}
	 	 	\setlength\abovedisplayskip{0.03in} 
	 	 	\setlength\belowdisplayskip{0.03in}
	\begin{eqnarray}
	\label{eqn:objective}
	\theta^* & = \arg \min_{\theta} \mathbb{E}_{(\mathbf{x},\mathbf{y})\in D} \left[\mathcal{L} \left(\mathbf{y}, f \left(\mathbf{x}; \theta \right) \right) \right]  \nonumber \\
	& s.t.~~~~~  \mathbf{W}^{l} (\mathbf{W}^l)^{T} =\mathbf{I}, ~~~~l=1,2,...,L.
	\end{eqnarray}
\end{small}
\hspace{-0.07in} To solve this problem directly, Huang \etal \cite{2018_AAAI_Huang} proposed to use the proxy parameters $\mathbf{V}$ and construct the orthogonal weight matrix $\mathbf{W}$ by minimizing them in a Frobenius norm over the feasible transformation sets,  where the objective is:
\vspace{-0.06in}
\begin{small}
	\begin{eqnarray}
	\label{eqn:orthogonalTransform}
	& \min_{\phi(\mathbf{V})} tr \left( \left(\mathbf{W}-\mathbf{V} \right) \left(\mathbf{W}-\mathbf{V} \right)^T \right)  \nonumber \\
	& ~~~~~~ s.t. ~~~~ \mathbf{W}=\phi(\mathbf{V}) ~~~and ~~\mathbf{W} \mathbf{W}^T = \mathbf{I}.
	\end{eqnarray}
\end{small}
\hspace{-0.05in}They solved this in a closed-form, with the orthogonal transformation as:
\begin{small}
		\setlength\abovedisplayskip{0.03in} 
		\setlength\belowdisplayskip{0.03in}
	\begin{eqnarray}
	\label{eqn:orthogonalTransform_SVDSolution}
	\mathbf{W}=\phi(\mathbf{V})=\mathbf{D}\Lambda^{-1/2} \mathbf{D}^T \mathbf{V},
	\end{eqnarray}
\end{small}
\hspace{-0.05in}where $\Lambda=\{\lambda_1,...,\lambda_n \}$ and $\mathbf{D}$ are the eigenvalues and corresponding eigenvectors of the covariance matrix $\mathbf{S}=\mathbf{V} \mathbf{V}^T$. 
Given the gradient  $\frac{\partial \mathcal{L}}{\partial \mathbf{W}}$, back-propagation must pass through the orthogonal transformation to calculate $\frac{\partial \mathcal{L}}{\partial \mathbf{V}}$ for updating $\mathbf{V}$.
The closed formulation is concise; however, it encounters the following problems in practice: 1) Eigen decomposition is required, which is computationally expensive, especially on GPU devices \cite{2017_BMVC_Lin};
2) The back-propagation through the eigen decomposition requires the element-wise multiplication of a matrix $\mathbf{K}$ \cite{2018_AAAI_Huang}, whose elements are given by $\mathbf{K}_{i,j}=\frac{1}{(\lambda_{i}-\lambda_{j})}$, where $i\neq j$. This may cause numerical instability when there exists equal eigenvalues of  $\mathbf{S}$, which is discussed in \cite{2015_ICCV_Ionescu,2017_BMVC_Lin} and observed in our preliminary experiments, especially for high-dimensional space.

We observe that the solution of Eqn. \ref{eqn:orthogonalTransform}  can be represented as $\mathbf{W}=\mathbf{S}^{- \frac{1}{2}} \mathbf{V} $, where  $\mathbf{S}^{-\frac{1}{2}}$ can be computed by Newton's iteration \cite{1950_ChemicalPhysics_Lwdin,2005_NumerialAlg,2019_CVPR_Huang}, which avoids eigen decomposition in the forward pass and potential numerical instability during the back-propagation. 

	\begin{algorithm}[t]
		\algsetup{linenosize=\footnotesize}
		\small
		\caption{\small  ONI with Acceleration.}
		\label{alg_forward_acc}
		\begin{algorithmic}[1]
			\STATE \textbf{Input}: proxy parameters $\mathbf{Z} \in \mathbb{R}^{n \times d} $ and iteration numbers $N$.
			\STATE  Centering: $\mathbf{Z}_c= \mathbf{Z} - \frac{1}{d}\mathbf{Z}\mathbf{1} \mathbf{1}^T$.
			\STATE  Bounding  $\mathbf{Z}$'s singular values: $\mathbf{V}=\frac{\mathbf{Z}_c}{\sqrt{\| \mathbf{Z}_c \mathbf{Z}_c^T \|_F}}$.
			\STATE  Execute Step. 3 to 8 in Algorithm \ref{alg_forward}.
			\STATE \textbf{Output}:  orthogonalized weight matrix: $\mathbf{W} \in \mathbb{R}^{n \times d}$.
		\end{algorithmic}
	\end{algorithm}
	\vspace{-0.1in}
\subsection{Orthogonalization by Newton's Iteration}
\label{sec:motivation}

Newton's iteration calculates $\mathbf{S}^{-\frac{1}{2}}$ as follows:
\begin{small}
	 	 	\setlength\abovedisplayskip{0.03in} 
	 	 	\setlength\belowdisplayskip{0.03in}
	\begin{equation}
	\label{eqn:Iteration}
	\begin{cases}
	\mathbf{B}_0=\mathbf{I} \\
	\mathbf{B}_{t}=\frac{1}{2} (3 \mathbf{B}_{t-1} - \mathbf{B}_{t-1}^{3} \mathbf{S}), ~~ t=1,2,...,T,
	\end{cases}
	\end{equation}
\end{small}
\hspace{-0.06in} where $T$ is the number of iterations.
Under the condition that $\|\mathbf{I} -\mathbf{S} \|_2 <1$, $\mathbf{B}_T$  will converge to $\mathbf{S}^{-\frac{1}{2}}$ \cite{2005_NumerialAlg,2019_CVPR_Huang}. 

$\mathbf{V}$ in Eqn. \ref{eqn:orthogonalTransform} can be initialized to ensure that $\mathbf{S}=\mathbf{V} \mathbf{V}^T$  initially satisfies the convergence condition, e.g. ensuring $0 \leq \sigma(\mathbf{V}) \leq 1$, where $\sigma(\mathbf{V})$ are the singular values of $\mathbf{V}$. However, the condition is very likely to be violated when training DNNs, 
since $\mathbf{V}$ varies. 

To address this problem, we propose to maintain another proxy parameter $\mathbf{Z}$ and conduct a transformation $\mathbf{V}=\phi_N(\mathbf{Z})$, such that $0 \leq \sigma(\mathbf{V}) \leq 1$, inspired by the re-parameterization method \cite{2016_CoRR_Salimans, 2018_AAAI_Huang}.
One straightforward way to ensure $0 \leq \sigma(\mathbf{V}) \leq 1$ is to divide the spectral norm of  $\mathbf{Z}$, like the spectral normalization method does \cite{2018_ICLR_Miyato}. However, it is computationally expensive to accurately calculate the spectral norm, since singular value decomposition is required.
We thus propose to divide the Frobenius norm of $\mathbf{Z}$ to perform spectral bounding: 
\begin{small}
	 	 	\setlength\abovedisplayskip{0.03in} 
	 	 	\setlength\belowdisplayskip{0.03in}
	\begin{equation}
	\label{eqn:FrobeniusNorm}
	\mathbf{V}=\phi_N(\mathbf{Z}) =\frac{\mathbf{Z}}{\| \mathbf{Z} \|_F}.
	\end{equation}
\end{small}
\hspace{-0.07in} It's easy to demonstrate that Eqn. \ref{eqn:FrobeniusNorm}  satisfies the convergence condition of Newton's iteration and we show that this method is equivalent to the Newton's iteration proposed in \cite{2019_CVPR_Huang} (See \TODO{\SM}~\ref{sup:proofConverge}  for details).
  Algorithm \ref{alg_forward} describes the proposed method, referred to as Orthogonalization by Newton's Iteration (ONI),
and its corresponding back-propagation is shown in \TODO{\SM}~\ref{sec-sup-backProp}.
We find that Algorithm \ref{alg_forward} converges well (Figure \ref{fig:NI}). However, the concern is the speed of convergence, since 10 iterations are required in order to obtain a good orthogonalization.  We thus further explore methods to speed up the convergence of ONI.

	\vspace{-0.1in}
\subsection{Speeding up Convergence of Newton's Iteration}
\label{sec-speedONI}
Our Newton's iteration proposed for obtaining orthogonal matrix $\mathbf{W}$ works by iteratively stretching the singular values of $\mathbf{V}$ towards 1, as shown in Figure \ref{fig:NI} (b). The speed of convergence depends on how close the singular values of $\mathbf{V}$ initially are to $1$ \cite{2005_NumerialAlg}. 
We observe that the following factors benefit the convergence of Newton's iteration: 1) The singular values of $\mathbf{Z}$ have  a  balanced distribution, which can be evaluated by the condition number of the matrix $\mathbf{Z}$; 2) The singular values of $\mathbf{V}$  should be as close to 1 as possible after spectral bounding (Eqn. \ref{eqn:FrobeniusNorm}).

\begin{figure}[t]
	\centering
		\vspace{-0.1in}
	\subfigure[]{
		\begin{minipage}[c]{.46\linewidth}
			\centering
			\includegraphics[width=3.8cm]{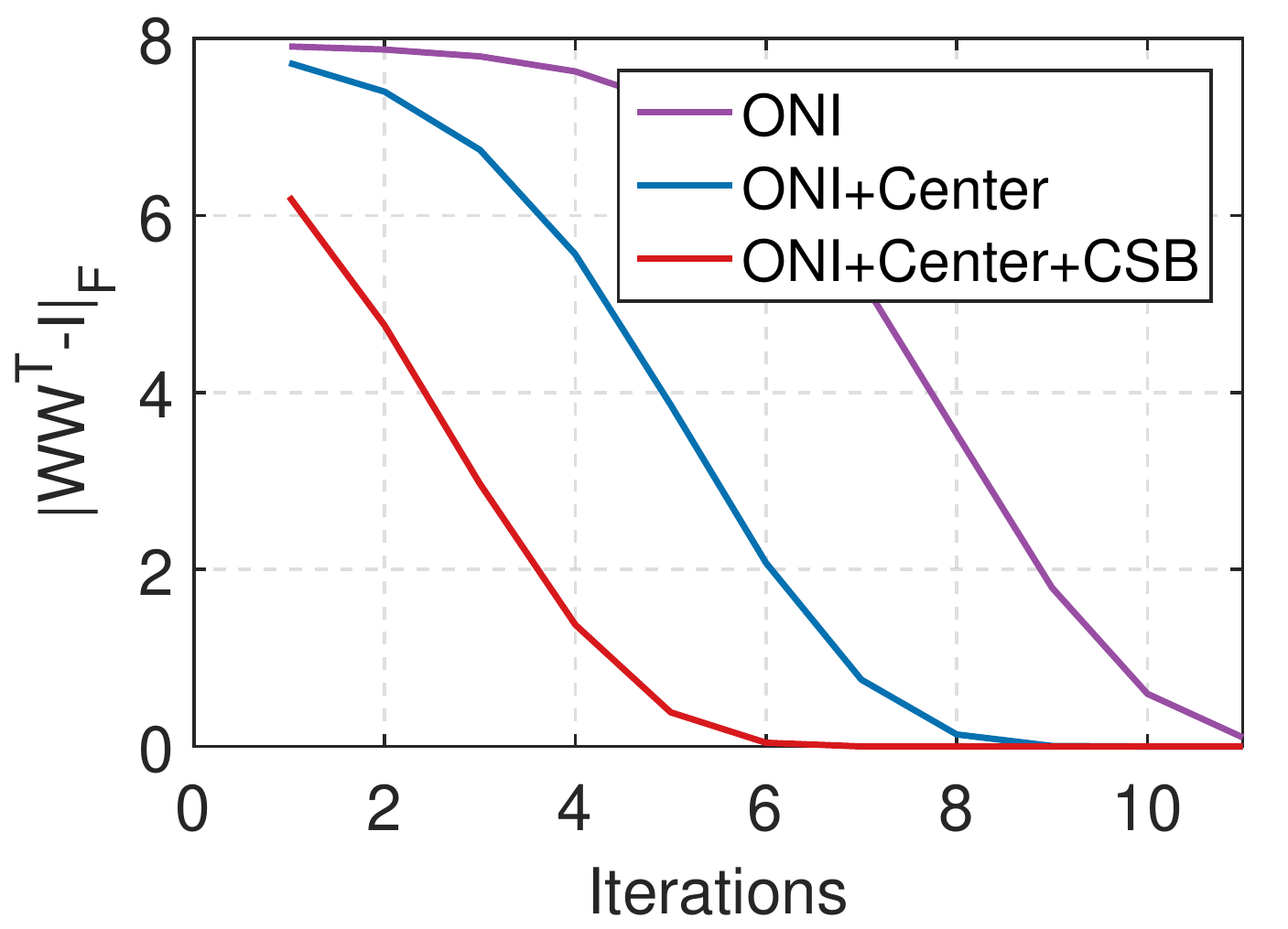}
		\end{minipage}
	}
	\subfigure[]{
		\begin{minipage}[c]{.46\linewidth}
			\centering
			\includegraphics[width=3.8cm]{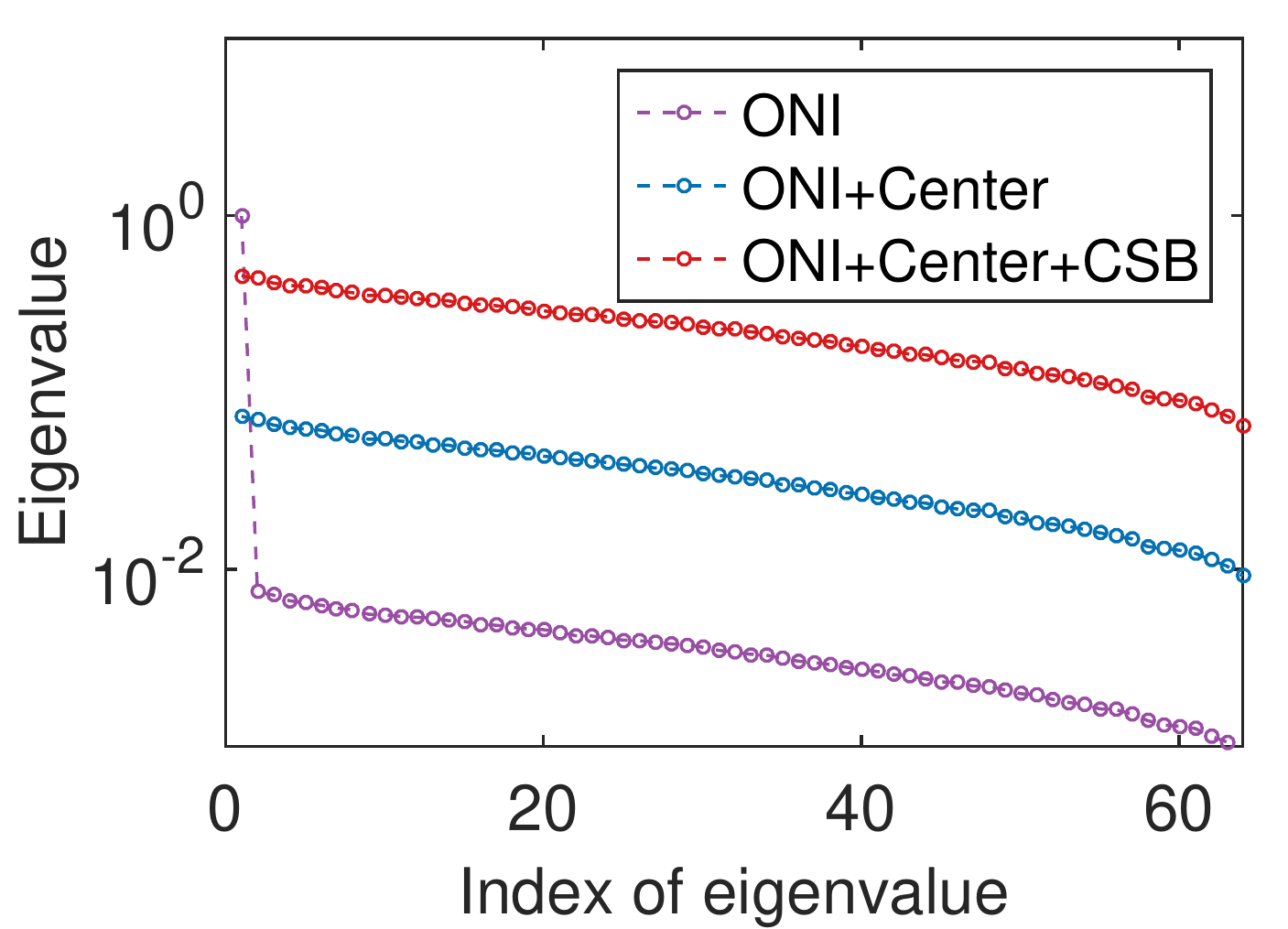}
		\end{minipage}
	}
		\vspace{-0.05in}
	\caption{Analysis of speeding up Newton's iteration. The entries of proxy matrix $\mathbf{Z} \in \mathbb{R}^{64 \times 256}$ are sampled from the Gaussian distribution $N(3, 1)$.  (a) Comparison of convergence;  (b) Comparison of the distribution of the eigenvalues of $\mathbf{W} \mathbf{W}^T$ at iteration $t=1$. }
	\label{fig:NI_accelerate}
	\vspace{-0.08in}
\end{figure}

\vspace{-0.16in}
\paragraph{Centering}
To achieve more balanced distributions for the eigenvalues of $\mathbf{Z}$, we perform  a centering operation over the proxy parameters $\mathbf{Z}$, as follows
\begin{small}
	 	 	\setlength\abovedisplayskip{0.03in} 
	 	 	\setlength\belowdisplayskip{0.03in}
	\begin{equation}
	\label{eqn:centering}
	\mathbf{Z}_c= \mathbf{Z} - \frac{1}{d}\mathbf{Z}\mathbf{1} \mathbf{1}^T.
	\end{equation}
\end{small}
\hspace{-0.07in}  The orthogonal transformation is then performed over the centered parameters $\mathbf{Z}_c$. As shown in \cite{1998_NN_Yann,1998_Schraudolph}, the covariance matrix of centered matrix $\mathbf{Z_c}$ is  better conditioned than $\mathbf{Z}$.
We also experimentally observe that  orthogonalization over centered parameters $\mathbf{Z}_c$ (indicated as `ONI+Center')  produces larger singular values on average at the initial stage (Figure \ref{fig:NI_accelerate} (b)), and thus converges  faster than the original ONI (Figure \ref{fig:NI_accelerate} (a)). 

\vspace{-0.14in}
\paragraph{Compact Spectral Bounding}
To achieve larger singular values of $\mathbf{V}$ after spectral bounding, we seek a more compact spectral bounding factor $f(\mathbf{Z})$ such that $f(\mathbf{Z})> \| \mathbf{Z} \|_F $ and $\mathbf{V}$ satisfies the convergence condition. We find that $f(\mathbf{Z})=\sqrt{\| \mathbf{Z} \mathbf{Z}^T \|_F}$ satisfies the requirements, which is demonstrated in  \TODO{\SM}~\ref{sup:proofConverge}. We thus perform spectral bounding based on the following formulation: 
\begin{small}
	 	 	\setlength\abovedisplayskip{0.03in} 
	 	 	\setlength\belowdisplayskip{0.03in}
	\begin{equation}
	\label{eqn:FrobeniusNorm_sqrt}
	\mathbf{V}=\phi_N(\mathbf{Z}) =\frac{\mathbf{Z}}{\sqrt{\| \mathbf{Z} \mathbf{Z}^T \|_F}}.
	\end{equation}
\end{small}
\hspace{-0.05in}   More compact spectral bounding (CSB) is achieved using Eqn. \ref{eqn:FrobeniusNorm_sqrt}, compared to Eqn. \ref{eqn:FrobeniusNorm}. 
For example, assuming that $\mathbf{Z}$ has $n$ equivalent singular values, the initial singular values of $\mathbf{V}$ after spectral bounding will be  $\frac{1}{\sqrt[4]{n}}$ when using Eqn. \ref{eqn:FrobeniusNorm_sqrt}, while $\frac{1}{\sqrt{n}}$ when using Eqn. \ref{eqn:FrobeniusNorm}.
We also experimentally observe that  using Eqn. \ref{eqn:FrobeniusNorm_sqrt}  (denoted with `+CSB' in Figure \ref{fig:NI_accelerate})  results in a significantly faster convergence.   

Algorithm \ref{alg_forward_acc} describes the accelerated ONI method with centering and more compact spectral bounding (Eqn. \ref{eqn:FrobeniusNorm_sqrt}).
	\vspace{-0.06in}
\begin{figure}[t]
	\centering
	\vspace{-0.1in}
	\subfigure[]{
		\begin{minipage}[c]{.46\linewidth}
			\centering
			\includegraphics[width=3.8cm]{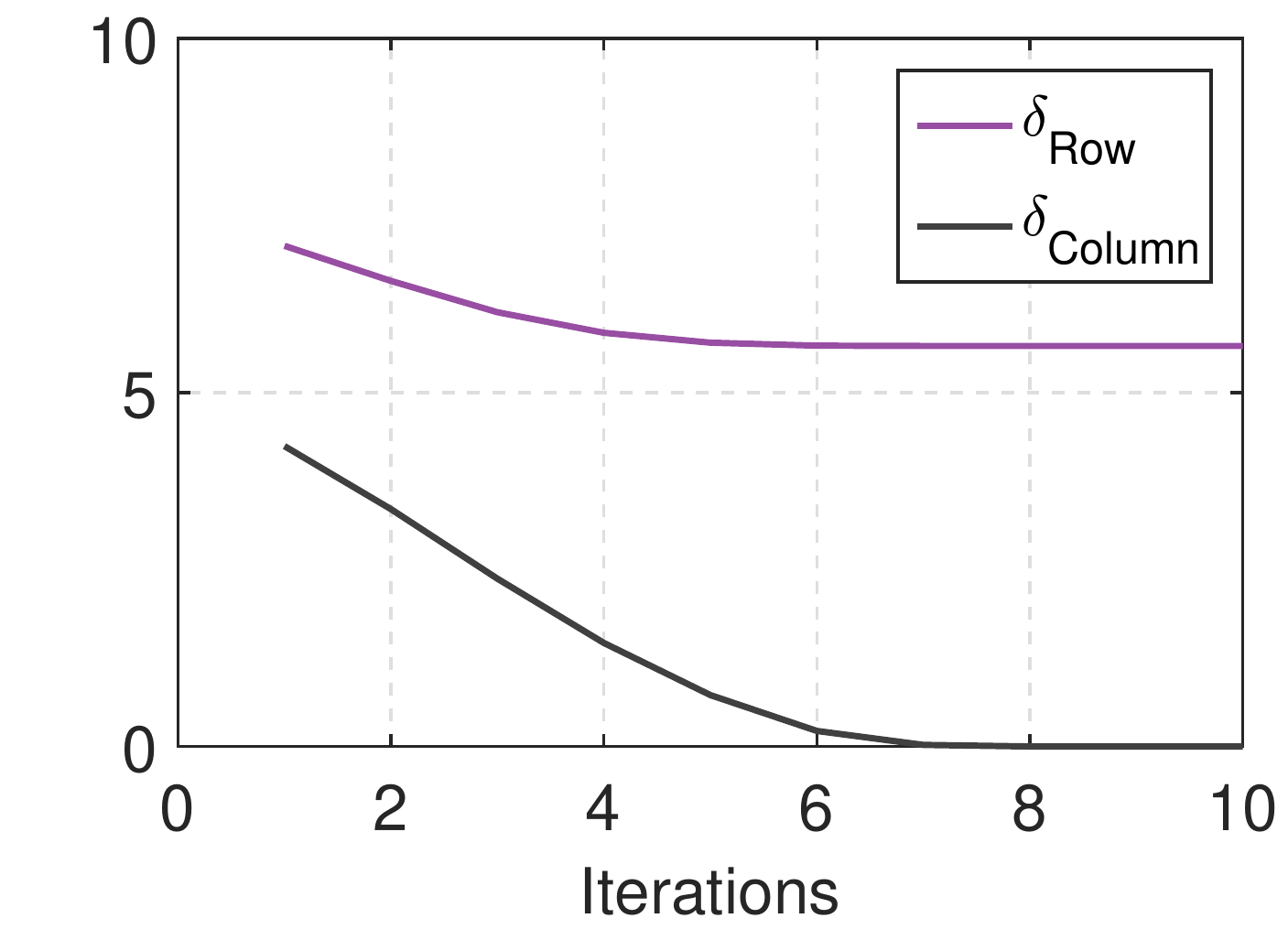}
		\end{minipage}
	}
	\subfigure[]{
		\begin{minipage}[c]{.46\linewidth}
			\centering
			\includegraphics[width=3.8cm]{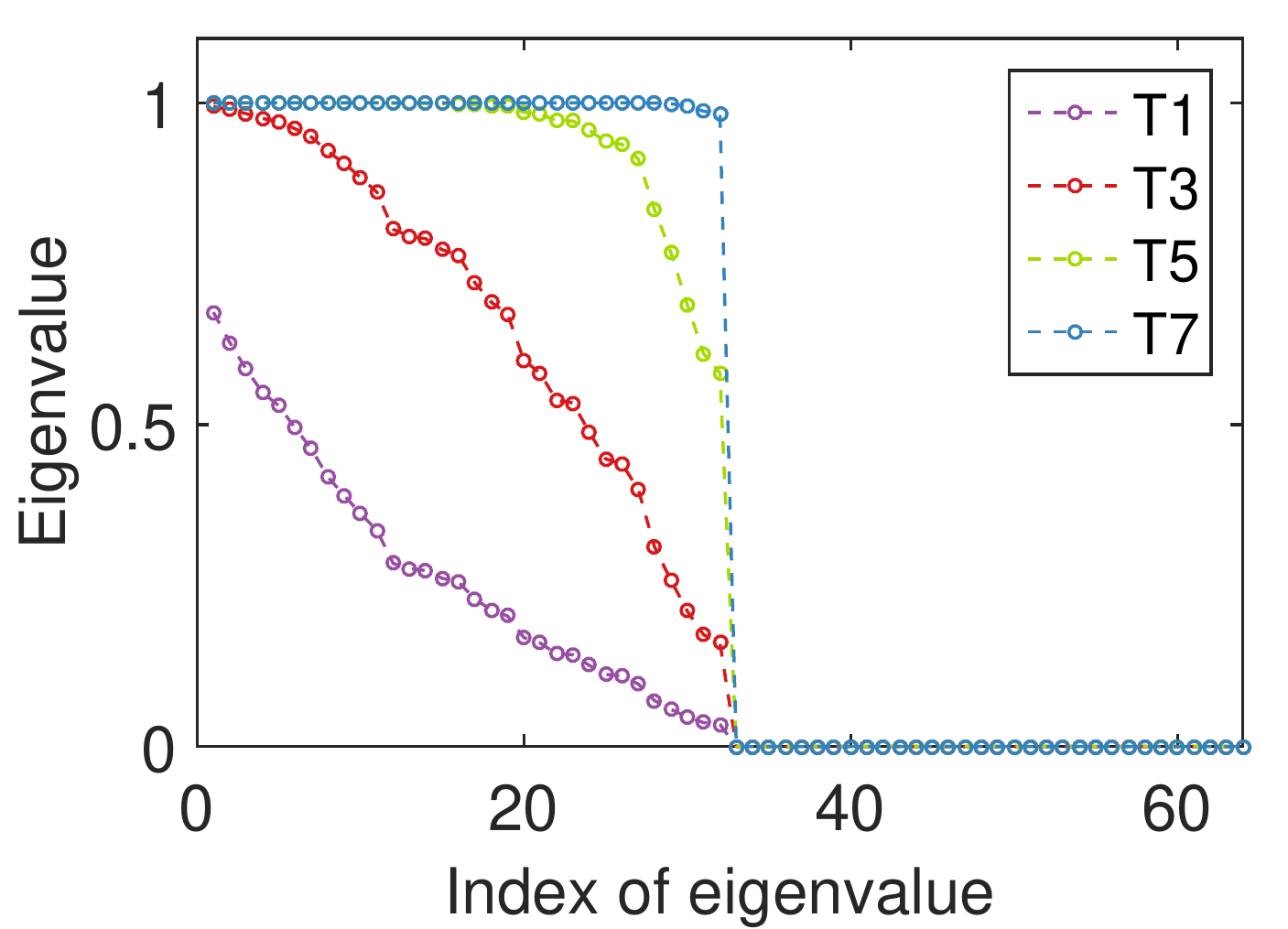}
		\end{minipage}
	}
		\vspace{-0.05in}
	\caption{Unified row and column orthogonalization. The entries of proxy matrix $\mathbf{Z} \in \mathbb{R}^{64 \times 32}$ are sampled from the Gaussian distribution $N(0, 1)$.  (a) Orthogonalization comparison between $\delta_{Row}=\| \mathbf{W} \mathbf{W}^T - \mathbf{I} \|_F$ and  $\delta_{Column}=\| \mathbf{W}^T \mathbf{W} - \mathbf{I} \|_F$;  (b) The distribution of the eigenvalues of $\mathbf{W} \mathbf{W}^T$ with different iterations.}
	\label{fig:Column-Orthonalization}
	\vspace{-0.18in}
\end{figure}

\subsection{Unified Row and Column Orthogonalization}
\label{sec:RowAndColumn}
In previous sections, we assume $n \leq d$, and obtain an  orthogonalization solution. One question that remains is how to handle the situation when $n>d$. When $n>d$, the rows of $\mathbf{W}$ cannot be orthogonal, because the rank of $\mathbf{W}$ is  less than/equal to $d$. Under this situation, 
full orthogonalization using the eigenvalue decomposition based solution (Eqn. \ref{eqn:orthogonalTransform_SVDSolution}) may cause numerical instability, since there exists  at least $n-d$ zero eigenvalues for the covariance matrix.  These zero eigenvalues specifically lead to  numerical instability during back-propagation (when element-wisely multiplying the scaling matrix $\mathbf{K}$, as discussed in Section \ref{sec:prelim}).   

Our orthogonalization solution by Newton's iteration can avoid such problems, since there are no operations relating to dividing the eigenvalues of the covariance matrix. Therefore, our ONI can solve Eqn. \ref{eqn:orthogonalTransform} under the situation  $n>d$. 
More interestingly, our method  can achieve column orthogonality for the weight matrix $\mathbf{W}$ (that is, $\mathbf{W}^T \mathbf{W}=\mathbf{I}$) by solving Eqn. \ref{eqn:orthogonalTransform} directly under $n>d$. Figure \ref{fig:Column-Orthonalization} shows the convergence behaviors of the row and column orthogonalizations.
We observe ONI stretches the non-zero eigenvalues of the covariance matrix $\mathbf{S}$ towards 1 in an iterative manner, and thus equivalently stretches the singular values of the weight matrix $\mathbf{V}$ towards 1. Therefore, it ensures  column orthogonality under the situation $n>d$.
Our method unifies the row and column orthogonalizations, and  we further show in Section \ref{sec:property} that they both benefit in preserving the norm/distribution of the activation/gradient when training DNNs.



Note that, for $n>d$, Huang \etal \cite{2018_AAAI_Huang} proposed the group based methods  by dividing the weights $\{ w_i\}_{i=1}^{n}$ into groups of size $N_G \leq d$ and performing orthogonalization over each group, such that the weights in each group are row orthogonal. However, such a method cannot ensure the whole matrix $\mathbf{W}$ to be either row or column orthogonal (See \TODO{\SM}~\ref{sup:sec-Group} for details).

\vspace{-0.06in}
\subsection{Controlling Orthogonality}
\label{sec:property}
One remarkable property of the orthogonal matrix is that it can preserve the norm and distribution of the activation for a linear transformation, given appropriate assumptions.
Such properties are  described in the following theorem.
\begin{small}
\begin{theorem}
	\label{th:norm}
	Let $\mathbf{\hat{h}}= \mathbf{W} \mathbf{x}$, where $\mathbf{W} \mathbf{W}^T = \mathbf{I}$ and $\mathbf{W} \in \mathbb{R}^{n \times d}$.	
	Assume:  (1) $\mathbb{E}_{\mathbf{x}}(\mathbf{x})=\mathbf{0}$, $cov(\mathbf{x})=\sigma_1^2 \mathbf{I}$,
	and (2) $\mathbb{E}_{\frac{\partial \mathcal{L}}{\partial \mathbf{\hat{h}}}}(\frac{\partial \mathcal{L}}{\partial \mathbf{\hat{h}}})=\mathbf{0}$,
	$cov(\frac{\partial \mathcal{L}}{\partial \mathbf{\hat{h}}})=\sigma_2^2 \mathbf{I}$.	
	If $n=d$, we have the following properties:
	(1) $\| \mathbf{\hat{h}}  \|=  \|  \mathbf{x}  \|$;
	(2) $\mathbb{E}_{\mathbf{\hat{h}}}(\mathbf{\hat{h}})=\mathbf{0}$, $cov(\mathbf{\hat{h}})=\sigma_1^2 \mathbf{I}$;
	(3) $\| \frac{\partial \mathcal{L}}{\partial \mathbf{x}}  \|=  \|  \frac{\partial \mathcal{L}}{\partial \mathbf{\hat{h}}} \|$;
	(4) $\mathbb{E}_{  \frac{\partial \mathcal{L}}{\partial \mathbf{x}}  }( \frac{\partial \mathcal{L}}{\partial \mathbf{x}} )=\mathbf{0}$, $cov(\frac{\partial \mathcal{L}}{\partial \mathbf{x}})=\sigma_2^2 \mathbf{I}$.
	In particular, if $n<d$, property (2) and (3) hold; if $n>d$, property (1) and (4) hold.
\end{theorem}
\end{small}
The  proof is provided in \TODO{\SM}~\ref{sup:sec:proofTheorem}. Theorem \ref{th:norm} shows the benefits of orthogonality in preventing gradients from exploding/vanishing, from an optimization perspective. 
Besides, the orthonormal weight matrix can be viewed  as the embedded Stiefel manifold $\mathcal{O}^{n \times d}$ with a degree of freedom $nd- n(n+1)/2$~\cite{2008_Book_Absil,2018_AAAI_Huang}, which regularizes the neural networks and can improve the model's generalization~\cite{2008_Book_Absil,2018_AAAI_Huang}.  

However, this regularization may harm the representational capacity and result in degenerated performance, as shown in \cite{2018_AAAI_Huang} and observed in our experiments. Therefore, controlling orthogonality is necessary to balance the increase in optimization benefit and reduction in representational capacity, when training DNNs. Our ONI can effectively control orthogonality using different numbers of iterations.

\vspace{-0.06in}
\subsection{Learning Orthogonal Weight Matrices in DNNs }
\label{sec:analysis}
Based on Algorithm \ref{alg_forward_acc} and its corresponding backward pass, we can wrap our method in linear modules \cite{2016_CoRR_Salimans, 2018_AAAI_Huang}, to learn filters/weights with orthogonality constraints for DNNs. After training,  we calculate the weight matrix $\mathbf{W}$ and save it for inference, as in the  standard module.

\vspace{-0.16in}
\paragraph{Layer-wise Dynamical Isometry }
Theorem \ref{th:norm} shows 
that the orthogonal matrix has remarkable properties for preserving the norm/distributions of activations during  the forward and backward passes, for linear transformations. However, in practice, we need to consider the nonlinearity function as well.
Here, we show that we can use an additional constant to scale the magnitude of the weight matrix for ReLU nonlinearity \cite{2010_ICML_Nair}, such that the output-input Jacobian matrix of each layer has dynamical isometry.
\vspace{-0.06in}
\begin{small}
\begin{theorem}
	\label{th:Jacobi}
	Let $\mathbf{h}= max (0,\mathbf{W} \mathbf{x})$, where $\mathbf{W} \mathbf{W}^T = \sigma^2 \mathbf{I}$ and $\mathbf{W} \in \mathbb{R}^{n \times d}$.  Assume $\mathbf{x}$ is a normal distribution with  $\mathbb{E}_{\mathbf{x}}(\mathbf{x})=\mathbf{0}$, $cov(\mathbf{x})=\mathbf{I}$. Denote the Jacobian matrix as $ \mathbf{J}= \frac{\partial  \mathbf{h}}{\partial \mathbf{x}} $. If $\sigma^2 =2$, we have  $\mathbb{E}_\mathbf{x} (\mathbf{J} \mathbf{J}^T) = \mathbf{I}$.
\end{theorem}
\end{small}
The  proof is shown in \TODO{\SM}~\ref{sup:sec:proofTheorem}.
We propose to multiply the orthogonal weight matrix $\mathbf{W}$ by a factor of $\sqrt{2}$ for networks with ReLU activation. 
We experimentally show  this improves the training efficiency in Section \ref{sec:classification}. 
Note that  Theorems \ref{th:norm} and \ref{th:Jacobi} are based on the assumption  that the layer-wise input is Gaussian. Such a property can be approximately satisfied  using batch normalization (BN) \cite{2015_ICML_Ioffe}. Besides, if we apply BN before the linear transformation, there is no need to apply it again after the linear module, since the normalized property of BN is  preserved according to Theorem \ref{alg_forward}.   We experimentally show that  such a process  improves performance in Section \ref{sec:expImageNet}.

\vspace{-0.2in}
\paragraph{Learnable Scalar}
Following \cite{2018_AAAI_Huang}, we relax the constraint of orthonormal to orthogonal, with $\mathbf{W}^T \mathbf{W}=\Lambda$, where $\Lambda$ is the diagonal matrix.
This can be viewed as the orthogonal filters having different contributions to the activations.
To achieve this, we propose to use a learnable scalar parameter to fine-tune the norm of each filter \cite{2016_CoRR_Salimans,2018_AAAI_Huang}.
\vspace{-0.2in}
\paragraph{Convolutional Layer} With regards to the convolutional layer parameterized by weights $\mathbf{W}^C \in \mathbb{R}^{n \times d \times F_h \times F_w}$, where  $F_h$ and  $F_w$ are the height and width of the filter, we reshape $\mathbf{W}^C$ as $\mathbf{W} \in \mathbb{R}^{n \times p }$,  where $p=d \cdot F_h \cdot F_w$, and the  orthogonalization is executed over the unrolled weight matrix $\mathbf{W} \in \mathbb{R}^{n \times (d \cdot F_h \cdot F_w)}$.



\begin{figure}[t]
	\centering
	\vspace{-0.12in}
	\subfigure[]{
		\begin{minipage}[c]{.46\linewidth}
			\centering
			\includegraphics[width=3.8cm]{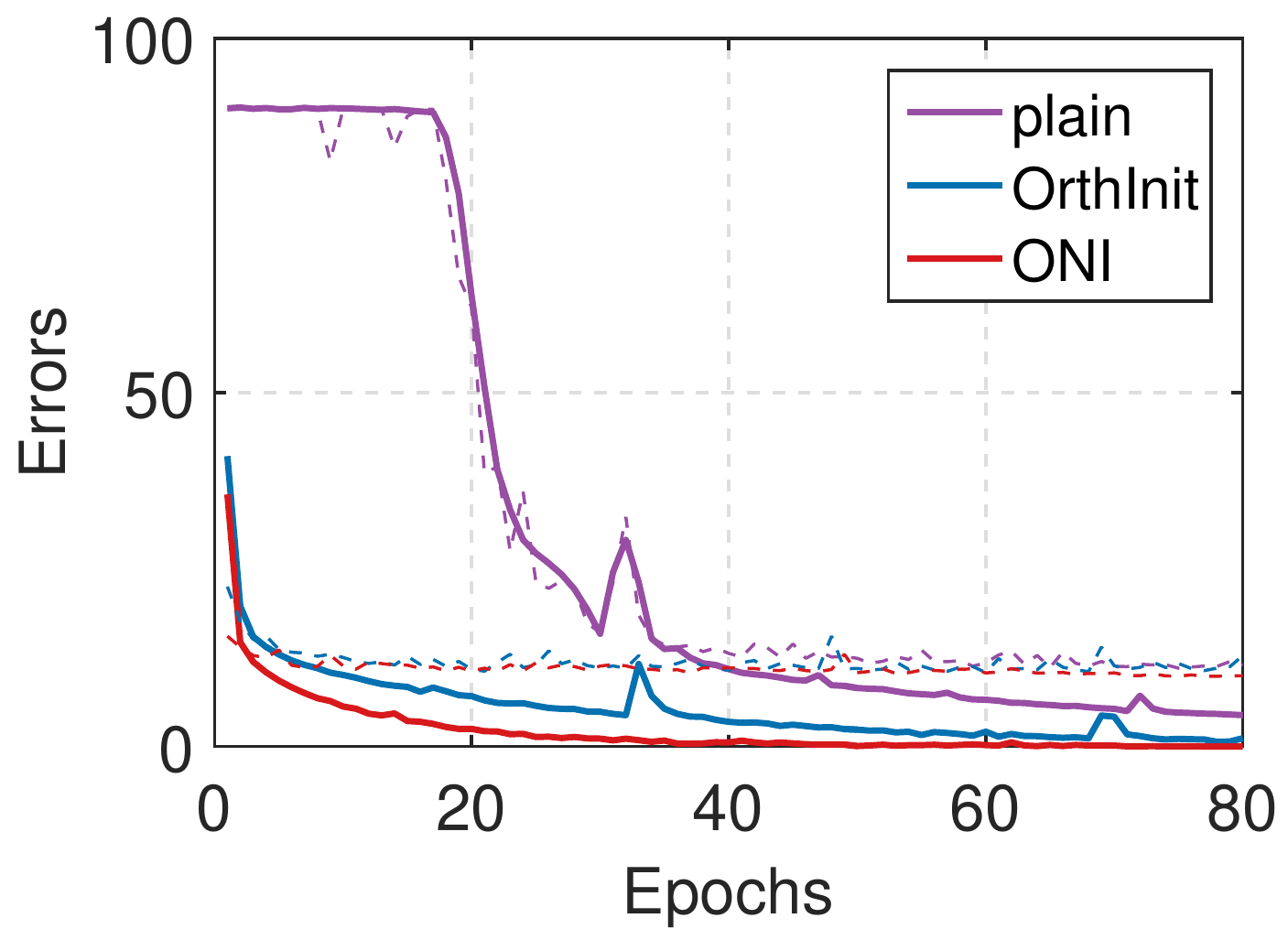}
		\end{minipage}
	}
	\subfigure[]{
		\begin{minipage}[c]{.46\linewidth}
			\centering
			\includegraphics[width=3.8cm]{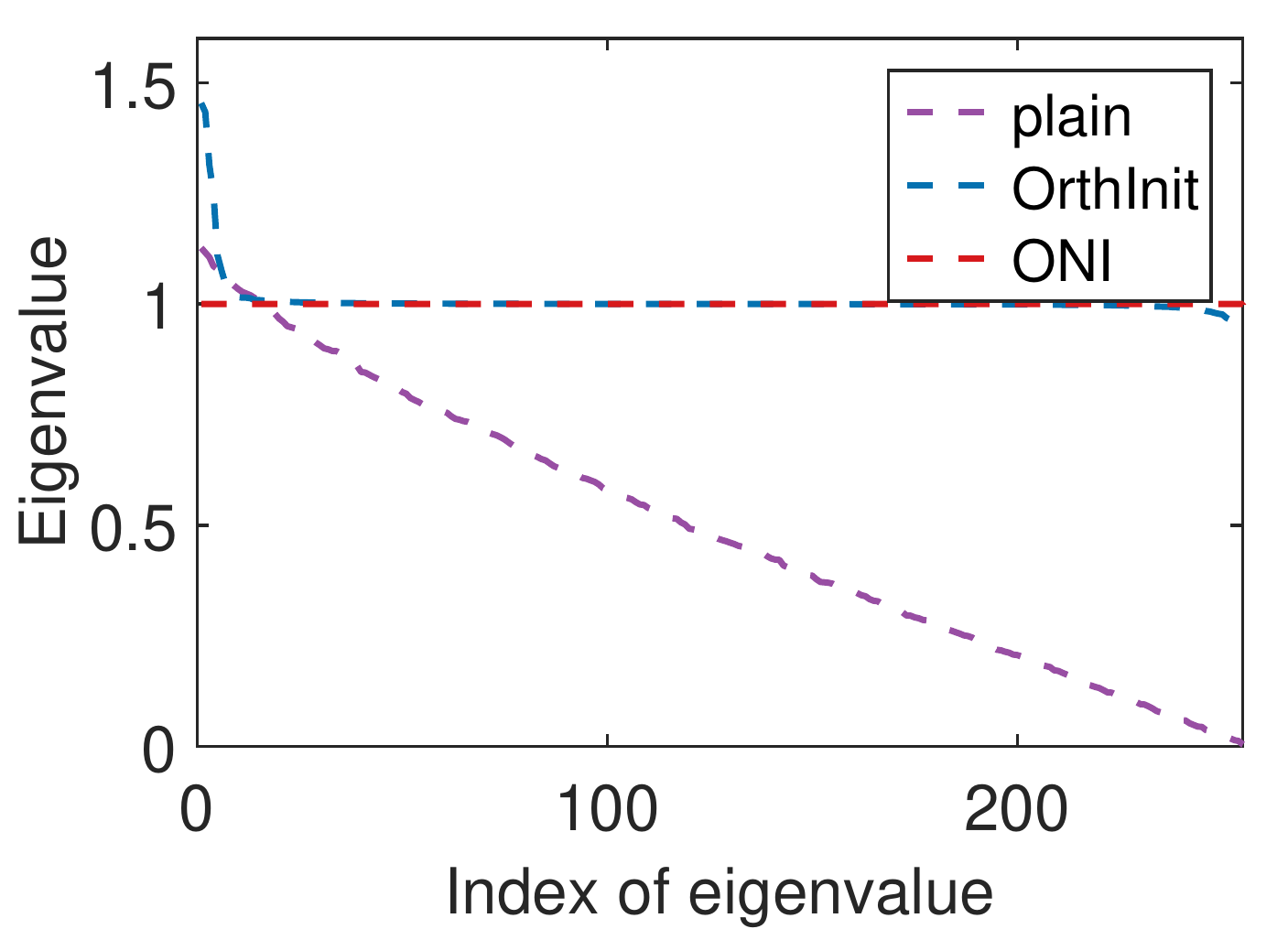}
		\end{minipage}
	}
		\vspace{-0.05in}
	\caption{Effects of maintaining orthogonality. Experiments are performed on a 10-layer MLP. (a) The training (solid lines) and testing (dashed lines) errors with respect to the training epochs; (b) The distribution of eigenvalues of the weight matrix $\mathbf{W}$ of the 5th layer, at the 200 iteration. }
	\label{fig:MO}
	\vspace{-0.1in}
\end{figure}

\begin{figure}[t]
	\centering
	\vspace{-0.06in}
	\subfigure[6-layer MLP]{
		\begin{minipage}[c]{.46\linewidth}
			\centering
			\includegraphics[width=4.0cm]{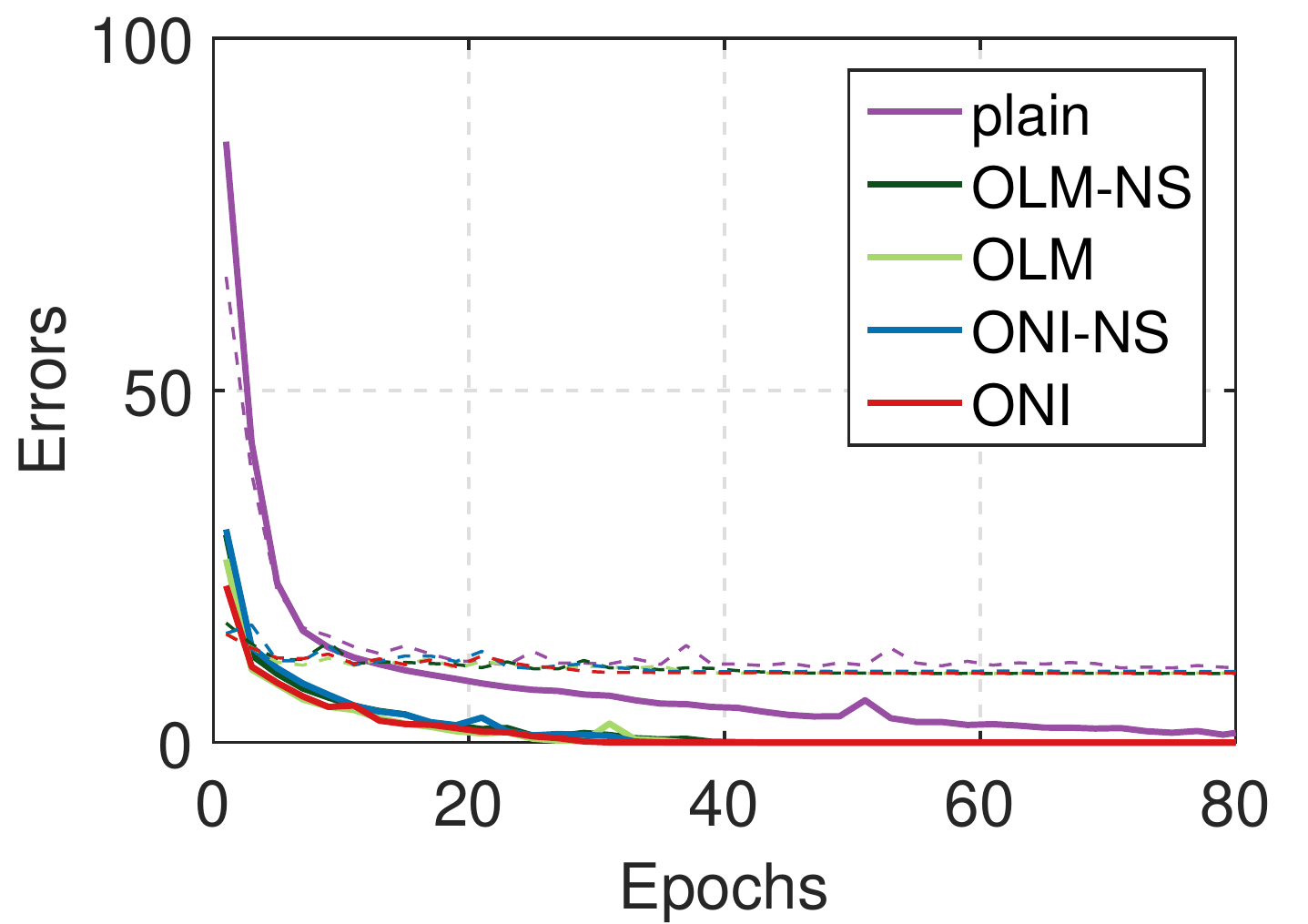}
		\end{minipage}
	}
	\subfigure[20-layer MLP]{
		\begin{minipage}[c]{.46\linewidth}
			\centering
			\includegraphics[width=4.0cm]{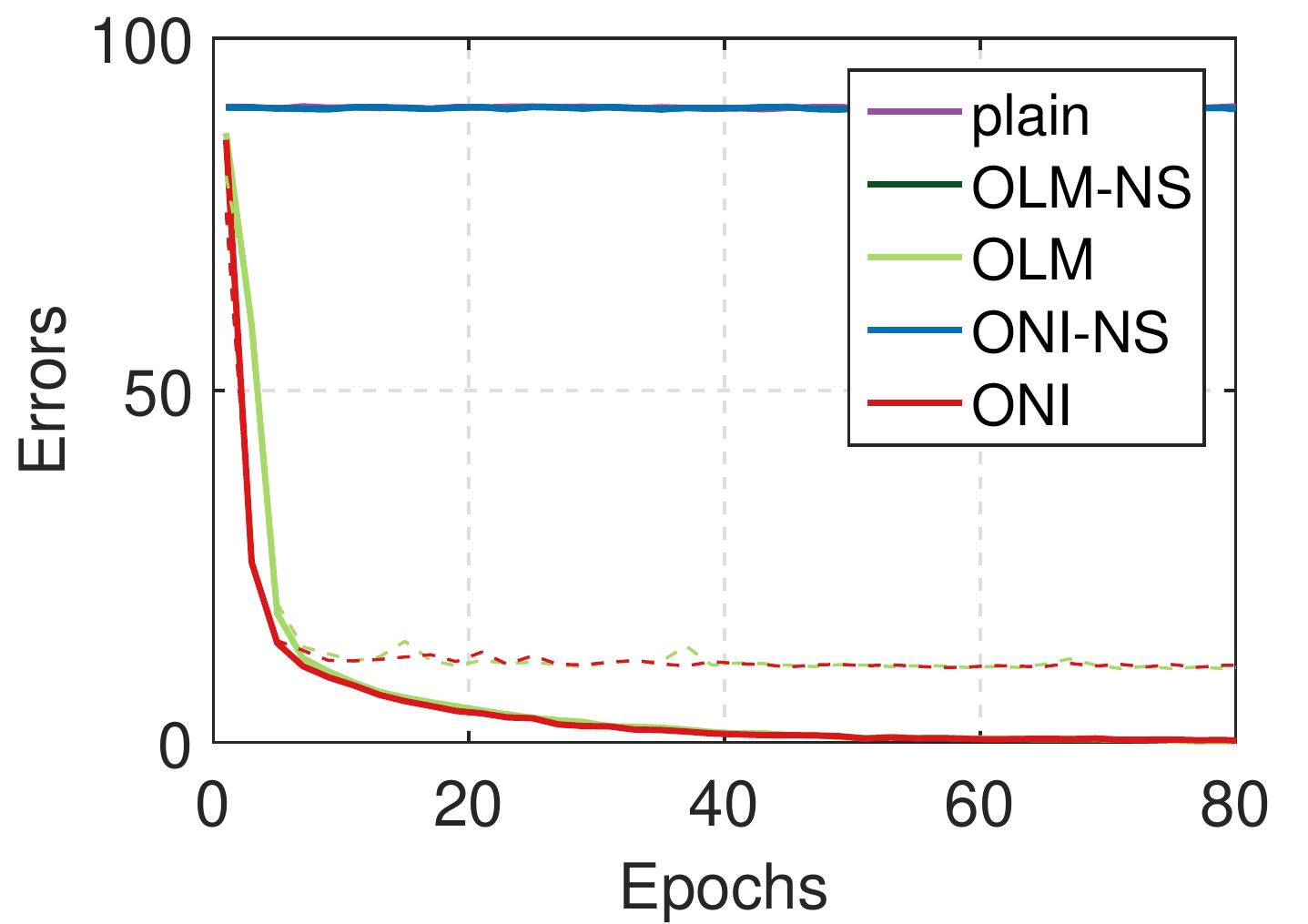}
		\end{minipage}
	}
	\vspace{-0.05in}
	\caption{Effects of scaling the orthogonal weights. `-NS' indicates orthogonalization without scaling by $\sqrt{2}$.  We evaluate the training (solid lines) and testing (dashed lines) errors on (a) a 6-layer MLP and (b) a 20-layer MLP.}
	\label{fig:MLP-scale}
	\vspace{-0.2in}
\end{figure}

\vspace{-0.20in}
\paragraph{Computational Complexity}
Consider a convolutional layer with filters $\mathbf{W} \in \mathbb{R}^{n \times d \times F_h \times F_w}$, and $m$ mini-batch data $\{\mathbf{x}_i \in \mathbb{R}^{d \times h \times w}\}_{i=1}^m$. 
The computational cost of our method, coming mainly from the Lines 3,  6 and  8 in Algorithm \ref{alg_forward}, is $2 n^2 d F_h F_w + 3 N n^3$ for each iteration during training. The relative cost of ONI over the constitutional layer is $\frac{2 n}{m h w} + \frac{3 N n^2}{m d h w F_h F_w}$. 
During inference, we use the orthogonalized weight matrix $\mathbf{W}$, and thus do not introduce additional computational or memory costs.
We provide the wall-clock times in \TODO{\SM}~\ref{sup:sec:time}. 

\begin{table*}[t]	
	\centering
		\vspace{-0.12in}
	\begin{small}
		\begin{tabular}{l|ccccccccc}
			\toprule[1pt]
			methods	& g=2,k=1   &  g=2,k=2   &   g=2,k=3   &   g=3,k=1   &   g=3,k=2   &   g=3,k=3   &   g=4,k=1  &   g=4,k=2  &  g=4,k=3   \\
			\hline
			plain	& 11.34   &  9.84   & 9.47  & 10.32  &   8.73   &  8.55   &   10.66  &   9.00  &  8.43   \\
			WN 	    & 11.19   &   9.55   &   9.49   &  10.26   &   9.26   &   8.19   &   9.90  &  9.33  & 8.90  \\
			OrthInit& 10.57   &   9.49   &   9.33   &   10.34  &   8.94   &   8.28   &   10.35  &  10.6  &  9.39  \\
			OrthReg & 12.01  &  10.33  &   10.31   &  9.78   &   8.69   &   8.61  &  9.39 &  7.92  & 7.24 \\
			
			OLM-1 \cite{2018_AAAI_Huang}  & 10.65   &   8.98   &   8.32   &  9.23   &   8.05   &   7.23   &   9.38  &  7.45  & 7.04  \\
			OLM-$\sqrt{2}$	    & 10.15   &   8.32   &   7.80  &  8.74   &   7.23   &   6.87   &    \textbf{8.02}  &  6.79  & 6.56  \\			
			ONI    & \textbf{9.95}   &  \textbf{ 8.20}   &   \textbf{7.73}   &  \textbf{8.64}   & \textbf{ 7.16}   &  \textbf{ 6.70}   & 8.27 &   \textbf{6.72}  & \textbf{ 6.52}  \\
			\toprule[1pt]
		\end{tabular}
	\end{small}
	\caption{Test errors ($\%$) on VGG-style networks for CIFAR-10 classification. The results are averaged over three independent runs.
	}
	\label{tab:Cifar-VGG}
	\vspace{-0.16in}
\end{table*}

\vspace{-0.06in}
\section{Experiments}
\label{sec:exp}

\vspace{-0.02in}
\subsection{Image Classification}
\label{sec:classification}
\vspace{-0.05in}
We evaluate our ONI on the Fashion-MNIST \cite{2017_FashionMNIST}, CIFAR-10 \cite{2009_TR_Alex} and ImageNet \cite{2015_ImageNet} datasets. We provide an ablation study on the iteration number $T$ of ONI in Section \ref{sec:ablation}. 
Due to space limitations, we only provide essential components of the experimental setup; for more details, please refer to \TODO{\SM}~\ref{sup:sec:experiments}. The code  is available at \textcolor[rgb]{0.33,0.33,1.00}{https://github.com/huangleiBuaa/ONI}.

\vspace{-0.10in}
\subsubsection{MLPs on Fashion-MNIST}
\label{sec:MLP}
We use an MLP with a ReLU activation \cite{2010_ICML_Nair}, and vary the depth. The number of neurons in each layer is 256. 
 We employ stochastic gradient descent (SGD) optimization with a batch size of 256, and the learning rates are selected based on the validation set ($5,000$ samples from the training set) from $\{0.05, 0.1, 0.5, 1\}$. 
\vspace{-0.18in}
\paragraph{Maintaining Orthogonality} We first show that maintaining orthogonality can improve the training performance. We  compare two baselines: 1) `plain', the original network; and 2) `OrthInit', in which the orthogonal initialization \cite{2013_CoRR_Saxe} is used. The training performances are shown in Figure \ref{fig:MO} (a).  We observe orthogonal initialization can improve the training efficiency in the initial phase (comparing with `plain'), after which the benefits of orthogonality degenerate (comparing with `ONI') due to the updating of weights (Figure \ref{fig:MO} (b)). 

\vspace{-0.16in}
\paragraph{Effects of Scaling} We experimentally show the effects of initially scaling the orthogonal weight matrix  by a factor of $\sqrt{2}$. We also apply this technique to the `OLM' method \cite{2018_AAAI_Huang}, in which the orthogonalization is solved by eigen decomposition. We refer to `OLM-NS'/`ONI-NS' as the  `OLM'/`ONI' without  scaling by $\sqrt{2}$. The results are shown in Figure \ref{fig:MLP-scale}. We observe that the scaling technique has no significant effect on shallow neural networks, \eg, the 6-layer MLP. However, for deeper neural networks, it produces significant performance boosts. For example, for the 20-layer MLP, neither `OLM' nor `ONI'  can converge without the additional scaling factors, because the activation and gradient exponentially vanish  (See  \TODO{\SM}~\ref{sup:sec:exp:MLP}).  
Besides,  our `ONI' has a nearly identical performance compared to `OLM', which indicates the effectiveness of our approximate orthogonalization with few iterations (\eg 5). 

\vspace{-0.16in}
\subsubsection{CNNs on CIFAR-10}
\label{sec:CNN_CIFAR10}
\begin{table}[t]
	\centering
	\begin{small}
		\begin{tabular}{l|c|c}
			\toprule[1pt]
			&  BatchSize=128   &  BatchSize=2 \\ 
			\hline
			w/BN*~\cite{2016_CoRR_Zagoruyko}   & 6.61 & --\\
			Xavier Init* \cite{2018_AAAI_Huang}  & 7.78  & -- \\
			Fixup-init* \cite{2019_ICLR_Zhang}  & 7.24 & -- \\
			w/BN  & 6.82 & 7.24\\
			Xavier Init   & 8.43 & 9.74 \\
			GroupNorm  & 7.33 & 7.36 \\
			ONI  & \textbf{ 6.56}  &  \textbf{6.67}\\ 
			\toprule[1pt]
		\end{tabular}
	\end{small}
	\caption{Test errors ($\%$) comparison on 110-layer residual network \cite{2015_CVPR_He} without BN \cite{2015_ICML_Ioffe} under CIFAR-10. 'w/BN' indicates with BN. We report the median of five independent runs. The methods with `*' indicate the results reported in the cited paper. 
	}
	\label{tab:Cifar-NoBN}
	\vspace{-0.16in}
\end{table}
\vspace{-0.02in}
\paragraph{VGG-Style Networks}
Here, we evaluate ONI on VGG-style neural networks with  $3 \times 3$ convolutional layers. The network starts with a convolutional layer of $32k$ filters, where $k$ is the varying width based on different configurations. We then sequentially stack  three blocks, each of which has $g$ convolutional layers with filter numbers of $32k$, $64k$ and $128k$, respectively.
We vary the depth with $g$ in  $\{2,3,4 \}$ and the width with $k$ in $\{1,2,3\}$.
We use SGD with a  momentum of 0.9 and batch size of 128. The best initial learning rate is chosen from $\{0.01, 0.02, 0.05\}$ over the validation set of 5,000 samples from the training set, and we divide the learning rate by 5 at 80 and 120 epochs, ending the training at 160 epochs.
 We compare our `ONI' to several baselines, including  orthogonal initialization \cite{2013_CoRR_Saxe} (`OrthInit'), using soft orthogonal constraints as the penalty term \cite{2017_CVPR_Xie}  (`OrthReg'), weight normalization \cite{2016_CoRR_Salimans} (`WN'), `OLM' \cite{2018_AAAI_Huang} and the `plain' network.
   Note that OLM \cite{2018_AAAI_Huang} originally uses a scale of 1 (indicated as `OLM-1'), and we also apply the proposed scaling by $\sqrt{2}$ (indicated as `OLM-$\sqrt{2}$').


Table \ref{tab:Cifar-VGG} shows the results.  `ONI' and `OLM-$\sqrt{2}$' have significantly better performance under all network configurations (different depths and widths), which demonstrates the beneficial effects of maintaining orthogonality during training. We also observe `ONI' and `OLM-$\sqrt{2}$' converge faster than other baselines, in terms of training epochs (See \TODO{\SM}~\ref{sup:sec:exp:CNN}).
Besides,  our proposed `ONI' achieves slightly better performance than `OLM-$\sqrt{2}$' on average, over all configurations.
Note that we train `OLM-$\sqrt{2}$' with a group size of $G=64$, as suggested in \cite{2018_AAAI_Huang}. We also try  full orthogonalization for `OLM-$\sqrt{2}$'. However, we observe either performance degeneration or numerical instability (\eg, the eigen decomposition cannot converge). 
We argue that the main reason for this is that full orthogonalization solved by OLM over-constrains  the weight matrix, which harms the performance. Moreover, eigen decomposition based methods are more likely to result in numerical instability in  high-dimensional space, due to the element-wise multiplication of a matrix $\mathbf{K}$ during back-propagation \cite{2017_BMVC_Lin}, as discussed in Section \ref{sec:prelim}.


\begin{table}[t]
		\vspace{-0.07in}
	\centering
	\begin{small}
		\begin{tabular}{l|c|c|c}
			\toprule[1pt]
			&  Top-1 ($\%$)   &  Top-5 ($\%$) & Time (min./epoch) \\ 
			\hline
			plain   & 27.47 & 9.08 & 97\\
			WN & 27.33  & 9.07 & 98\\
			OrthInit   & 27.75  & 9.21 & 97\\
			OrthReg  & 27.22 & 8.94& 98\\
			ONI  & \textbf{ 26.31}  &  \textbf{8.38}& 104\\ 
			\toprule[1pt]
		\end{tabular}
	\end{small}
	\caption{Test errors ($\%$)  on  ImageNet validation set (single model and single crop test) evaluated with VGG-16 \cite{2014_CoRR_Simonyan}. The time cost for each epoch is averaged over the training epochs.}
	\label{tab:ImageNetVGG}
	\vspace{-0.1in}
\end{table}

\vspace{-0.16in}
\paragraph{Residual Network without Batch Normalization}
Batch normalization (BN) is essential for stabilizing and accelerating the training \cite{2015_ICML_Ioffe} of DNNs \cite{2015_CVPR_He,2016_CoRR_Huang_a,2016_CoRR_He,2016_CoRR_Szegedy}. It is a standard configuration in residual networks \cite{2015_CVPR_He}. However, it sometimes suffers from the small batch size problem \cite{2017_NIPS_Ioffe,2018_ECCV_Wu} and introduces too much stochasticity  \cite{2018_ICML_Teye} when debugging neural networks. Several studies have tried to train deep residual networks without BN \cite{2017_AAAI_Shang,2019_ICLR_Zhang}.
 Here, we show that, when using our ONI, the residual network without BN  can also be well trained.

The experiments are executed on a 110-layer residual network (Res-110). We follow the same experimental setup as in \cite{2015_CVPR_He}, except that we run the experiments on one GPU. We also compare against the Xavier Init \cite{2010_AISTATS_Glorot,2018_NIPS_Bjorck}, and group normalization (GN) \cite{2018_ECCV_Wu}. ONI can be trained with a large learning rate of 0.1 and  converge faster than BN, in terms of training epochs (See  \TODO{\SM}~\ref{sup:sec:exp:CNN}).  We observe that ONI has slightly  better test performance than BN (Table \ref{tab:Cifar-NoBN}). 
Finally, we also test the performance on a small batch size of 2. We find  ONI continues to  have better performance than BN in this case, and is not sensitive to the batch size, like GN \cite{2018_ECCV_Wu}. 

\vspace{-0.12in}
\subsubsection{Large-scale ImageNet Classification}
\label{sec:expImageNet}

\begin{table}[t]
	\centering
	\begin{small}
		\begin{tabular}{c|cc|cc|cc}
			\toprule
			& \multicolumn{2}{c|}{ResNet w/o BN} &    \multicolumn{2}{c|}{ResNet}  &    \multicolumn{2}{c}{ResNetVar} \\
			Method   & Train   & Test  & Train   & Test  & Train   & Test\\
			\hline
			plain   & 31.76  &  33.84   & 29.33 & 29.64 & 28.82  & 29.56 \\
			ONI  &\textbf{27.05 }&\textbf{31.17}&\textbf{29.28} & \textbf{29.57} & \textbf{28.12}  & \textbf{28.92}\\
			\toprule[1pt]
		\end{tabular}
		\caption{Ablation study on ImageNet with an 18-layer ResNet. We evaluate the top-1 training and test errors ($\%$). }
		\label{tab:res18}
	\end{small}
	\vspace{-0.1in}
\end{table}

\begin{table}[t]
	\centering
	\begin{small}
		\begin{tabular}{c|cc|cc}
			\toprule
			& \multicolumn{2}{c|}{Test error ($\%$)} &    \multicolumn{2}{c}{Time (min./epoch)}   \\
			Method   & 50   & 101  & 50  & 101 \\
			\hline
			ResNet   & 23.85& 22.40  &  66 & 78   \\
			ResNet ~+~ONI   &23.55 &  22.17   & 74 &  92 \\
			ResNetVar  &23.94&22.76& 66 &  78\\
			ResNetVar~+~ONI  &\textbf{23.30}&\textbf{21.89} & 74 &  92 \\
			\toprule[1pt]
		\end{tabular}
		\caption{Results on  ImageNet with the 50- and 101-layer ResNets.}
		\label{tabresnet}
	\end{small}
	\vspace{-0.12in}
\end{table}

To further validate the effectiveness of our ONI on a large-scale dataset, we evaluate it on the ImageNet-2012 dataset.
We  keep almost all the experimental settings the same as the publicly available PyTorch implementation \cite{2017_NIPS_pyTorch}:  We apply SGD with a momentum of 0.9, and a weight decay of 0.0001. We train for 100 epochs in total and  set the initial learning rate to 0.1, lowering it  by a factor of 10 at epochs 30, 60 and 90. For more details on the slight differences among different architectures and methods, see  \TODO{\SM}~\ref{sup:sec:exp:ImageNet}.

\vspace{-0.16in}
\paragraph{VGG Network}
Table \ref{tab:ImageNetVGG} shows the results on the 16-layer VGG \cite{2014_CoRR_Simonyan}. Our `ONI' outperforms `plain', `WN', `OrthInit' and `OrthReg' by a significant margin.
Besides, `ONI' can be trained with a large learning rate of 0.1,  while the other methods cannot (the results are reported for an initial learning rate of 0.01). We also provide the running times in Table \ref{tab:ImageNetVGG}. The additional cost introduced by `ONI' compared to  `plain' is negligible ($7.2\%$).
\vspace{-0.16in}
\paragraph{Residual Network}
We first perform an ablation study on an 18-layer residual network (ResNet)~\cite{2015_CVPR_He}, applying our ONI. We use the original ResNet and the ResNet without BN \cite{2015_ICML_Ioffe}.  We also consider the architecture with BN inserted  after the nonlinearity, which we refer to as `ResNetVar'. We observe that our ONI improves the performance over all three architectures, as shown in Table \ref{tab:res18}. One interesting observation is that ONI achieves the lowest training error on the ResNet without BN, which demonstrates its ability to facilitate optimization for large-scale datasets. We also observe that ONI has no significant difference in performance compared to `plain' on ResNet. One possible reason is that the BN module and residual connection are well-suited for information propagation, causing ONI to have a lower net gain for such a large-scale classification task. However, we observe that, on ResNetVar,  ONI obtains obviously better performance than `plain'. We argue that this boost is attributed to the orthogonal matrix's ability to achieve  approximate dynamical isometry, as described in Theorem \ref{th:Jacobi}. 

 We also apply our ONI on a 50- and 101-layer residual network.  The results are shown in Table \ref{tabresnet}. We again observe that ONI can improve the performance, without introducing significant computational cost. 
\begin{figure}[t]
	\centering
	\vspace{-0.19in}
\hspace{-0.1in}	\subfigure[]{
		\begin{minipage}[c]{.31\linewidth}
			\centering
			\includegraphics[width=2.8cm]{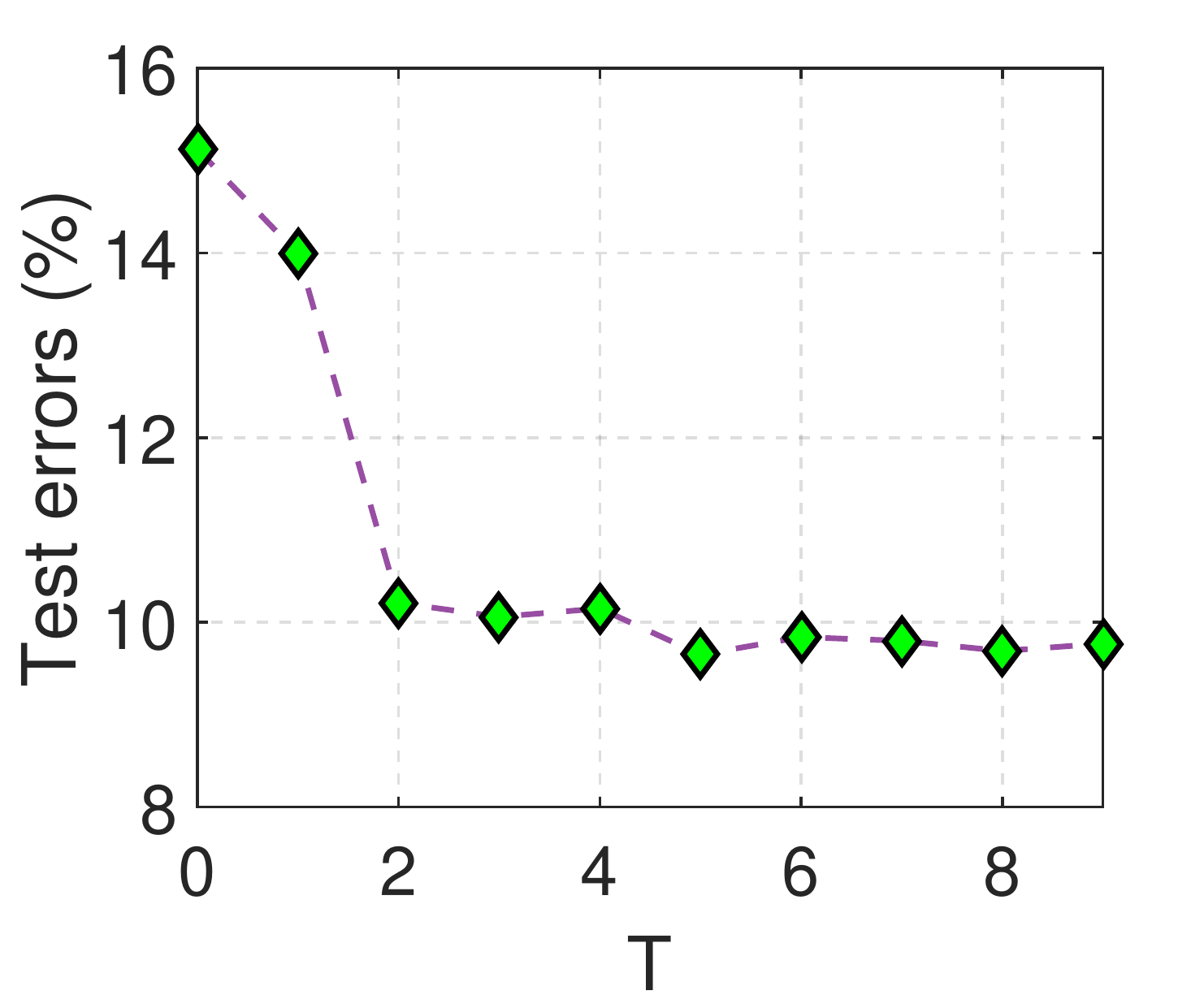}
		\end{minipage}
	}
	\subfigure[]{
		\begin{minipage}[c]{.31\linewidth}
			\centering
			\includegraphics[width=2.8cm]{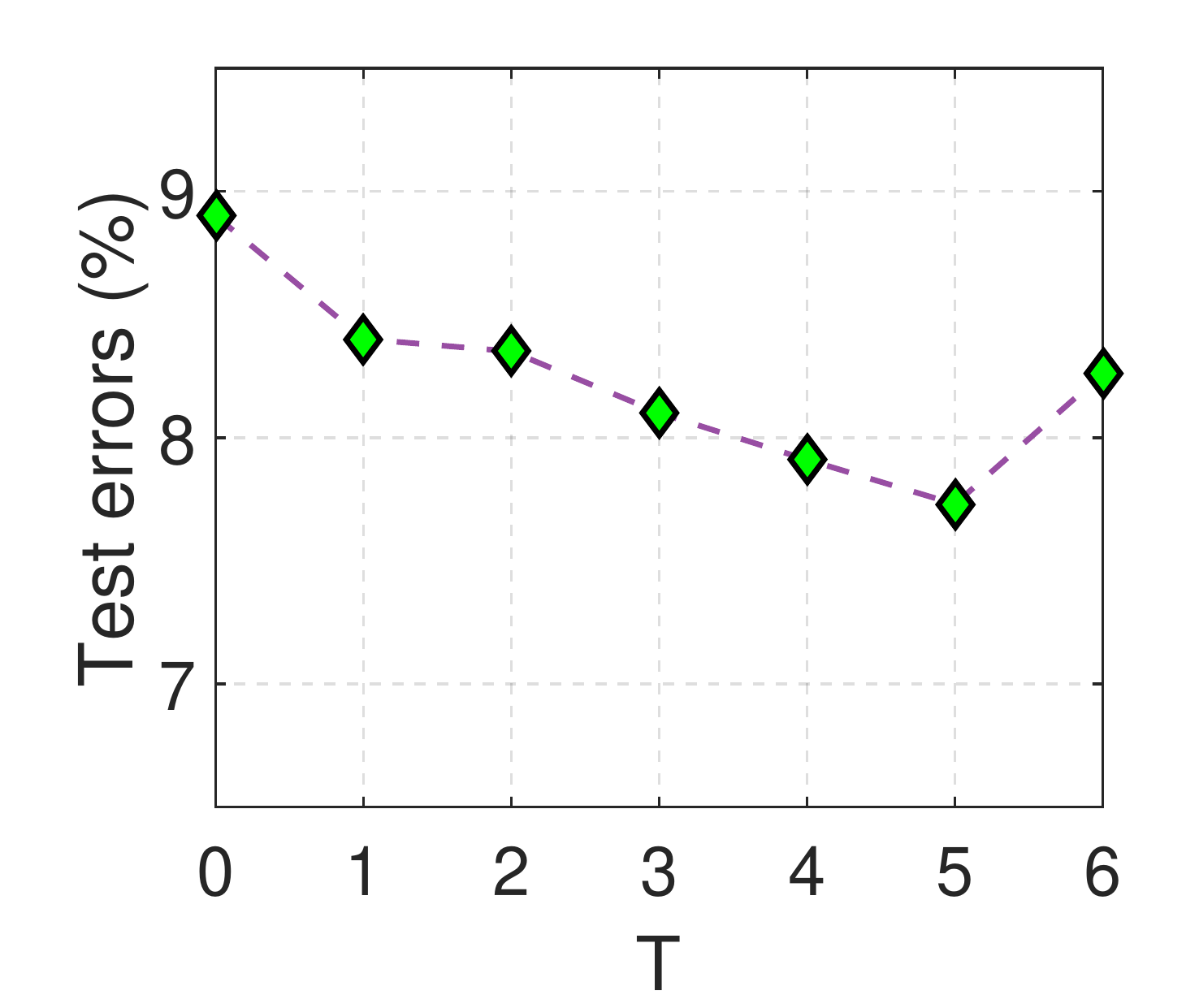}
		\end{minipage}
	}
	\subfigure[]{
		\begin{minipage}[c]{.31\linewidth}
			\centering
			\includegraphics[width=2.8cm]{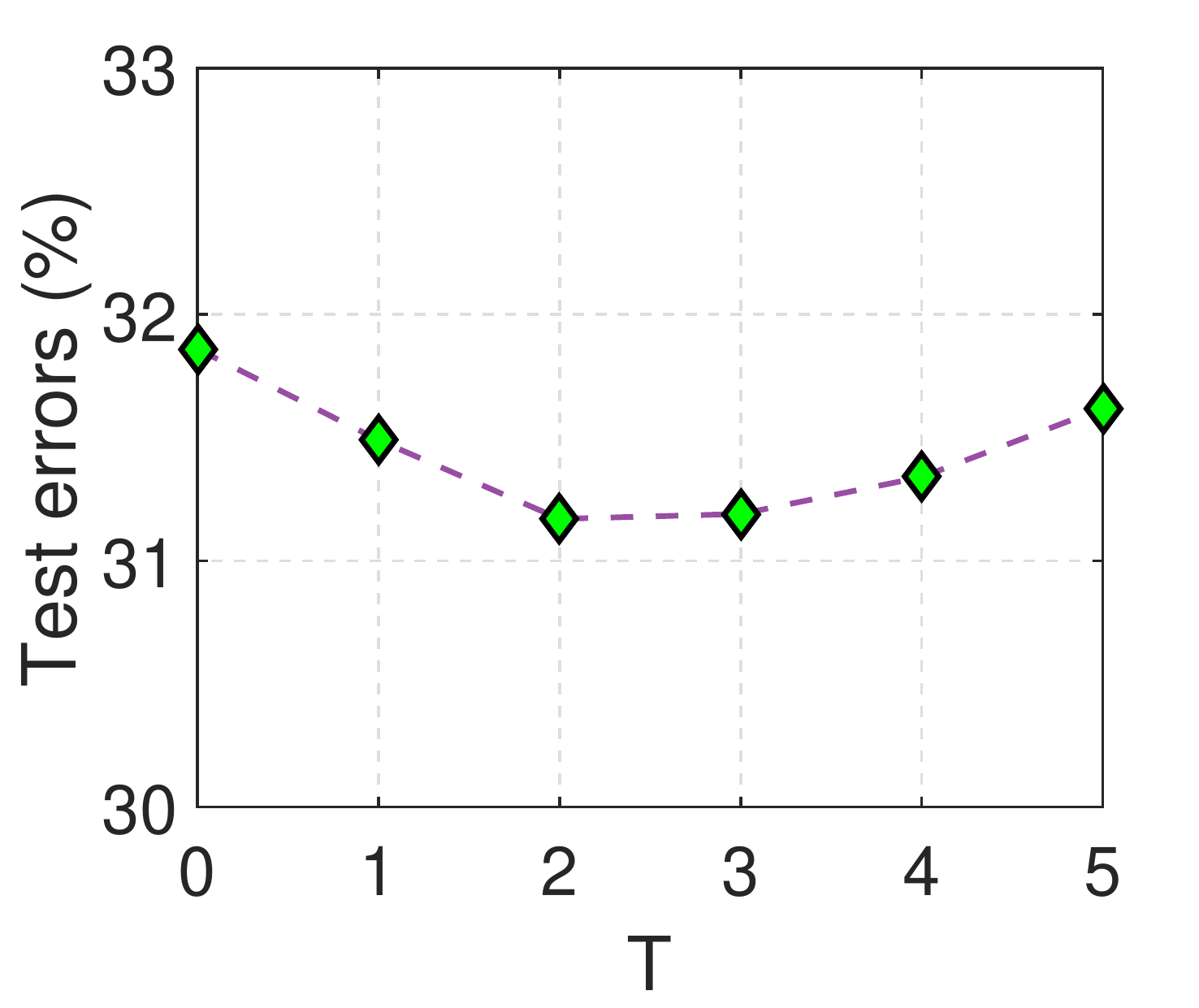}
		\end{minipage}
	}
	\vspace{-0.06in}
	\caption{Effects of the iteration number $T$ of the proposed ONI. (a) 6-layer MLP for Fashion-MNIST; (b) VGG-Style network with $(g=2,k=3)$ for CIFAR-10; (c) 18-layer ResNet for ImageNet. } 
	\label{fig:iteration}
	\vspace{-0.20in}
\end{figure}
\vspace{-0.16in}
\subsubsection{Ablation Study on Iteration Number}
\label{sec:ablation}
ONI controls the spectrum of the weight matrix by the iteration number $T$, as discussed before.  Here, we explore the effect of $T$ on the performance of ONI over different datasets and architectures.
We consider three configurations: 1) the 6-layer MLP for Fashion-MNIST; 2) the VGG-Style network with $(g=2,k=3)$ for CIFAR-10; and 3) the 18-layer ResNet without BN for ImageNet.  The corresponding experimental setups are the same as  described before.
We vary $T$ and show the results in Figure \ref{fig:iteration}. 
Our primary observation is that using either a small or large T degrades performance. This indicates that we need to control the magnitude of orthogonality to balance the increased optimization benefit and  diminished representational capacity. Our empirical observation is that $T=5$ usually works the best for networks without residual connections, whereas $T=2$ usually works better for residual networks. We argue that the residual network itself already has good optimization \cite{2015_CVPR_He}, which reduces the optimization benefits of orthogonality.

Besides, we also observe that larger $T$s have nearly equivalent performance for simple datasets, \eg Fashion-MNIST, as shown in \ref{fig:iteration} (a). This suggests that amplifying the eigenbasis corresponding to a small singular value cannot help more, even though the network with a fully orthogonalized weight matrix can well fit the dataset.  We further show the distributions of the singular values of the orthogonalized weight matrix in \TODO{\SM}~\ref{sup:sec:exp:ablation}.


\begin{figure}[t]
	\centering
		\vspace{-0.16in}
	\subfigure[]{
		\begin{minipage}[c]{.46\linewidth}
			\centering
			\includegraphics[width=3.6cm]{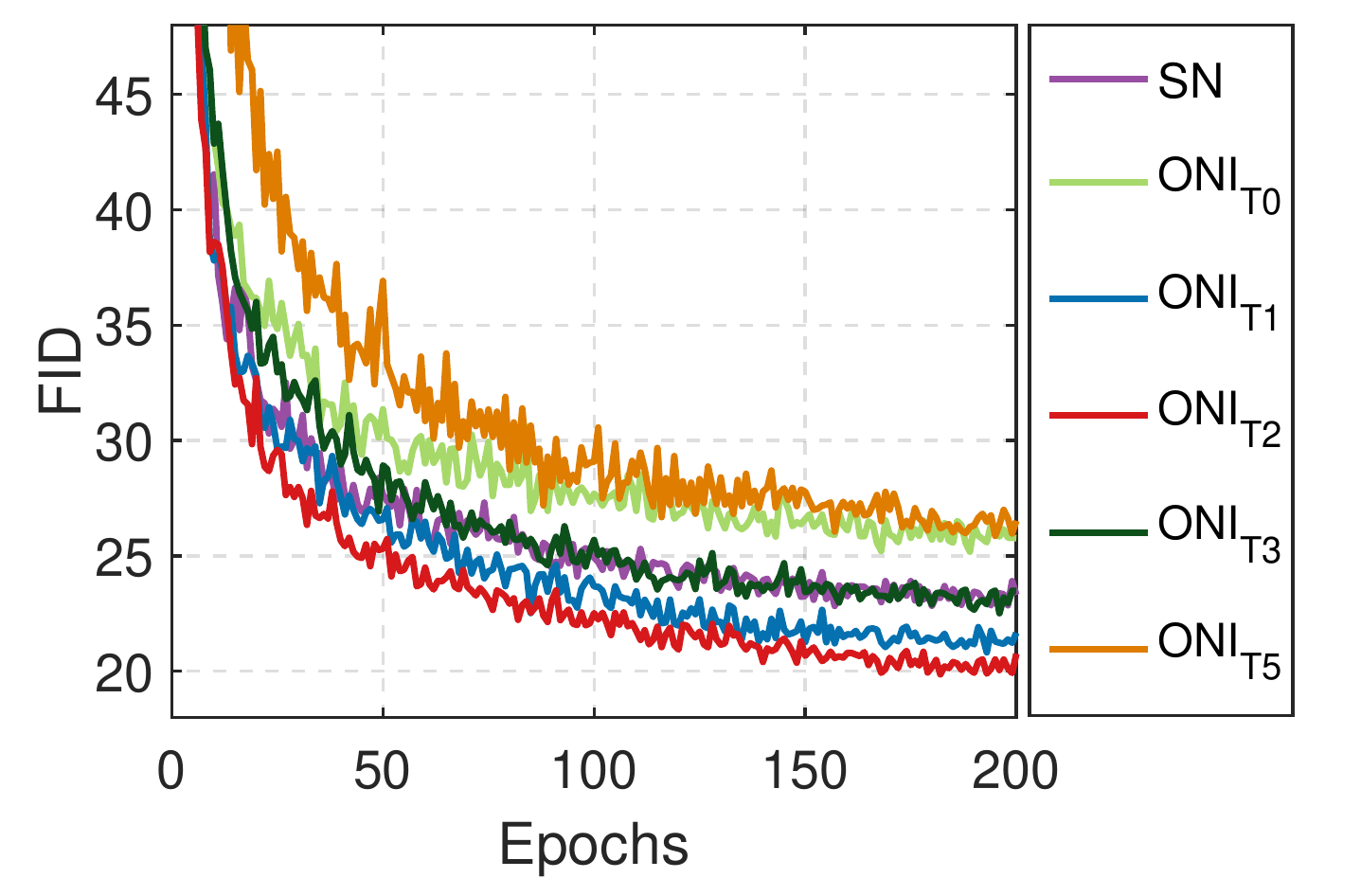}
		\end{minipage}
	}
	\subfigure[]{
		\begin{minipage}[c]{.46\linewidth}
			\centering
			\includegraphics[width=3.6cm]{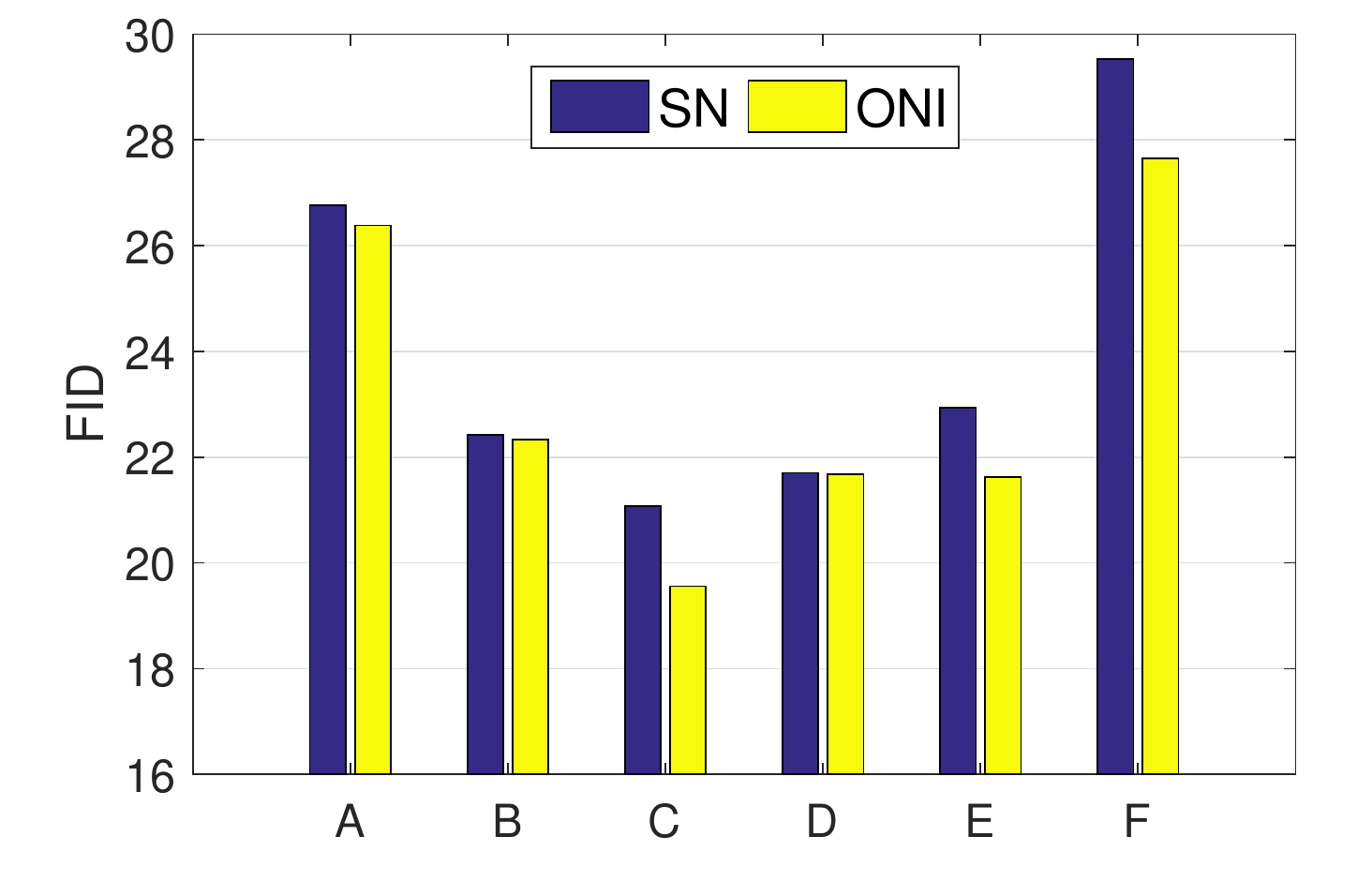}
		\end{minipage}
	}
	\vspace{-0.08in}
	\caption{Comparison of SN and ONI on DCGAN. (a) The FID with respect to training epochs. (b) Stability experiments on six configurations, described in \cite{2018_ICLR_Miyato}.} 
	\label{fig:GAN-dcagan}
		\vspace{-0.2in}
\end{figure}

\vspace{-0.05in}
\subsection{Stabilizing Training of GANs}
\label{sec:GAN}
\vspace{-0.02in}
How to stabilize GAN training is an open research problem \cite{2014_NIPS_goodfellow,2016_NIPS_Salimans,2017_NIPS_Gulrajani}. One pioneering work is spectral normalization (SN) \cite{2018_ICLR_Miyato}, which can maintain the Lipschitz continuity of a network by bounding the maximum eigenvalue of it's weight matrices as 1. This technique has been extensively used in current GAN architectures \cite{2018_ICLR_Miyato_cGAN,2019_ICML_Zhang,2019_ICLR_Brock,2019_ICML_Kurach}.
 As stated before, our method is not only capable of bounding the maximum eigenvalue as 1, but can also control the orthogonality to amplify other eigenbasis with increased iterations, meanwhile orthogonal regularization is also a good technique for training GANs \cite{2019_ICLR_Brock}.
Here, we conduct a series of experiments for unsupervised image generation on CIFAR-10, and compare our method against the widely used SN \cite{2018_ICLR_Miyato}.  
\vspace{-0.18in}
\paragraph{Experimental Setup}
We strictly follow the network architecture
and training protocol reported in the SN paper \cite{2018_ICLR_Miyato}. We use both DCGAN \cite{2016_ICLR_Radford} and ResNet \cite{2015_CVPR_He,2017_NIPS_Gulrajani} architectures. We provide implementation details in \TODO{\SM}~\ref{sup:sec:GAN}. We replace all the SN modules in the corresponding network with our ONI. 
 Our main metric for evaluating the quality of generated samples is the Fr\'{e}chet Inception Distance (FID) \cite{2017_NIPS_Heusel} (the lower the better). We also provide the corresponding Inception Score (IS) \cite{2016_NIPS_Salimans} in  \TODO{\SM}~\ref{sup:sec:GAN}. 
 


 \vspace{-0.18in}
\paragraph{DCGAN}
We use the standard non-saturating function as the adversarial loss \cite{2014_NIPS_goodfellow, 2019_ICML_Kurach} in the DCGAN architecture, following \cite{2018_ICLR_Miyato}. 
For optimization, we use the Adam optimizer \cite{2014_CoRR_Kingma} with the default hyper-parameters, as in \cite{2018_ICLR_Miyato}: learning rate $\alpha = 0.0002$, first momentum $\beta_1 = 0$, second momentum  $\beta_2 = 0.9$, and the number of discriminator updates per generator update $n_{dis}=5$.  We train the network over 200 epochs with a batch size of 64 (nearly 200k generator updates) to determine whether it suffers from training instability. 
 Figure \ref{fig:GAN-dcagan} (a) shows the FID of SN and ONI when varying Newton's iteration number $T$ from 0 to 5.
 One interesting observation is that the ONI with only the initial  spectral bounding described in Formula \ref{eqn:FrobeniusNorm_sqrt} ($T=0$) can also stabilize training, even though it has downgraded performance compared to SN. 
 When $T=1$, ONI achieves better performance than SN. This is because,  based on what we observed,  ONI stretches the maximum eigenvalue to nearly 1, while simultaneously amplifying other eigenvalues. 
Finally, we find that ONI achieves the best performance when $T=2$, yielding an $FID=20.75$, compared to SN's $FID=23.31$. Further increasing $T$ harms the training, possibly because too strong an orthogonalization downgrades the capacity of a network, as discussed in \cite{2018_ICLR_Miyato,2019_ICLR_Brock}. 

We also conduct experiments to validate the stability of our proposed ONI under different experimental configurations: we use six configurations, following \cite{2018_ICLR_Miyato}, by varying $\alpha, \beta_1, \beta_2$ and $n_{dis}$ (denoted by A-F, for more details please see \TODO{\SM})~\ref{sup:sec:DCGAN}. Figure \ref{fig:GAN-dcagan} (b) shows the results of SN and ONI (with T=2) under these six configurations. We observe that our ONI is consistently better than SN. 
 \begin{figure}[t]
 	\centering
 	\vspace{-0.16in}
 	\subfigure[]{
 		\begin{minipage}[c]{.46\linewidth}
 			\centering
 			\includegraphics[width=3.6cm]{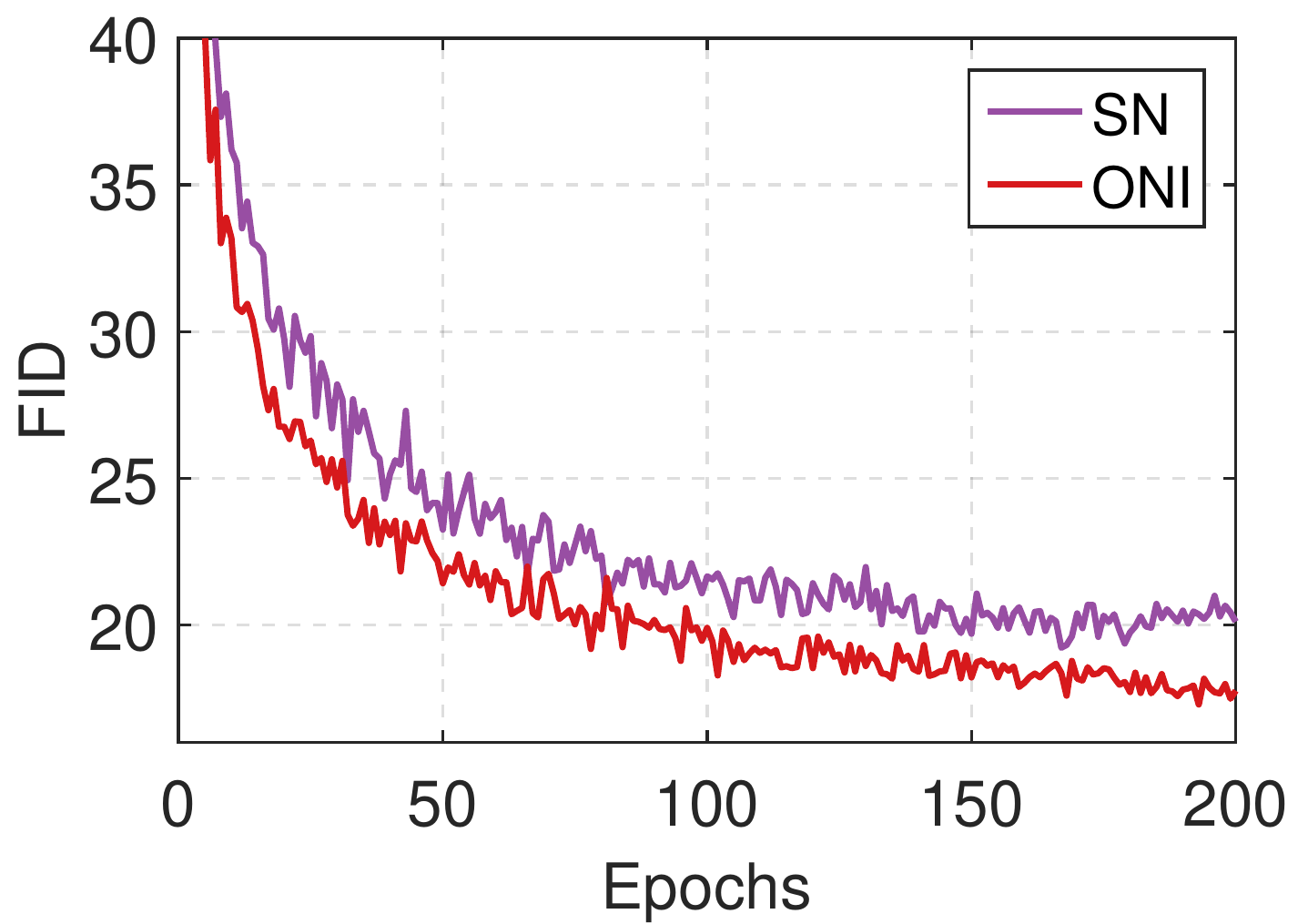}
 		\end{minipage}
 	}
 	\subfigure[]{
 		\begin{minipage}[c]{.46\linewidth}
 			\centering
 			\includegraphics[width=3.6cm]{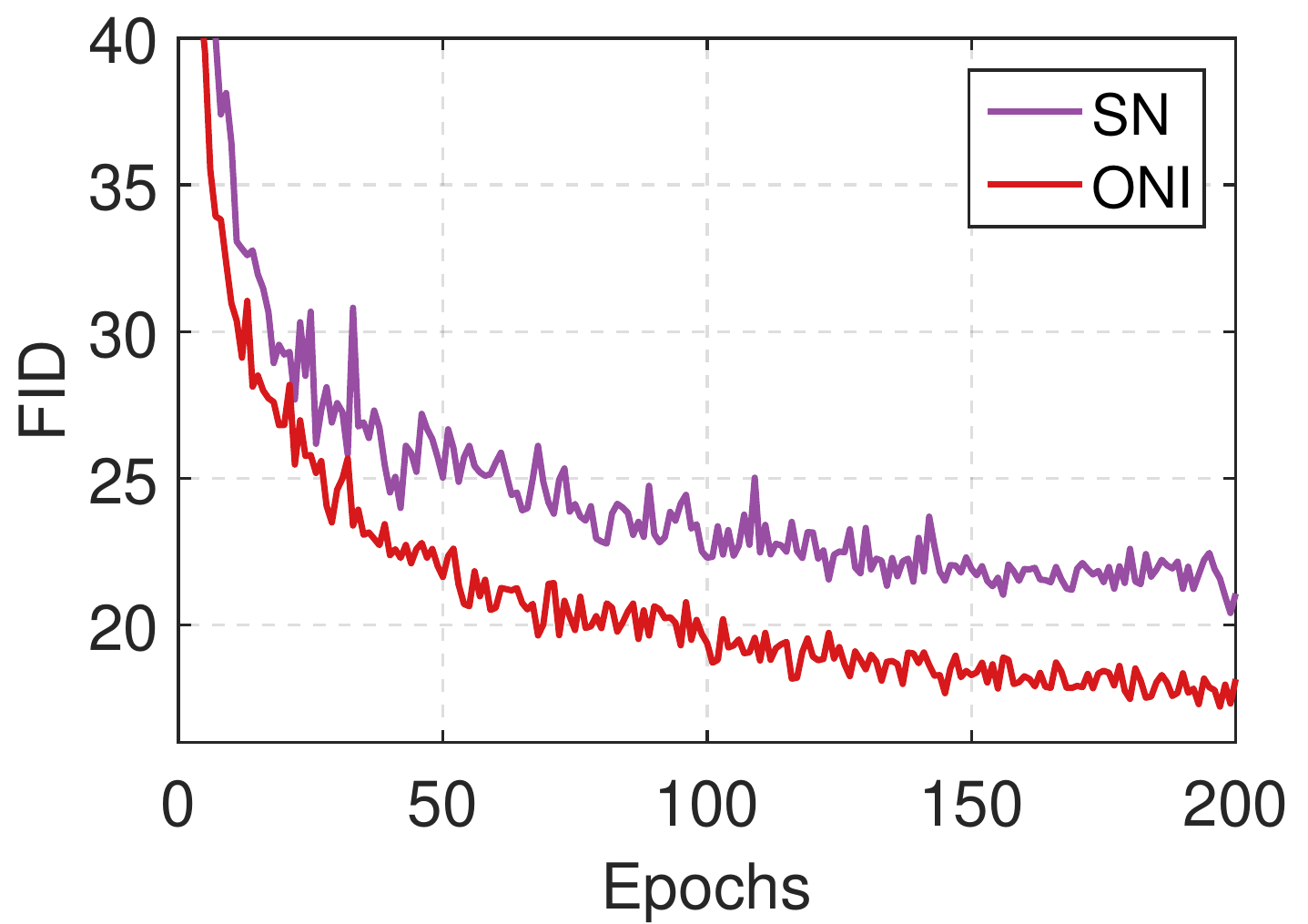}
 		\end{minipage}
 	}
 	\vspace{-0.08in}
 	\caption{Comparison of SN and ONI on ResNet GAN. We show the FID with respect to training epochs when using (a) the non-saturating loss and (b) the hinge loss.} 
 	\label{fig:GAN-resnet}
 	\vspace{-0.2in}
 \end{figure}

%
%

\paragraph{ResNet GAN}
\vspace{-0.18in}
For experiments on the ResNet architecture, we use the same setup as the DCGAN.
Besides the standard non-saturating loss \cite{2014_NIPS_goodfellow}, we also evaluate the recently popularized hinge loss \cite{2017_Corr_GGAN,2018_ICLR_Miyato,2019_ICLR_Brock}. Figure \ref{fig:GAN-resnet} shows the results. We again observe that our ONI achieves better performance than SN under the ResNet architecture, both when using the non-saturating loss and hinge loss.

\vspace{-0.1in}
\section{Conclusion}
\vspace{-0.05in}
In this paper, we proposed an efficient and stable orthogonalization method by  Newton's iteration (ONI) to learn layer-wise  orthogonal weight matrices in DNNs.
We provided insightful analysis for ONI and demonstrated its ability to control orthogonality, which is a desirable property in training DNNs. 
ONI can be implemented as a linear layer and used to learn an orthogonal weight matrix, by simply substituting it for the standard linear module. 

ONI can effectively bound the spectrum of a weight matrix in ($\sigma_{min}$, $\sigma_{max}$) during the course of training. This property makes ONI a potential tool for validating some theoretical results relating to DNN's generalization (\eg, the margin bounds shown in \cite{2017_NIPS_Bartlett}) and resisting attacks from adversarial examples~\cite{2017_ICML_Cisse}. Furthermore, the advantage of ONI in stabilizing training w/o BN (BN usually disturbs the theoretical analysis since it depends on the sampled mini-batch input with stochasticity~\cite{2015_ICML_Ioffe,2019_CVPR_Huang}) makes it possible to validate these theoretical arguments  under real scenarios. 


\vspace{0.1in}
\noindent\textbf{Acknowledgement}
We  thank Anna Hennig and Ying Hu for their help with proofreading.

{\small
	\bibliographystyle{ieee_fullname}
	\bibliography{OWN}

\begin{thebibliography}{10}\itemsep=-1pt

\bibitem{2008_Book_Absil}
P.-A. Absil, R. Mahony, and R. Sepulchre.
\newblock {\em Optimization Algorithms on Matrix Manifolds}.
\newblock Princeton University Press, Princeton, NJ, 2008.

\bibitem{2012_SIAM_Absil}
Pierre-Antoine Absil and Jerome Malick.
\newblock Projection-like retractions on matrix manifolds.
\newblock {\em SIAM Journal on Optimization}, 22(1):135--158, 2012.

\bibitem{2019_Arxiv_Amjad}
Jaweria Amjad, Zhaoyan Lyu, and Miguel~RD Rodrigues.
\newblock Deep learning for inverse problems: Bounds and regularizers.
\newblock {\em arXiv preprint arXiv:1901.11352}, 2019.

\bibitem{2016_ICML_Arjovsky}
Mart{\'{\i}}n Arjovsky, Amar Shah, and Yoshua Bengio.
\newblock Unitary evolution recurrent neural networks.
\newblock In {\em ICML}, 2016.

\bibitem{2018_NIPS_WANG}
Nitin Bansal, Xiaohan Chen, and Zhangyang Wang.
\newblock Can we gain more from orthogonality regularizations in training deep
  cnns?
\newblock In {\em NeurIPS}, 2018.

\bibitem{2017_NIPS_Bartlett}
Peter~L Bartlett, Dylan~J Foster, and Matus~J Telgarsky.
\newblock Spectrally-normalized margin bounds for neural networks.
\newblock In {\em NeurIPS}. 2017.

\bibitem{1994_TNN_Bengio}
Y. Bengio, P. Simard, and P. Frasconi.
\newblock Learning long-term dependencies with gradient descent is difficult.
\newblock {\em Trans. Neur. Netw.}, 5(2):157--166, Mar. 1994.

\bibitem{2005_NumerialAlg}
Dario~A. Bini, Nicholas~J. Higham, and Beatrice Meini.
\newblock Algorithms for the matrix pth root.
\newblock {\em Numerical Algorithms}, 39(4):349--378, Aug 2005.

\bibitem{2018_NIPS_Bjorck}
Nils Bjorck, Carla~P Gomes, Bart Selman, and Kilian~Q Weinberger.
\newblock Understanding batch normalization.
\newblock In {\em NeurIPS}. 2018.

\bibitem{2019_ICLR_Brock}
Andrew Brock, Jeff Donahue, and Karen Simonyan.
\newblock Large scale {GAN} training for high fidelity natural image synthesis.
\newblock In {\em ICLR}, 2019.

\bibitem{2017_ICLR_Brock}
Andrew Brock, Theodore Lim, James~M. Ritchie, and Nick Weston.
\newblock Neural photo editing with introspective adversarial networks.
\newblock In {\em ICLR}, 2017.

\bibitem{2014_cudnn}
Sharan Chetlur, Cliff Woolley, Philippe Vandermersch, Jonathan Cohen, John
  Tran, Bryan Catanzaro, and Evan Shelhamer.
\newblock cudnn: Efficient primitives for deep learning.
\newblock {\em CoRR}, abs/1410.0759, 2014.

\bibitem{2017_ICML_Cisse}
Moustapha Cisse, Piotr Bojanowski, Edouard Grave, Yann Dauphin, and Nicolas
  Usunier.
\newblock Parseval networks: Improving robustness to adversarial examples.
\newblock In {\em ICML}, 2017.

\bibitem{2011_torch}
R. Collobert, K. Kavukcuoglu, and C. Farabet.
\newblock Torch7: A matlab-like environment for machine learning.
\newblock In {\em BigLearn, NIPS Workshop}, 2011.

\bibitem{2016_CoRR_Dorobantu}
Victor Dorobantu, Per~Andre Stromhaug, and Jess Renteria.
\newblock Dizzyrnn: Reparameterizing recurrent neural networks for
  norm-preserving backpropagation.
\newblock {\em CoRR}, abs/1612.04035, 2016.

\bibitem{2010_AISTATS_Glorot}
Xavier Glorot and Yoshua Bengio.
\newblock Understanding the difficulty of training deep feedforward neural
  networks.
\newblock In {\em AISTATS}, 2010.

\bibitem{2014_NIPS_goodfellow}
Ian Goodfellow, Jean Pouget-Abadie, Mehdi Mirza, Bing Xu, David Warde-Farley,
  Sherjil Ozair, Aaron Courville, and Yoshua Bengio.
\newblock Generative adversarial nets.
\newblock In {\em NeurIPS}. 2014.

\bibitem{2017_Arxiv_Priya}
Priya Goyal, Piotr Doll{\'{a}}r, Ross~B. Girshick, Pieter Noordhuis, Lukasz
  Wesolowski, Aapo Kyrola, Andrew Tulloch, Yangqing Jia, and Kaiming He.
\newblock Accurate, large minibatch {SGD:} training imagenet in 1 hour.
\newblock {\em CoRR}, abs/1706.02677, 2017.

\bibitem{2017_NIPS_Gulrajani}
Ishaan Gulrajani, Faruk Ahmed, Martin Arjovsky, Vincent Dumoulin, and Aaron~C
  Courville.
\newblock Improved training of wasserstein gans.
\newblock In {\em NeurIPS}. 2017.

\bibitem{2017_Corr_Harandi}
Mehrtash Harandi and Basura Fernando.
\newblock Generalized backpropagation, etude de cas: Orthogonality.
\newblock {\em CoRR}, abs/1611.05927, 2016.

\bibitem{2015_ICCV_He}
Kaiming He, Xiangyu Zhang, Shaoqing Ren, and Jian Sun.
\newblock Delving deep into rectifiers: Surpassing human-level performance on
  imagenet classification.
\newblock In {\em {ICCV}}, 2015.

\bibitem{2015_CVPR_He}
Kaiming He, Xiangyu Zhang, Shaoqing Ren, and Jian Sun.
\newblock Deep residual learning for image recognition.
\newblock In {\em CVPR}, 2016.

\bibitem{2016_CoRR_He}
Kaiming He, Xiangyu Zhang, Shaoqing Ren, and Jian Sun.
\newblock Identity mappings in deep residual networks.
\newblock In {\em ECCV}, 2016.

\bibitem{2018_ICML_Kyle}
Kyle Helfrich, Devin Willmott, and Qiang Ye.
\newblock Orthogonal recurrent neural networks with scaled {C}ayley transform.
\newblock In {\em ICML}, 2018.

\bibitem{2017_NIPS_Heusel}
Martin Heusel, Hubert Ramsauer, Thomas Unterthiner, Bernhard Nessler, and Sepp
  Hochreiter.
\newblock Gans trained by a two time-scale update rule converge to a local nash
  equilibrium.
\newblock In {\em NeurIPS}. 2017.

\bibitem{2016_CoRR_Huang_a}
Gao Huang, Zhuang Liu, and Kilian~Q. Weinberger.
\newblock Densely connected convolutional networks.
\newblock In {\em CVPR}, 2017.

\bibitem{2018_AAAI_Huang}
Lei Huang, Xianglong Liu, Bo Lang, Adams~Wei Yu, Yongliang Wang, and Bo Li.
\newblock Orthogonal weight normalization: Solution to optimization over
  multiple dependent stiefel manifolds in deep neural networks.
\newblock In {\em AAAI}, 2018.

\bibitem{2018_CVPR_Huang}
Lei Huang, Dawei Yang, Bo Lang, and Jia Deng.
\newblock Decorrelated batch normalization.
\newblock In {\em CVPR}, 2018.

\bibitem{2019_CVPR_Huang}
Lei Huang, Yi Zhou, Fan Zhu, Li Liu, and Ling Shao.
\newblock Iterative normalization: Beyond standardization towards efficient
  whitening.
\newblock In {\em CVPR}, 2019.

\bibitem{2017_AAAI_Hyland}
Stephanie Hyland and Gunnar Rätsch.
\newblock Learning unitary operators with help from u(n).
\newblock In {\em AAAI}, 2017.

\bibitem{2017_NIPS_Ioffe}
Sergey Ioffe.
\newblock Batch renormalization: Towards reducing minibatch dependence in
  batch-normalized models.
\newblock In {\em NeurIPS}, 2017.

\bibitem{2015_ICML_Ioffe}
Sergey Ioffe and Christian Szegedy.
\newblock Batch normalization: Accelerating deep network training by reducing
  internal covariate shift.
\newblock In {\em ICML}, 2015.

\bibitem{2015_ICCV_Ionescu}
Catalin Ionescu, Orestis Vantzos, and Cristian Sminchisescu.
\newblock Training deep networks with structured layers by matrix
  backpropagation.
\newblock In {\em ICCV}, 2015.

\bibitem{2017_CVPR_Jia}
Kui Jia.
\newblock Improving training of deep neural networks via singular value
  bounding.
\newblock In {\em CVPR}, 2017.

\bibitem{2017_GRU_Jing}
Li Jing, {\c{C}}aglar G{\"{u}}l{\c{c}}ehre, John Peurifoy, Yichen Shen, Max
  Tegmark, Marin Soljacic, and Yoshua Bengio.
\newblock Gated orthogonal recurrent units: On learning to forget.
\newblock {\em CoRR}, abs/1706.02761, 2017.

\bibitem{2014_CoRR_Kingma}
Diederik~P. Kingma and Jimmy Ba.
\newblock Adam: {A} method for stochastic optimization.
\newblock In {\em ICLR}, 2015.

\bibitem{2009_TR_Alex}
Alex Krizhevsky.
\newblock Learning multiple layers of features from tiny images.
\newblock Technical report, 2009.

\bibitem{2019_ICML_Kurach}
Karol Kurach, Mario Lu{\v{c}}i{\'c}, Xiaohua Zhai, Marcin Michalski, and
  Sylvain Gelly.
\newblock A large-scale study on regularization and normalization in {GAN}s.
\newblock In {\em ICML}, 2019.

\bibitem{1998_NN_Yann}
Yann LeCun, L{\'e}on Bottou, Genevieve~B. Orr, and Klaus-Robert M\"{u}ller.
\newblock Effiicient backprop.
\newblock In {\em Neural Networks: Tricks of the Trade}, 1998.

\bibitem{2018_CVPR_Lezama}
José Lezama, Qiang Qiu, Pablo Musé, and Guillermo Sapiro.
\newblock OlÉ: Orthogonal low-rank embedding - a plug and play geometric loss
  for deep learning.
\newblock In {\em CVPR}, June 2018.

\bibitem{2018_CVPR_Li}
Peihua Li, Jiangtao Xie, Qilong Wang, and Zilin Gao.
\newblock Towards faster training of global covariance pooling networks by
  iterative matrix square root normalization.
\newblock In {\em CVPR}, 2018.

\bibitem{2017_Corr_GGAN}
Jae~Hyun Lim and Jong~Chul Ye.
\newblock Geometric gan.
\newblock {\em CoRR}, abs/1705.02894, 2017.

\bibitem{2017_BMVC_Lin}
Tsung{-}Yu Lin and Subhransu Maji.
\newblock Improved bilinear pooling with cnns.
\newblock In {\em BMVC}, 2017.

\bibitem{1950_ChemicalPhysics_Lwdin}
Per-Olov L{\"o}wdin.
\newblock On the non-orthogonality problem connected with the use of atomic
  wave functions in the theory of molecules and crystals.
\newblock {\em The Journal of Chemical Physics}, 18(3):365--375, 1950.

\bibitem{2013_ICMLW_Maas}
Andrew~L Maas, Awni~Y Hannun, and Andrew~Y Ng.
\newblock Rectifier nonlinearities improve neural network acoustic models.
\newblock In {\em in ICML Workshop on Deep Learning for Audio, Speech and
  Language Processing}, 2013.

\bibitem{2016_ICLR_Mishkin}
Dmytro Mishkin and Jiri Matas.
\newblock All you need is a good init.
\newblock In {\em ICLR}, 2016.

\bibitem{2018_ICLR_Miyato}
Takeru Miyato, Toshiki Kataoka, Masanori Koyama, and Yuichi Yoshida.
\newblock Spectral normalization for generative adversarial networks.
\newblock In {\em ICLR}, 2018.

\bibitem{2018_ICLR_Miyato_cGAN}
Takeru Miyato and Masanori Koyama.
\newblock cgans with projection discriminator.
\newblock In {\em ICLR}, 2018.

\bibitem{2010_ICML_Nair}
Vinod Nair and Geoffrey~E. Hinton.
\newblock Rectified linear units improve restricted boltzmann machines.
\newblock In {\em {ICML}}, 2010.

\bibitem{2016_Corr_Ozay}
Mete Ozay and Takayuki Okatani.
\newblock Optimization on submanifolds of convolution kernels in cnns.
\newblock {\em CoRR}, abs/1610.07008, 2016.

\bibitem{2013_ICML_Pascanu}
Razvan Pascanu, Tomas Mikolov, and Yoshua Bengio.
\newblock On the difficulty of training recurrent neural networks.
\newblock In {\em ICML}, 2013.

\bibitem{2017_NIPS_pyTorch}
Adam Paszke, Sam Gross, Soumith Chintala, Gregory Chanan, Edward Yang, Zachary
  DeVito, Zeming Lin, Alban Desmaison, Luca Antiga, and Adam Lerer.
\newblock Automatic differentiation in {PyTorch}.
\newblock In {\em NeurIPS Autodiff Workshop}, 2017.

\bibitem{2017_NIPS_Pennington_non}
Jeffrey Pennington, Samuel Schoenholz, and Surya Ganguli.
\newblock Resurrecting the sigmoid in deep learning through dynamical isometry:
  theory and practice.
\newblock In {\em NeurIPS}. 2017.

\bibitem{2016_ICLR_Radford}
Alec Radford, Luke Metz, and Soumith Chintala.
\newblock Unsupervised representation learning with deep convolutional
  generative adversarial networks.
\newblock In {\em ICLR}, 2016.

\bibitem{2015_ImageNet}
Olga Russakovsky, Jia Deng, Hao Su, Jonathan Krause, Sanjeev Satheesh, Sean Ma,
  Zhiheng Huang, Andrej Karpathy, Aditya Khosla, Michael Bernstein,
  Alexander~C. Berg, and Li Fei-Fei.
\newblock {ImageNet Large Scale Visual Recognition Challenge}.
\newblock {\em International Journal of Computer Vision (IJCV)},
  115(3):211--252, 2015.

\bibitem{2016_NIPS_Salimans}
Tim Salimans, Ian Goodfellow, Wojciech Zaremba, Vicki Cheung, Alec Radford, Xi
  Chen, and Xi Chen.
\newblock Improved techniques for training gans.
\newblock In {\em NeurIPS}, 2016.

\bibitem{2016_CoRR_Salimans}
Tim Salimans and Diederik~P. Kingma.
\newblock Weight normalization: {A} simple reparameterization to accelerate
  training of deep neural networks.
\newblock In {\em {NeurIPS}}, 2016.

\bibitem{2013_CoRR_Saxe}
Andrew~M. Saxe, James~L. McClelland, and Surya Ganguli.
\newblock Exact solutions to the nonlinear dynamics of learning in deep linear
  neural networks.
\newblock {\em CoRR}, abs/1312.6120, 2013.

\bibitem{1998_Schraudolph}
Nicol~N. Schraudolph.
\newblock Accelerated gradient descent by factor-centering decomposition.
\newblock Technical report, 1998.

\bibitem{2017_AAAI_Shang}
Wenling Shang, Justin Chiu, and Kihyuk Sohn.
\newblock Exploring normalization in deep residual networks with concatenated
  rectified linear units.
\newblock In {\em AAAI}, 2017.

\bibitem{2014_CoRR_Simonyan}
Karen Simonyan and Andrew Zisserman.
\newblock Very deep convolutional networks for large-scale image recognition.
\newblock In {\em ICLR}, 2015.

\bibitem{2018_Arxiv_Piotr}
Piotr~A. Sokol and Il~Memming Park.
\newblock Information geometry of orthogonal initializations and training.
\newblock {\em CoRR}, abs/1810.03785, 2018.

\bibitem{2016_CoRR_Szegedy}
Christian Szegedy, Sergey Ioffe, and Vincent Vanhoucke.
\newblock Inception-v4, inception-resnet and the impact of residual connections
  on learning.
\newblock {\em CoRR}, abs/1602.07261, 2016.

\bibitem{2014_CoRR_Szegedy}
C. Szegedy, Wei Liu, Yangqing Jia, P. Sermanet, S. Reed, D. Anguelov, D. Erhan,
  V. Vanhoucke, and A. Rabinovich.
\newblock Going deeper with convolutions.
\newblock In {\em CVPR}, 2015.

\bibitem{2018_ICML_Teye}
Mattias Teye, Hossein Azizpour, and Kevin Smith.
\newblock {B}ayesian uncertainty estimation for batch normalized deep networks.
\newblock In {\em ICML}, 2018.

\bibitem{2017_ICML_Eugene}
Eugene Vorontsov, Chiheb Trabelsi, Samuel Kadoury, and Chris Pal.
\newblock On orthogonality and learning recurrent networks with long term
  dependencies.
\newblock In {\em ICML}, 2017.

\bibitem{2016_NIPS_Wisdom}
Scott Wisdom, Thomas Powers, John Hershey, Jonathan Le~Roux, and Les Atlas.
\newblock Full-capacity unitary recurrent neural networks.
\newblock In {\em NeurIPS}. 2016.

\bibitem{2018_ECCV_Wu}
Yuxin Wu and Kaiming He.
\newblock Group normalization.
\newblock In {\em ECCV}, 2018.

\bibitem{2017_FashionMNIST}
Han Xiao, Kashif Rasul, and Roland Vollgraf.
\newblock Fashion-mnist: a novel image dataset for benchmarking machine
  learning algorithms.
\newblock {\em CoRR}, abs/1708.07747, 2017.

\bibitem{2018_ICML_Xiao}
Lechao Xiao, Yasaman Bahri, Jascha Sohl-Dickstein, Samuel S.Schoenholz, and
  Jeffrey Pennington.
\newblock Dynamical isometry and a mean field theory of cnns: How to train
  10,000-layer vanilla convolutional neural networks.
\newblock In {\em ICML}, 2018.

\bibitem{2017_CVPR_Xie}
Di Xie, Jiang Xiong, and Shiliang Pu.
\newblock All you need is beyond a good init: Exploring better solution for
  training extremely deep convolutional neural networks with orthonormality and
  modulation.
\newblock In {\em CVPR}, 2017.

\bibitem{2019_ICLR_Yang}
Greg Yang, Jeffrey Pennington, Vinay Rao, Jascha Sohl{-}Dickstein, and
  Samuel~S. Schoenholz.
\newblock A mean field theory of batch normalization.
\newblock In {\em ICLR}, 2019.

\bibitem{2016_CoRR_Zagoruyko}
Sergey Zagoruyko and Nikos Komodakis.
\newblock Wide residual networks.
\newblock In {\em BMVC}, 2016.

\bibitem{2019_ICLR_Zhang}
Hongyi Zhang, Yann~N. Dauphin, and Tengyu Ma.
\newblock Fixup initialization: Residual learning without normalization.
\newblock In {\em ICLR}, 2019.

\bibitem{2019_ICML_Zhang}
Han Zhang, Ian Goodfellow, Dimitris Metaxas, and Augustus Odena.
\newblock Self-attention generative adversarial networks.
\newblock In {\em ICML}, 2019.

\bibitem{2018_NIPS_Zhang}
Liheng Zhang, Marzieh Edraki, and Guo-Jun Qi.
\newblock Cappronet: Deep feature learning via orthogonal projections onto
  capsule subspaces.
\newblock In {\em NeurIPS}. 2018.

\bibitem{2006_TIP_Zhou}
Jianping Zhou, Minh~N. Do, and Jelena Kovacevic.
\newblock Special paraunitary matrices, cayley transform, and multidimensional
  orthogonal filter banks.
\newblock {\em {IEEE} Trans. Image Processing}, 15(2):511--519, 2006.

\end{thebibliography}
}

\appendix

\clearpage

\renewcommand{\thealgorithm}{\Roman{algorithm}}
\setcounter{algorithm}{0}

\renewcommand{\thetable}{A\arabic{table}}
\setcounter{table}{0}

\renewcommand{\thefigure}{A\arabic{figure}}
\setcounter{figure}{0}

\renewcommand{\theequation}{A\arabic{equation}}
\setcounter{equation}{0}
\linespread{1}
\section{Derivation of Back-Propagation}
\label{sec-sup-backProp}

Given the layer-wise orthogonal weight matrix $\mathbf{W}$, we can perform the forward pass to calculate the loss of the deep neural networks (DNNs).  It's necessary to back-propagate through the orthogonalization transformation,  because we aim to  update the proxy parameters $\mathbf{Z}$.   For illustration, we first describe the proposed orthogonalization by Newton's iteration (ONI) in Algorithm \ref{sup:alg_forward}.  
Given the gradient with respect to the orthogonalized weight matrix $\frac{\partial \mathcal{L}}{\partial \mathbf{W}}$, we target to compute  $\frac{\partial \mathcal{L}}{\partial \mathbf{Z}}$.
The back-propagation is based on the chain rule.
From Line 2 in Algorithm \ref{sup:alg_forward}, we have: 
\begin{small}
	\begin{align}
	\label{sup:eqn:v}
	\frac{\partial \mathcal{L}}{\partial \mathbf{Z}}&= \frac{1}{\|\mathbf{Z}\|_F}  \frac{\partial \mathcal{L}}{\partial \mathbf{V}} +
	tr(\frac{\partial \mathcal{L}}{\partial \mathbf{V}}^T \mathbf{Z}) \frac{
		\partial \frac{1}{\|\mathbf{Z}\|_F}	} { \partial \| \mathbf{Z} \|_F}  \frac{\partial \| \mathbf{Z} \|_F}{\partial  \mathbf{Z}}   \nonumber \\
	&= \frac{1}{\|\mathbf{Z}\|_F}  \frac{\partial \mathcal{L}}{\partial \mathbf{V}} -
	tr(\frac{\partial \mathcal{L}}{\partial \mathbf{V}}^T \mathbf{Z}) \frac{1}{\|\mathbf{Z}\|_F^2} \frac{\mathbf{Z}}{\|\mathbf{Z}\|_F}   \nonumber \\
	&=  \frac{1}{\| \mathbf{Z} \|_F} (\frac{\partial \mathcal{L}}{\partial \mathbf{V}} - \frac{tr(\frac{\partial \mathcal{L}}{\partial \mathbf{V}}^T \mathbf{Z})}{\| \mathbf{Z} \|_F^2} 
	\mathbf{Z}),                                              
	\end{align}
\end{small}
\hspace{-0.05in}where $tr(\cdot)$ indicates the trace of the corresponding matrix and  $\frac{\partial \mathcal{L}}{\partial \mathbf{V}}$ can be calculated from Lines 3 and 8 in Algorithm \ref{sup:alg_forward}:
\begin{eqnarray}
\label{eqn:v}
\frac{\partial \mathcal{L}}{\partial \mathbf{V}}=  (\mathbf{B}_{T})^T  \frac{\partial \mathcal{L}}{\partial \mathbf{W}} 
+  (\frac{\partial \mathcal{L}}{\partial \mathbf{S}} +  \frac{\partial \mathcal{L}}{\partial \mathbf{S}}^T ) \mathbf{V}.  
\end{eqnarray}

We thus need to calculate $\frac{\partial \mathcal{L}}{\partial \mathbf{S}}$, which can be computed from Lines 5, 6 and 7 in Algorithm \ref{sup:alg_forward}:
\begin{eqnarray}
\frac{\partial{L}}{\partial{\mathbf{S}}}= -\frac{1}{2} \sum_{t=1}^{T} (\mathbf{B}_{t-1}^3)^T \frac{\partial{L}}{\partial{\mathbf{B}_t}},
\end{eqnarray}
where $\frac{\partial \mathcal{L}}{\partial \mathbf{B}_{T}} = \frac{\partial \mathcal{L}}{\partial \mathbf{W}} \mathbf{V}^T$ and $\{ \frac{\partial \mathcal{L}}{\partial \mathbf{B}_{t-1}}, t=T,...,1 \}$ can be iteratively calculated from Line 6 in Algorithm \ref{sup:alg_forward} as follows:
\begin{small}
	\begin{align}
	\frac{\partial \mathcal{L}}{\partial \mathbf{B}_{t-1} }&=- \frac{1}{2}( \frac{\partial \mathcal{L}}{\partial \mathbf{B}_{t} }(\mathbf{B}_{t-1}^2 \mathbf{S})^T +  (\mathbf{B}_{t-1}^{2})^T   \frac{\partial \mathcal{L}}{\partial \mathbf{B}_{t} } \mathbf{S}^T \nonumber \\
	&\phantom{{}={}} +  \mathbf{B}_{t-1}^{T} \frac{\partial \mathcal{L}}{\partial \mathbf{B}_{t} }( \mathbf{B}_{t-1} \mathbf{S})^{T} )
	+ \frac{3}{2}  \frac{\partial \mathcal{L}}{\partial \mathbf{B}_{t} }.
	\end{align}
\end{small}
\hspace{-0.05in}In summary, the back-propagation of Algorithm \ref{sup:alg_forward} is shown in Algorithm \ref{alg_backward}.

\begin{small}
	\begin{algorithm}[b]
		\caption{Orthogonalization by Newton's Iteration.}
		\label{sup:alg_forward}
		\begin{algorithmic}[1]
			\STATE \textbf{Input}: proxy parameters $\mathbf{Z} \in \mathbb{R}^{n \times d} $ and iteration numbers $T$.
			\STATE  Bounding $\mathbf{Z}$'s singular values:  $\mathbf{V}= \frac{\mathbf{Z}}{\| \mathbf{Z} \|_F}$.
			\STATE  Calculate covariance matrix: $\mathbf{S}= \mathbf{V} \mathbf{V}^T$.
			\STATE $\mathbf{B}_0=\mathbf{I}$.
			\FOR {$t = 1$ to T}
			\STATE $\mathbf{B}_{t}=\frac{3}{2} \mathbf{B}_{t-1}-\frac{1}{2}\mathbf{B}_{t-1}^3  \mathbf{S} $.
			\ENDFOR		
			\STATE $\mathbf{W}=\mathbf{B}_{T} \mathbf{V}$.
			\STATE \textbf{Output}:  orthogonalized weight matrix: $\mathbf{W} \in \mathbb{R}^{n \times d}$.
		\end{algorithmic}
	\end{algorithm}
\end{small}

\begin{small}
	\begin{algorithm}[b]
		\caption{Back-propagation of ONI.}
		\label{alg_backward}
		\begin{algorithmic}[1]
			\STATE \textbf{Input}:
			$\frac{\partial \mathcal{L}}{\partial \mathbf{W}} \in \mathbb{R}^{n \times d} $ and variables from respective forward pass: $\mathbf{Z}$, $\mathbf{V}$, $\mathbf{S}$, $ \{\mathbf{B}_t \}_{t=1}^{T}$.
			\STATE $\frac{\partial \mathcal{L}}{\partial \mathbf{B}_{T}} = \frac{\partial \mathcal{L}}{\partial \mathbf{W}} \mathbf{V}^T$.
			\FOR {$t = T$ down to 1}
			\STATE $\frac{\partial \mathcal{L}}{\partial \mathbf{B}_{t-1} }=- \frac{1}{2}( \frac{\partial \mathcal{L}}{\partial \mathbf{B}_{t} }(\mathbf{B}_{t-1}^2 \mathbf{S})^T +  (\mathbf{B}_{t-1}^{2})^T   \frac{\partial \mathcal{L}}{\partial \mathbf{B}_{t} } \mathbf{S}^T 
			+  \mathbf{B}_{t-1}^{T} \frac{\partial \mathcal{L}}{\partial \mathbf{B}_{t} }( \mathbf{B}_{t-1} \mathbf{S})^{T} )
			+ \frac{3}{2}  \frac{\partial \mathcal{L}}{\partial \mathbf{B}_{t} }.
			$
			\ENDFOR	
			\STATE $\frac{\partial{L}}{\partial{\mathbf{S}}}= -\frac{1}{2} \sum_{t=1}^{T} (\mathbf{B}_{t-1}^3)^T \frac{\partial{L}}{\partial{\mathbf{B}_t}}$.  	
			\STATE $ \frac{\partial \mathcal{L}}{\partial \mathbf{V}}=  (\mathbf{B}_{T})^T  \frac{\partial \mathcal{L}}{\partial \mathbf{W}} 
			+  (\frac{\partial \mathcal{L}}{\partial \mathbf{S}} +  \frac{\partial \mathcal{L}}{\partial \mathbf{S}}^T ) \mathbf{V}   $.
			\STATE $\frac{\partial \mathcal{L}}{\partial \mathbf{Z}}= \frac{1}{\| \mathbf{Z} \|_F} (\frac{\partial \mathcal{L}}{\partial \mathbf{V}} - \frac{tr(\frac{\partial \mathcal{L}}{\partial \mathbf{V}}^T \mathbf{Z})}{\| \mathbf{Z} \|_F^2} 
			\mathbf{Z}) $.	
			\STATE \textbf{Output}: $  \frac{\partial \mathcal{L}}{\partial \mathbf{Z} } \in \mathbb{R}^{n \times d}$.
		\end{algorithmic}
	\end{algorithm}
\end{small}

We further derive the back-propagation of the accelerated ONI method with the centering and more compact spectral bounding operation, as described in Section~\ref{sec-speedONI} of the  paper. For illustration, Algorithm \ref{sup:alg_forward_acc} describes the forward pass of the accelerated ONI. Following the calculation in Algorithm \ref{alg_backward}, we can obtain  $\frac{\partial \mathcal{L}}{\partial \mathbf{V}}$.  
To simplify the derivation, we represent Line 3 of Algorithm \ref{sup:alg_forward_acc} as the following formulations:
\begin{small}
	\begin{align}
	\label{eqn:caculateM}
	\mathbf{M} =& \mathbf{Z}_c \mathbf{Z}_c^T  \\
	\label{eqn:caculateDelta}
	\delta=& \sqrt{ \| \mathbf{M} \|_F }   \\
	\label{eqn:caculateV}
	\mathbf{V} = &\frac{\mathbf{Z}_c}{\delta}.
	\end{align}
\end{small}

It's easy to calculate $\frac{\partial \mathcal{L}}{\partial \mathbf{Z}_c}$ from Eqn.~\ref{eqn:caculateM} and Eqn.~\ref{eqn:caculateV} as follows:
\begin{small}
	\begin{align}
	\label{eqn:caculateDZC}
	\frac{\partial \mathcal{L}}{\partial \mathbf{Z}_c}= \frac{1}{\delta}  \frac{\partial \mathcal{L}}{\partial \mathbf{V}} + (\frac{\partial \mathcal{L}}{\partial \mathbf{M}}  +  \frac{\partial \mathcal{L}}{\partial \mathbf{M}}^T) \mathbf{Z}_c,  
	\end{align}
\end{small}
where $\frac{\partial \mathcal{L}}{\partial \mathbf{M}}$ can be computed based on Eqn.~\ref{eqn:caculateDelta} and Eqn.~\ref{eqn:caculateV}:
\begin{small}
	\begin{align}
	\label{eqn:calculateDM}
	\frac{\partial \mathcal{L}}{\partial \mathbf{M}} &= \frac{\partial \mathcal{L}}{\partial \delta} \frac{\partial \delta} {\partial \| \mathbf{M}  \|_F}  \frac{\partial \| \mathbf{M} \|_F }{ \partial \mathbf{M}} \nonumber \\
	&= tr(\frac {\partial \mathcal{L}}{\partial \mathbf{V}}^T  \mathbf{Z}_c ) (-\frac{1} {\delta^2})    
	\frac{1} {2 \sqrt{\| \mathbf{M} \|_F}}  \frac{\mathbf{M}}{\| \mathbf{M} \|_F} \nonumber \\
	&= -\frac{tr(\frac {\partial \mathcal{L}}{\partial \mathbf{V}}^T  \mathbf{Z}_c )}{2 \delta^5}  \mathbf{M}.  
	\end{align}
\end{small}

Based on Line 2 in Algorithm \ref{sup:alg_forward_acc}, we can achieve $\frac{\partial \mathcal{L}}{\partial \mathbf{Z}}$ as follows:
\begin{small}
	\begin{align}
	\label{eqn:caculateDZ}
	\frac{\partial \mathcal{L}}{\partial \mathbf{Z}}=  \frac{\partial \mathcal{L}}{\partial \mathbf{Z}}_c -\frac{1}{d} \mathbf{1} \mathbf{1}^T  \frac{\partial \mathcal{L}}{\partial \mathbf{Z}}_c. 
	\end{align}
\end{small}

In summary, Algorithm \ref{alg_forward_back} describes the back-propagation of the Algorithm \ref{sup:alg_forward_acc}.

\begin{small}
	\begin{algorithm}[t]
		\caption{ ONI with acceleration.}
		\label{sup:alg_forward_acc}
		\begin{algorithmic}[1]
			\STATE \textbf{Input}: proxy parameters $\mathbf{Z} \in \mathbb{R}^{n \times d} $ and iteration numbers $N$.
			\STATE  Centering: $\mathbf{Z}_c= \mathbf{Z} - \frac{1}{d}\mathbf{Z}\mathbf{1} \mathbf{1}^T$.
			\STATE  Bounding  $\mathbf{Z}$'s singular values: $\mathbf{V}=\frac{\mathbf{Z}_c}{\sqrt{\| \mathbf{Z}_c \mathbf{Z}_c^T \|_F}}$.
			\STATE  Execute Step. 3 to 8 in Algorithm \ref{sup:alg_forward}.
			\STATE \textbf{Output}:  orthogonalized weight matrix: $\mathbf{W} \in \mathbb{R}^{n \times d}$.
		\end{algorithmic}
	\end{algorithm}
\end{small}

\section{Proof of Convergence Condition for Newton's Iteration}
\label{sup:proofConverge}

In Section~\ref{sec:solve_optimization} of the paper, we show that bounding the spectral of the proxy parameters matrix by 
\begin{equation}
\label{sup:eqn:FrobeniusNorm}
\mathbf{V}=\phi_N(\mathbf{Z}) =\frac{\mathbf{Z}}{\| \mathbf{Z} \|_F}
\end{equation}
and 
\begin{equation}
\label{sup:eqn:FrobeniusNorm_sqrt}
\mathbf{V}=\phi_N(\mathbf{Z}) =\frac{\mathbf{Z}}{\sqrt{\| \mathbf{Z} \mathbf{Z}^T \|_F}}
\end{equation}
can satisfy the convergence condition of Newton's Iterations as follows:
\begin{equation}
\label{eqn:converge}
\|\mathbf{I} -\mathbf{S} \|_2 <1,
\end{equation}
where $\mathbf{S}=\mathbf{V} \mathbf{V}^T$ and the singular values  of $\mathbf{Z}$  are nonzero. Here we will prove this conclusion, and we also prove that $\| \mathbf{Z} \|_F  > \sqrt{\| \mathbf{Z} \mathbf{Z}^T \|_F} $.

\begin{proof}
	
	By definition, $\| \mathbf{Z} \|_F$ can be represented as $\| \mathbf{Z} \|_F = \sqrt{tr(\mathbf{Z} \mathbf{Z}^T)}$. Given Eqn.~\ref{sup:eqn:FrobeniusNorm}, we calculate
	\begin{equation}
	\label{eqn:Equivalent}
	\mathbf{S} =\mathbf{V} \mathbf{V}^T= \frac{\mathbf{Z} \mathbf{Z}^T}{tr(\mathbf{Z} \mathbf{Z}^T)}.
	\end{equation}
	Let's denote $\mathbf{M}= \mathbf{Z} \mathbf{Z}^T$ and the  eigenvalues of  $\mathbf{M}$ are $\{\lambda_1, ..., \lambda_n \}$. We  have $\lambda_i > 0 $, since $\mathbf{M}$ is a real symmetric matrix and the singular values of $\mathbf{Z}$ are nonzero. 
	We also have $\mathbf{S}=\frac{\mathbf{M}}{tr(\mathbf{M})}$ and the eigenvalues of $\mathbf{S}$ are $\frac{\lambda_i}{\sum_{i=1}^n \lambda_i}$. 
	Furthermore,  the  eigenvalues of $(\mathbf{I}-\mathbf{S})$ are $1- \frac{\lambda_i}{\sum_{i=1}^n \lambda_i}$, thus satisfying the convergence condition described by Eqn.~\ref{eqn:converge}.

	Similarly, given $\mathbf{V}=\phi_N(\mathbf{Z}) =\frac{\mathbf{Z}}{ \sqrt{\| \mathbf{Z} \mathbf{Z}^T \|_F}}$, we have $\mathbf{S} = \frac{\mathbf{Z} \mathbf{Z}^T}{\| \mathbf{Z} \mathbf{Z}^T \|_F} =\frac{\mathbf{M}}{\|\mathbf{M} \|_F}$ and its corresponding eigenvalues are $\frac{\lambda_i}{\sqrt{\sum_{i=1}^n \lambda_i^2}}$.	Therefore,  the singular values of $(\mathbf{I}-\mathbf{S})$  are $1- \frac{\lambda_i}{\sqrt{\sum_{i=1}^n \lambda_i^2}}$, also satisfying the convergence condition described by Eqn.~\ref{eqn:converge}. 
	
	We have $\| \mathbf{Z} \|_F=\sqrt{tr(\mathbf{M})}= \sqrt{\sum_{i=1}^n \lambda_i}$  and $\sqrt{\| \mathbf{Z} \mathbf{Z}^T \|_F}=\sqrt{\|\mathbf{M}\|_F}=\sqrt[4]{\sum_{i=1}^n \lambda_i^2} $. 
	It's easy to demonstrate that $\| \mathbf{Z} \|_F  > \sqrt{\| \mathbf{Z} \mathbf{Z}^T \|_F} $, since $(\sum_{i=1}^n \lambda_i)^2 > \sum_{i=1}^n \lambda_i^2$.
\end{proof}

\begin{small}
	\begin{algorithm}[t]
		\caption{Back-propagation of ONI with acceleration.}
		\label{alg_forward_back}
		\begin{algorithmic}[1]
			\STATE \textbf{Input}:
			$\frac{\partial \mathcal{L}}{\partial \mathbf{W}} \in \mathbb{R}^{n \times d} $ and variables from respective forward pass: $\mathbf{Z}_c$, $\mathbf{V}$, $\mathbf{S}$, $ \{\mathbf{B}_t \}_{t=1}^{T}$.
			\STATE  Calculate $\frac{\partial \mathcal{L}}{\partial \mathbf{V}}$ from Line 2 to Line 7 in Algorithm \ref{alg_backward}.
			\STATE  Calculate $\mathbf{M} $ and $\delta$ from Eqn. \ref{eqn:caculateM} and Eqn. \ref{eqn:caculateDelta}.
			\STATE  Calculate $\frac{\partial \mathcal{L}}{\partial \mathbf{M} }$ based on Eqn. \ref{eqn:calculateDM}.
			\STATE  Calculate $\frac{\partial \mathcal{L}}{\partial \mathbf{Z}_c }$ based on Eqn. \ref{eqn:caculateDZC}.
			\STATE  Calculate $\frac{\partial \mathcal{L}}{\partial \mathbf{Z} }$ based on Eqn. \ref{eqn:caculateDZ}.
			\STATE \textbf{Output}: $  \frac{\partial \mathcal{L}}{\partial \mathbf{Z} } \in \mathbb{R}^{n \times d}$.
		\end{algorithmic}
	\end{algorithm}
\end{small}

In Section~\ref{sec:solve_optimization} of the paper, we show that the Newton's iteration by bounding the spectrum with Eqn.~\ref{sup:eqn:FrobeniusNorm} is equivalent to the Newton's iteration proposed in \cite{2019_CVPR_Huang}. Here, we provide the details. In \cite{2019_CVPR_Huang}, they bound the covariance matrix $\mathbf{M}=\mathbf{Z} \mathbf{Z}^T$ by the trace of $\mathbf{M}$ as $\frac{\mathbf{M}}{tr(\mathbf{M})}$. It's clear that $\mathbf{S}$ used in Algorithm~\ref{sup:alg_forward} is equal to $\frac{\mathbf{M}}{tr(\mathbf{M})}$, based on Eqn.~\ref{eqn:Equivalent} shown in the previous proof.

\begin{table*}[t]	
	\centering
	\begin{small}
		\begin{tabular}{l|ccccc}
			\toprule[1pt]
			configurations	& cudnn   &  cudnn~+~ONI-T1   &   cudnn~+~ONI-T3   &   cudnn~+~ONI-T5  &  cudnn~+~ONI-T7     \\
			\hline
			$F_h=F_w=3$,~ n=d=256,~ m=256 	& 118.6   &  122.1   & 122.9  & 124.4  &   125.7     \\
			$F_h=F_w=3$,~ n=d=256,~ m=32 	& 15.8   &  18.3   & 18.9  & 19.5  &   20.8     \\
			$F_h=F_w=3$,~ n=d=1024,~ m=32 	& 71.1   &  81.7   & 84.3  & 89.5  &   94.2     \\
			$F_h=F_w=1$,~ n=d=256,~ m=256 	& 28.7   &  31.5   & 32.1  & 33.7  &   34.6     \\
			$F_h=F_w=1$,~ n=d=256,~ m=32 	& 10.1   &  13   & 13.6  & 14.2  &   15.3     \\
			$F_h=F_w=1$,~ n=d=1024,~ m=32 	& 22.2   &  27.6   & 29.7  & 32.9  &   37.0     \\		
			\toprule[1pt]
		\end{tabular}
	\end{small}
	\vspace{0.05in}
	\caption{Comparison of wall-clock time ($ms$). We fix the input with size $h=w=32$. We evaluate the total wall-clock time of training for each iteration (forward pass + back-propagation pass). Note that `cudnn ~+~ ONI-T5' indicates the `cudnn' convolution wrapped in our ONI method, using an iteration number of 5.
	}
	\label{tab:timeCost}
\end{table*}

\begin{table}[t]
	
	\centering
	\begin{small}
		\begin{tabular}{l|c|c}
			\toprule[1pt]
			& $\delta_{Row} $   & $\delta_{Column}$ \\ 
			\hline
			ONI-Full   & 5.66 & 0 \\
			OLM-G32  & 8 &  5.66\\
			OLM-G16    & 9.85  &  8.07\\ 
			OLM-G8    & 10.58  &  8.94\\ 
			\toprule[1pt]
		\end{tabular}
	\end{small}
	\vspace{0.05in}
	\caption{Evaluation for row and column orthogonalization with the group based methods. The entries of proxy matrix $\mathbf{Z} \in \mathbb{R}^{32 \times 64}$ are sampled from the Gaussian distribution $N(0,1)$. We evaluate the row orthogonality  $\delta_{Row}=\| \mathbf{W} \mathbf{W}^T - \mathbf{I} \|_F$ and column orthogonality $\delta_{Column}=\| \mathbf{W}^T \mathbf{W} - \mathbf{I} \|_F$. `OLM-G32' indicates the eigen decomposition based orthogonalization method described in \cite{2018_AAAI_Huang},  with a group size of 32.
	}
	\label{tab:GroupOrth}
\end{table}

\section{Orthogonality for Group Based Method}
\label{sup:sec-Group}
In Section~\ref{sec:RowAndColumn} of the paper, we argue that group based methods cannot ensure the whole matrix $\mathbf{W}\in \mathbb{R}^{n \times d}$ to be either row or column orthogonal, when $n>d$. Here we provide more details. 

We follow the experiments described in Figure~\ref{fig:NI_accelerate} of the paper, where we sample the entries of proxy matrix $\mathbf{Z} \in \mathbb{R}^{64 \times 32}$ from the Gaussian distribution $N(0,1)$. We apply the eigen decomposition based orthogonalization method \cite{2018_AAAI_Huang} with group size $G$,  to obtain the orthogonalized matrix  $\mathbf{W}$. We vary the group size $G \in \{ 32, 16, 8\}$. We evaluate the corresponding row orthogonality $\delta_{Row}=\| \mathbf{W} \mathbf{W}^T - \mathbf{I} \|_F$ and   column orthogonality  $\delta_{Column}=\| \mathbf{W}^T \mathbf{W} - \mathbf{I} \|_F$. The results are shown in Table \ref{tab:GroupOrth}. We observe that the group based orthogonalization method cannot ensure the whole matrix $\mathbf{W}$ to be either row or column orthogonal, while our ONI can ensure column orthogonality. We also observe that the group based method has degenerated orthogonality, with decreasing group size. 

\begin{figure}[t]
	\centering
	\begin{minipage}[c]{.7\linewidth}
		\centering
		\includegraphics[width=5.6cm]{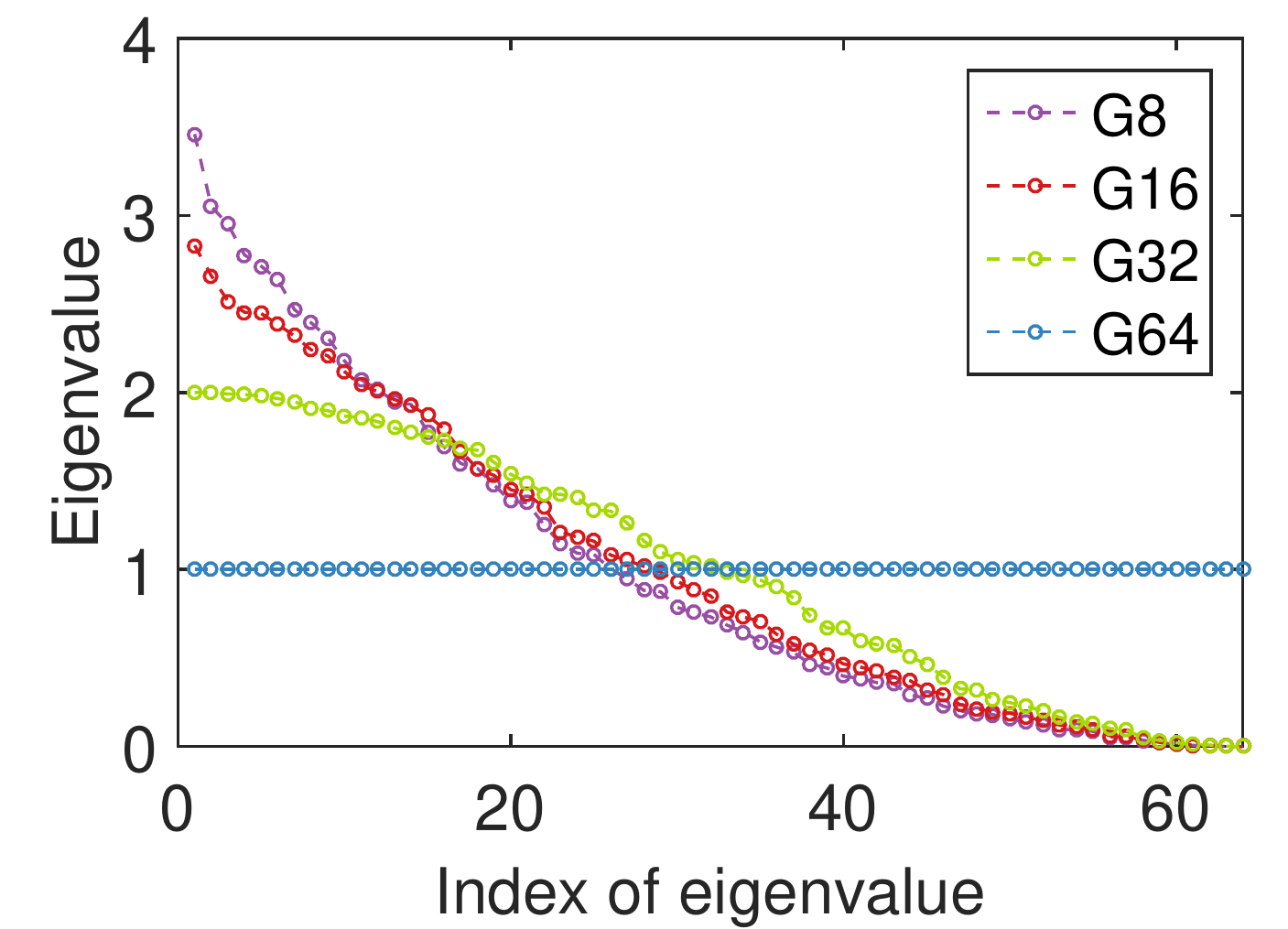}
	\end{minipage}
	\vspace{0.2in}
	\caption{The distribution of the eigenvalues of $\mathbf{W} \mathbf{W}^T$ with different group size $G$. The entries of proxy matrix $\mathbf{Z} \in \mathbb{R}^{64 \times 64}$ are sampled from the Gaussian distribution $N(0,1)$.}
	\label{fig:GroupSVD}
\end{figure}

We also conduct an experiment when $n=d$, where we sample the entries of proxy matrix $\mathbf{Z} \in \mathbb{R}^{64 \times 64}$ from the Gaussian distribution $N(0,1)$.  We vary the group size $G \in \{64, 32, 16, 8\}$. Note that $G=64$ represents full orthogonalization. Figure \ref{fig:GroupSVD} shows the distribution of the eigenvalues of $\mathbf{W} \mathbf{W}^T$. We again observe that the group based method cannot ensure the whole weight matrix to be row orthogonal. 
Furthermore, orthogonalization with smaller group size tends to be worse.

\section{Comparison of Wall Clock Times }
\label{sup:sec:time}

In Section~\ref{sec:analysis} of the paper, we show that, given a convolutional layer with filters $\mathbf{W} \in \mathbb{R}^{n \times d \times F_h \times F_w}$ and $m$ mini-batch data $\{\mathbf{x}_i \in \mathbb{R}^{d \times h \times w}\}_{i=1}^m$,  the relative computational cost of ONI over the convolutional layer is $\frac{2 n}{m h w} + \frac{3 N n^2}{m d h w F_h F_w}$.  In this section, we compare the of wall clock time between the convolution wrapping with our ONI and the standard convolution. In this experiment, our ONI is implemented based on Torch \cite{2011_torch} and we wrap it to the  `cudnn' convolution \cite{2014_cudnn}. The experiments are run on a TITAN Xp. 

We fix the input to size $h=w=32$, and vary the kernel size ($F_h$ and $F_w$), the feature dimensions ($n$ and $d$) and the batch size $m$.  Table \ref{tab:timeCost} shows the wall clock time under different configurations. We compare the standard `cudnn' convolution (denoted as `cudnn') and the `cudnn' wrapped with our ONI (denoted as `cudnn ~+~ ONI'). 

We observe that our method introduces negligible  computational costs when using a $3 \times 3$ convolution, feature dimension $n=d=256$ and batch size of $m=256$. Our method may degenerate in efficiency with a smaller kernel size, larger feature dimension and smaller batch size, based on the computational complexity analysis. However, our method (with iteration of 5) `cudnn ~+~ ONI-T5' only costs $1.48\times$ over the standard convolution `cudnn', under the worst configuration,  $F_h=F_w=1$, $n=d=1024$ and m=32. 	  

%

\section{Proof of Theorems}
\label{sup:sec:proofTheorem}
Here we prove the two theorems described in Sections~\ref{sec:property} and~\ref{sec:analysis} of the paper. 

\noindent\textbf{Theorem 1.}
	Let $\mathbf{\hat{h}}= \mathbf{W} \mathbf{x}$, where $\mathbf{W} \mathbf{W}^T = \mathbf{I}$ and $\mathbf{W} \in \mathbb{R}^{n \times d}$.	
	Assume:  (1) $\mathbb{E}_{\mathbf{x}}(\mathbf{x})=\mathbf{0}$, $cov(\mathbf{x})=\sigma_1^2 \mathbf{I}$,
	and (2) $\mathbb{E}_{\frac{\partial \mathcal{L}}{\partial \mathbf{\hat{h}}}}(\frac{\partial \mathcal{L}}{\partial \mathbf{\hat{h}}})=\mathbf{0}$,
	$cov(\frac{\partial \mathcal{L}}{\partial \mathbf{\hat{h}}})=\sigma_2^2 \mathbf{I}$.	
	If $n=d$, we have the following properties:
	(1) $\| \mathbf{\hat{h}}  \|=  \|  \mathbf{x}  \|$;
	(2) $\mathbb{E}_{\mathbf{\hat{h}}}(\mathbf{\hat{h}})=\mathbf{0}$, $cov(\mathbf{\hat{h}})=\sigma_1^2 \mathbf{I}$;
	(3) $\| \frac{\partial \mathcal{L}}{\partial \mathbf{x}}  \|=  \|  \frac{\partial \mathcal{L}}{\partial \mathbf{\hat{h}}} \|$;
	(4) $\mathbb{E}_{  \frac{\partial \mathcal{L}}{\partial \mathbf{x}}  }( \frac{\partial \mathcal{L}}{\partial \mathbf{x}} )=\mathbf{0}$, $cov(\frac{\partial \mathcal{L}}{\partial \mathbf{x}})=\sigma_2^2 \mathbf{I}$.
	In particular, if $n<d$, property (2) and (3) hold; if $n>d$, property (1) and (4) hold.

\begin{proof}
	
	Based on $n=d$ and  $\mathbf{W} \mathbf{W}^T = \mathbf{I}$, we have that $\mathbf{W}$ is a square orthogonal matrix. We thus have  $\mathbf{W}^T \mathbf{W}= \mathbf{I}$.
	Besides, we have $ \frac{\partial \mathcal{L}}{\partial \mathbf{x}} =  \frac{\partial \mathcal{L}}{\partial \hat{\mathbf{h}}} \mathbf{W}$\footnote{We follow the common setup where  the vectors are column vectors when their derivations are row vectors. }. 
	
	(1) 
	Therefore, we have
	\begin{equation}
	\| \hat{\mathbf{h}}  \|^2 =  \hat{\mathbf{h}}^T \hat{\mathbf{h}} =  \mathbf{x}^T  \mathbf{W}^T \mathbf{W} \mathbf{x}  = \mathbf{x}^T \mathbf{x}
	=\| \mathbf{x} \|^2.
	\end{equation}
	We thus get $\| \hat{\mathbf{h}}  \|=  \|  \mathbf{x}  \|$.
	
	(2) It's easy to calculate:
	\begin{eqnarray}
	&\mathbb{E}_{\hat{\mathbf{h}}}(\hat{\mathbf{h}})&=
	\mathbb{E}_{\mathbf{x}} ( \mathbf{W} \mathbf{x})=
	\mathbf{W} \mathbb{E}_{\mathbf{x}} ( \mathbf{x})= \mathbf{0}.
	\end{eqnarray}
	
	The covariance of $\hat{\mathbf{h}}$ is given by
	\begin{align}
	cov(\hat{\mathbf{h}})&=\mathbb{E}_{\hat{\mathbf{h}}} ((\hat{\mathbf{h}}-\mathbb{E}_{\hat{\mathbf{h}}}(\hat{\mathbf{h}}))  
	\cdot (\hat{\mathbf{h}}-\mathbb{E}_{\hat{\mathbf{h}}}(\hat{\mathbf{h}}))^T )\nonumber \\
	&=\mathbb{E}_{\mathbf{x}}(\mathbf{W}(\mathbf{x}-\mathbb{E}_{\mathbf{x}}(\mathbf{x})) 
	\cdot (\mathbf{W}(\mathbf{x}-\mathbb{E}_{\mathbf{x}}(\mathbf{x})))^T)  \nonumber \\
	&=\mathbf{W} \mathbb{E}_{\mathbf{x}}((\mathbf{x}-\mathbb{E}_{\mathbf{x}}(\mathbf{x})) \cdot (\mathbf{x}-\mathbb{E}_{\mathbf{x}}(\mathbf{x}))^T) \mathbf{W}^T  \nonumber \\
	&=\mathbf{W} cov(\mathbf{x}) \mathbf{W}^T  \nonumber \\
	&=\mathbf{W} \sigma_1^2 \mathbf{I} \mathbf{W}^T  \nonumber \\
	&=\sigma_1^2 \mathbf{W} \mathbf{W}^T  \nonumber \\
	&= \sigma_1^2.
	\end{align}
	
	(3) Similar to the proof of (1), 
	\begin{align}
	\| \frac{\partial \mathcal{L}}{\partial \mathbf{x}} \|^2
	&=  \frac{\partial \mathcal{L}}{\partial \mathbf{x}} \frac{\partial \mathcal{L}}{\partial \mathbf{x}}^T
	= \frac{\partial \mathcal{L}}{\partial \hat{\mathbf{h}}} \mathbf{W} \mathbf{W}^T \frac{\partial \mathcal{L}}{\partial \hat{\mathbf{h}}}^T \nonumber \\
	&= \frac{\partial \mathcal{L}}{\partial \hat{\mathbf{h}}} \frac{\partial \mathcal{L}}{\partial \hat{\mathbf{h}}}^T
	= \|\frac{\partial \mathcal{L}}{\partial \hat{\mathbf{h}}}\|^2.
	\end{align}
	We thus have $\| \frac{\partial \mathcal{L}}{\partial \mathbf{x}}  \|=  \|  \frac{\partial \mathcal{L}}{\partial \hat{\mathbf{h}}} \|$.
	
	(4) Similar to the proof of (2),  we have
	\begin{equation}
	\mathbb{E}_{  \frac{\partial \mathcal{L}}{\partial \mathbf{x}}  }( \frac{\partial \mathcal{L}}{\partial \mathbf{x}} )=
	\mathbb{E}_{  \frac{\partial \mathcal{L}}{\partial  \hat{\mathbf{h}}}} ( \frac{\partial \mathcal{L}}{\partial \hat{\mathbf{h}} } \mathbf{W} )
	=\mathbb{E}_{  \frac{\partial \mathcal{L}}{\partial  \hat{\mathbf{h}}}} ( \frac{\partial \mathcal{L}}{\partial \hat{\mathbf{h}} } ) \mathbf{W}
	=\mathbf{0}.
	\end{equation}

	The covariance of $\frac{\partial \mathcal{L}}{\partial \mathbf{x}} $ is given by
	\begin{small}
		\begin{align}
		cov(\frac{\partial \mathcal{L}}{\partial \mathbf{x}})&=\mathbb{E}_{ \frac{\partial \mathcal{L}}{\partial \mathbf{x}}  } ((  \frac{\partial \mathcal{L}}{\partial \mathbf{x}} -\mathbb{E}_{\frac{\partial \mathcal{L}}{\partial \mathbf{x}} }(\frac{\partial \mathcal{L}}{\partial \mathbf{x}} ))^T  
		(\frac{\partial \mathcal{L}}{\partial \mathbf{x}} -\mathbb{E}_{\frac{\partial \mathcal{L}}{\partial \mathbf{x}} }(\frac{\partial \mathcal{L}}{\partial \mathbf{x}} )) )\nonumber \\
		&=\mathbb{E}_{ \frac{\partial \mathcal{L}}{\partial \hat{\mathbf{h}} }}( (( \frac{\partial \mathcal{L}}{\partial \hat{\mathbf{h}} }-\mathbb{E}_{ \frac{\partial \mathcal{L}}{\partial \hat{\mathbf{h}} }}( \frac{\partial \mathcal{L}}{\partial \hat{\mathbf{h}} }))\mathbf{W})^T 
		( \frac{\partial \mathcal{L}}{\partial \hat{\mathbf{h}} }-\mathbb{E}_{ \frac{\partial \mathcal{L}}{\partial \hat{\mathbf{h}} }}( \frac{\partial \mathcal{L}}{\partial \hat{\mathbf{h}} }))\mathbf{W})  \nonumber \\
		&=\mathbf{W}^T \mathbb{E}_{ \frac{\partial \mathcal{L}}{\partial \hat{\mathbf{h}} }}(( \frac{\partial \mathcal{L}}{\partial \hat{\mathbf{h}} }-\mathbb{E}_{ \frac{\partial \mathcal{L}}{\partial \hat{\mathbf{h}} }}( \frac{\partial \mathcal{L}}{\partial \hat{\mathbf{h}} }))^T
		( \frac{\partial \mathcal{L}}{\partial \hat{\mathbf{h}} }-\mathbb{E}_{ \frac{\partial \mathcal{L}}{\partial \hat{\mathbf{h}} }}( \frac{\partial \mathcal{L}}{\partial \hat{\mathbf{h}} }))) \mathbf{W}  \nonumber \\
		&=\mathbf{W}^T cov( \frac{\partial \mathcal{L}}{\partial \hat{\mathbf{h}} }) \mathbf{W}  \nonumber \\
		&=\mathbf{W}^T \sigma_2^2 \mathbf{I} \mathbf{W}  \nonumber \\
		&=\sigma_2^2 \mathbf{W}^T \mathbf{W} \nonumber \\
		&= \sigma_2^2.
		\end{align}
	\end{small}	
	Besides, if $n<d$, it is easy to show that properties (2) and (3) hold; if $n>d$, properties (1) and (4) hold.	
	
\end{proof}

\noindent
\textbf{Theorem 2.}
	Let $\mathbf{h}= max (0,\mathbf{W} \mathbf{x})$, where $\mathbf{W} \mathbf{W}^T = \sigma^2 \mathbf{I}$ and $\mathbf{W} \in \mathbb{R}^{n \times d}$.  Assume $\mathbf{x}$ is a normal distribution with  $\mathbb{E}_{\mathbf{x}}(\mathbf{x})=\mathbf{0}$, $cov(\mathbf{x})=\mathbf{I}$. Denote the Jacobian matrix as $ \mathbf{J}= \frac{\partial  \mathbf{h}}{\partial \mathbf{x}} $. If $\sigma^2 =2$, we have  $\mathbb{E}_\mathbf{x} (\mathbf{J} \mathbf{J}^T) = \mathbf{I}$.

\begin{proof}
	For denotation, we use $\mathbf{A}_{i:}$ and $\mathbf{A}_{:j}$ to represent the $i$-th row and the $j$-th column of $\mathbf{A}$, respectively.
	Based on $\mathbf{W} \mathbf{W}^T=\sigma^2 \mathbf{I}$, we obtain $\mathbf{W}_{i:} (\mathbf{W}_{j:})^T=0$ for $i \neq j$ and  $\mathbf{W}_{i:} (\mathbf{W}_{i:})^T=\sigma^2 $ otherwise.  Let $\hat{\mathbf{h}}= \mathbf{W} \mathbf{x}$. This yields
	\begin{equation}
	\label{equ:jacobi}
	\mathbf{J}_{i:}=\frac{\partial \mathbf{h}_i}{\partial \hat{\mathbf{h}}_i } \frac{\partial \hat{\mathbf{h}}_i }{ \mathbf{x}} 
	= \frac{\partial \mathbf{h}_i }   {\partial  \hat{\mathbf{h}}_i } \mathbf{W}_{i:}.       
	\end{equation}
	
	Denote $\mathbf{M}=\mathbf{J}\mathbf{J}$. This yields the following equation from Eqn. \ref{equ:jacobi}:
	\begin{align}
	\mathbf{M}_{ij}&=\mathbf{J}_{i:} (\mathbf{J}_{j:})^T \nonumber \\
	& = \frac{\partial \mathbf{h}_i }   {\partial  \hat{\mathbf{h}}_i } \mathbf{W}_{i:} (\mathbf{W}_{j:})^T              
	\frac{\partial \mathbf{h}_j}{\partial \hat{\mathbf{h}}_j }   \nonumber \\
	&= \frac{\partial \mathbf{h}_i }   {\partial  \hat{\mathbf{h}}_i }  \frac{\partial \mathbf{h}_j }   {\partial  \hat{\mathbf{h}}_j } 
	(\mathbf{W}_{i:}   (\mathbf{W}_{j:} )^T   ).  
	\end{align}   
	If $i\neq j$, we obtain $\mathbf{M}_{ij}=0$.  For $i=j$, we have:
	\begin{equation}
	\mathbf{M}_{ii}=\mathbf{J}_{i:} (\mathbf{J}_{i:})^T
	= (\frac{\partial \mathbf{h}_i }   {\partial  \hat{\mathbf{h}}_i })^2  \sigma^2  =1(\hat{\mathbf{h}}_i>0) \sigma^2,
	\end{equation}  
	where $1(\hat{\mathbf{h}}_i>0)$ indicates 1 for $\hat{\mathbf{h}}_i>0$ and 0 otherwise. Since $\mathbf{x}$ is a normal distribution with $\mathbb{E}_{\mathbf{x}}(\mathbf{x})=\mathbf{0}$ and $cov(\mathbf{x})=\mathbf{I}$,  we have that $\hat{\mathbf{h}}$ is also a normal distribution, with $\mathbb{E}_{\hat{\mathbf{h}}}(\hat{\mathbf{h}})=\mathbf{0}$ and $cov(\hat{\mathbf{h}})= \mathbf{I}$, based on Theorem \ref{th:norm}.   We thus obtain 
	\begin{equation}
	\mathbb{E}_{\mathbf{x}} (\mathbf{M}_{ii})
	= \mathbb{E}_{\hat{\mathbf{h}}_i} 1(\hat{\mathbf{h}}_i>0) \sigma^2 = \frac{1}{2} \sigma^2.
	\end{equation} 
	
	Therefore, $\mathbb{E}_{\mathbf{x}} (\mathbf{J} \mathbf{J}^T)=\mathbb{E}_{\mathbf{x}} (\mathbf{M}) = \frac{\sigma^2}{2} \mathbf{I} =\mathbf{I}$.  
\end{proof}

\section{Details and More Experimental Results on Discriminative Classification}
\label{sup:sec:experiments}
\subsection{MLPs on Fashion-MNIST}
\label{sup:sec:exp:MLP}
\paragraph{Details of Experimental Setup}
Fashion-MNIST consists of  $60k$ training and $10k$ test images. Each image has a size of $28 \times 28$,  and is associated with a label from one of 10 classes. 
We use the MLP with varying depths  and the number of neurons in each layer is 256. We use  ReLU \cite{2010_ICML_Nair} as the nonlinearity. The weights in each layer are initialized by random initialization \cite{1998_NN_Yann} and we  use  an iteration number of 5 for ONI, unless otherwise stated.  We employ stochastic gradient descent (SGD) optimization with a batch size of 256, and the learning rates are selected, based on the validation set ($5,000$ samples from the training set), from $\{0.05, 0.1, 0.5, 1\}$. 

\subsubsection{Vanished Activations and Gradients}
In Section~\ref{sec:MLP} of the paper, we observe that the deeper neural networks with orthogonal weight matrices are difficult to train without scaling them by  a factor of $\sqrt{2}$. We argue the main reason for this is that the activation and gradient are exponentially vanished. Here we provide the details. 

\begin{figure}[t]
	\centering
	\vspace{-0.1in}
	\hspace{-0.2in}	\subfigure[]{
		\begin{minipage}[c]{.46\linewidth}
			\centering
			\includegraphics[width=4.20cm]{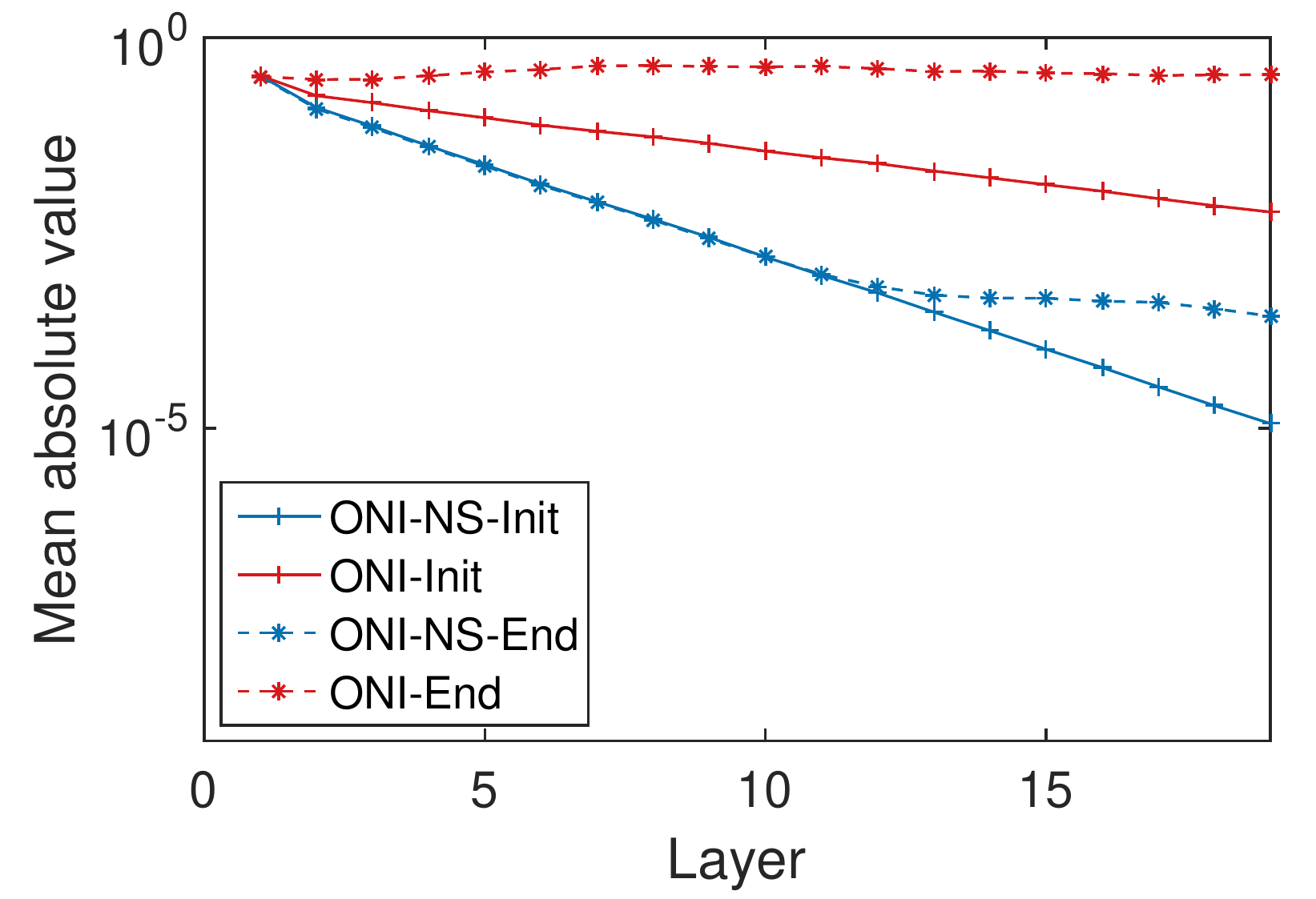}
		\end{minipage}
	}
	\hspace{0.05in}	\subfigure[]{
		\begin{minipage}[c]{.46\linewidth}
			\centering
			\includegraphics[width=4.20cm]{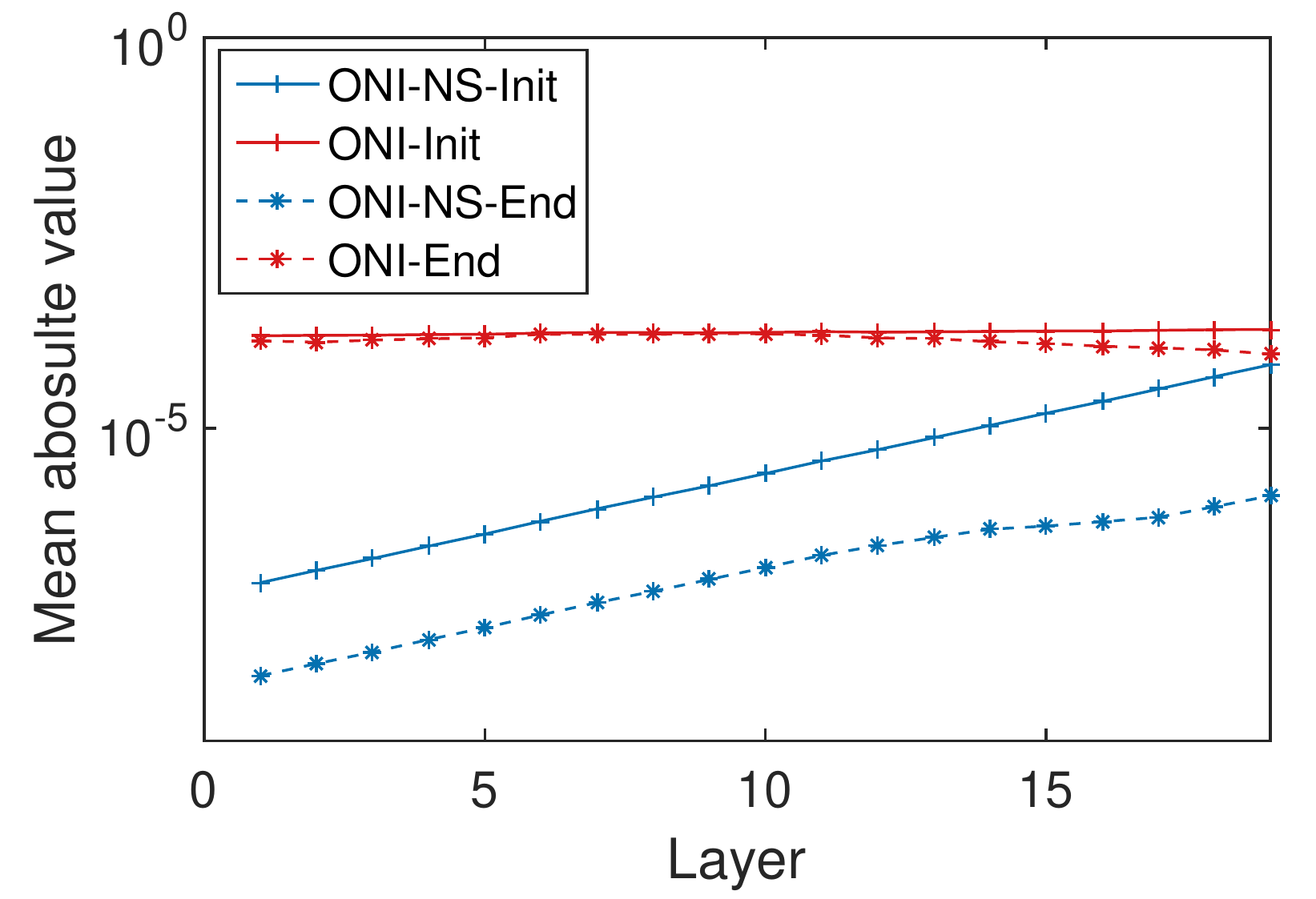}
		\end{minipage}
	}
	\caption{ The magnitude of the activation and gradient for each layer on a 20-layer MLP.  (a) The mean absolute value of the activation $\sigma_{\mathbf{x}}$ for each layer; and (b) the mean absolute value of the gradient $\sigma_{\frac{\partial \mathcal{L}}{\partial \mathbf{h}}}$ for each layer.}
	\label{fig:Norm_Gradient}
\end{figure}

We evaluate the mean absolute value of the activation: $\sigma_{\mathbf{x}}=\sum_{i=1}^{m} \sum_{j=1}^{d} | \mathbf{x}_{ij} |$ for the layer-wise input $\mathbf{x} \in \mathbb{R}^{m \times d}$, and the mean absolute value of the gradient: $\sigma_{\frac{\partial \mathcal{L}}{\partial \mathbf{h}}}=\sum_{i=1}^{m} \sum_{j=1}^{n} | \frac{\partial \mathcal{L}}{\partial \mathbf{h}}_{ij} |$ for the layer-wise gradient $\frac{\partial \mathcal{L}}{\partial \mathbf{h}} \in \mathbb{R}^{m \times n}$. Figure \ref{fig:Norm_Gradient} show the results on the 20-layer MLP. 
`ONI-NS-Init' (`ONI-NS-End') indicates the ONI method without scaling a factor of $\sqrt{2}$ during initialization (the end of training). We observe that `ONI-NS' suffers from a vanished activation and gradient during  training.  Our `ONI' with a scaling factor of  $\sqrt{2}$ has no vanished gradient. 

\vspace{-0.16in}
\subsubsection{Effects of Groups} 
\begin{figure}[t]
	\centering
	\vspace{-0.1in}
	\hspace{-0.15in}	\subfigure[Train Error]{
		\begin{minipage}[c]{.46\linewidth}
			\centering
			\includegraphics[width=4.1cm]{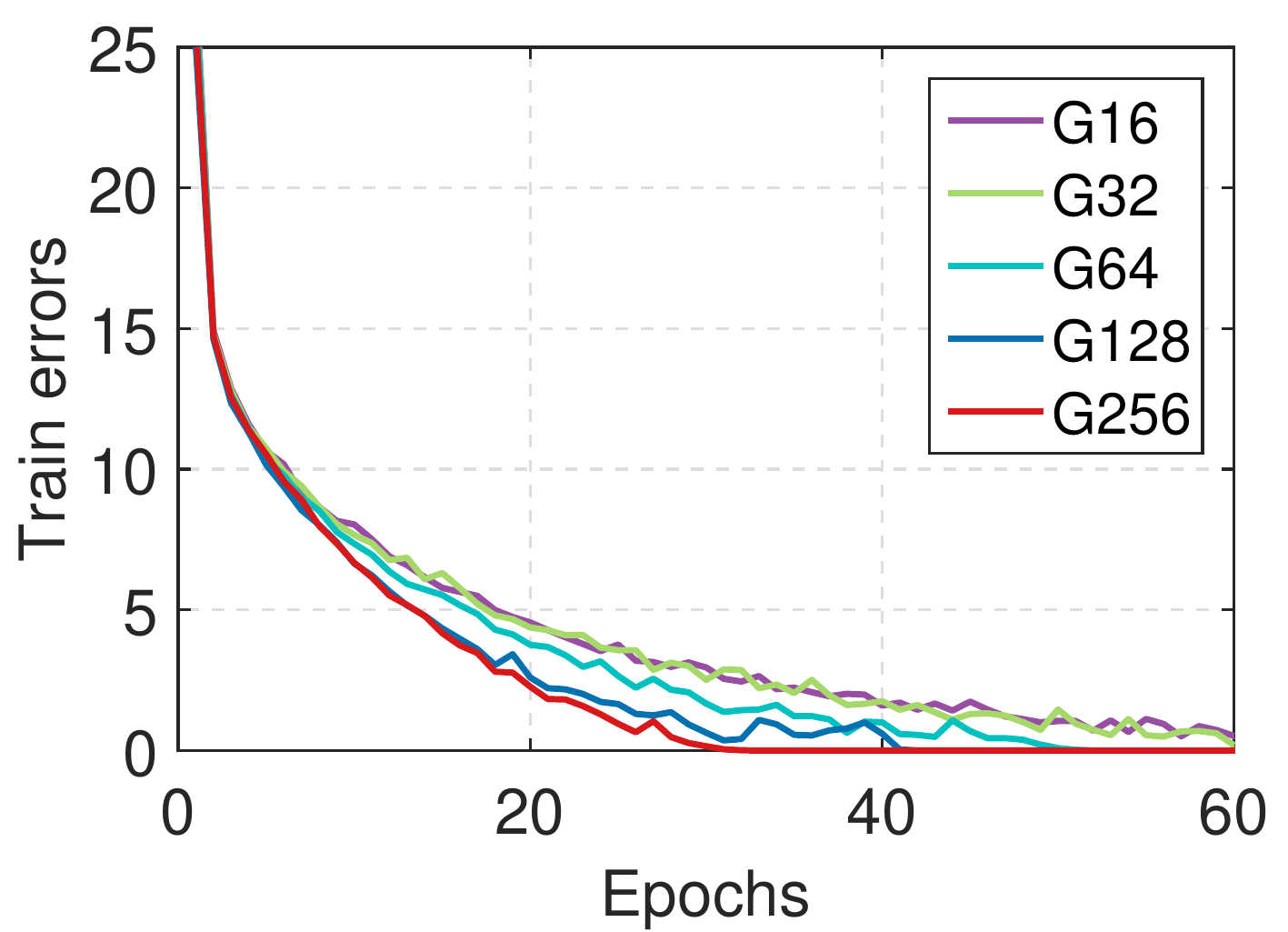}
		\end{minipage}
	}
	\hspace{0.05in}	\subfigure[Test Error]{
		\begin{minipage}[c]{.46\linewidth}
			\centering
			\includegraphics[width=4.1cm]{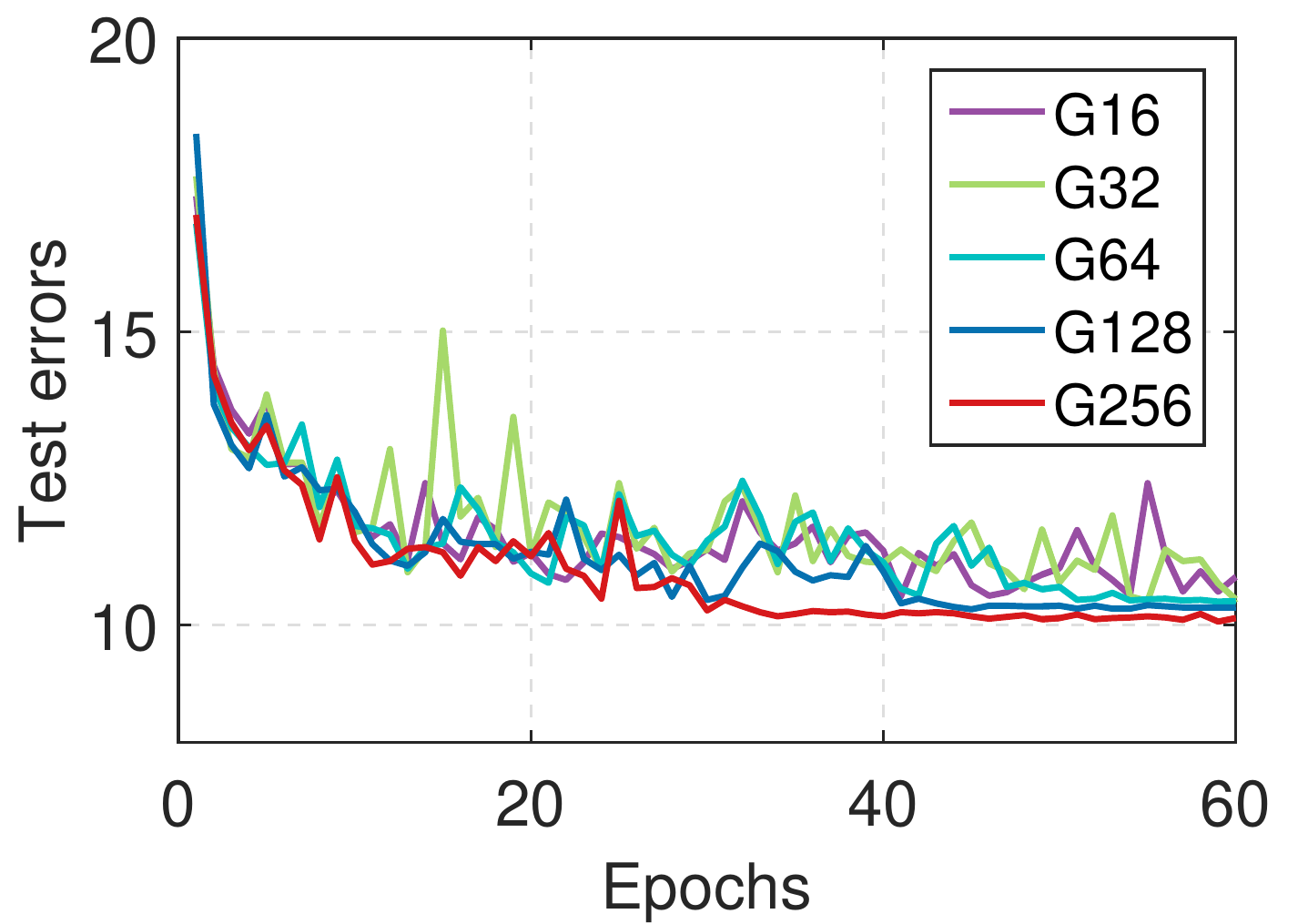}
		\end{minipage}
	}
	\caption{Effects of group size $G$ in proposed `ONI'. We evaluate the (a) training and (b) test errors on a 10-layer MLP.} 
	\label{fig:MLP-Group}
	\vspace{-0.1in}
\end{figure}

We further explore the group based orthogonalization method \cite{2018_AAAI_Huang} on ONI.  We vary the group size $G$ in $\{16, 32, 64, 128, 256 \}$, and show the results in Figure \ref{fig:MLP-Group}.  We observe that our ONI can achieve slightly better performance with an increasing group size. The main reason for this is that the group based orthogonalization cannot ensure the whole weight matrix to be   orthogonal. 

\subsection{CNNs on CIFAR10}
\label{sup:sec:exp:CNN}
We use the official training set of $50, 000$ images  and the standard test set of $10, 000$ images. The data preprocessing and data augmentation follow the commonly used mean$\&$std normalization and flip translation, as described in~\cite{2015_CVPR_He}. 

\subsubsection{VGG-Style Networks}
\paragraph{Details of Network Architectures}
The network starts with a convolutional layer of $32k$ filters, where $k$ is the varying width based on different configurations. We then sequentially stack  three blocks, each of which has $g$ convolutional layers, and the corresponding convolutional layers have a filter numbers of $32k$, $64k$ and $128k$, respectively, and feature maps sizes of $32\times32$,  $16\times16$ and $8\times8$, respectively. We use the first convolution in each block with  stride 2 to carry out spatial sub-sampling for feature maps.  The network ends with  global average pooling and follows a linear transformation.  
We vary the depth with $g$ in  $\{2,3,4 \}$ and the width with $k$ in $\{1,2,3\}$.

\vspace{-0.1in}
\paragraph{Experimental Setup }We use SGD with a  momentum of 0.9 and batch size of 128. The best initial learning rate is chosen from $\{0.01, 0.02, 0.05\}$ over the validation set of 5,000 samples from the training set, and we divide the learning rate by 5 at 80 and 120 epochs, ending the training at 160 epochs. 
For `OrthReg', we report the best results using a regularization coefficient $\lambda$ in $\{0.0001,0.0005\}$.
For `OLM', we use the group size of $G=64$ and full orthogonalization, and report the best result.

\begin{figure*}[t]
	\centering
	\subfigure[g=2, k=1]{
		\begin{minipage}[g=2, k=1]{.30\linewidth}
			\centering
			\includegraphics[width=4.80cm]{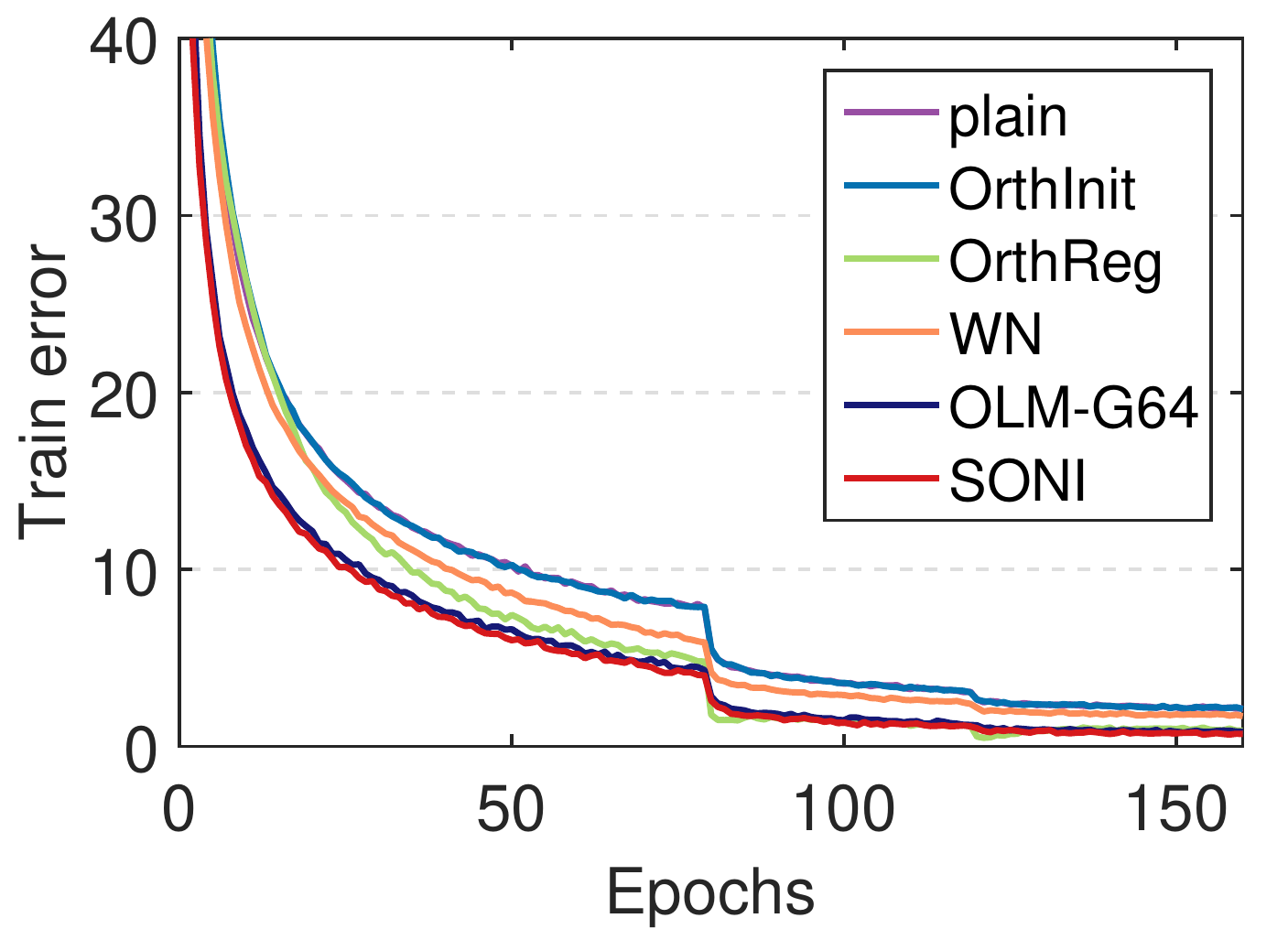}
		\end{minipage}
	}
	\subfigure[g=2, k=2]{
		\begin{minipage}[c]{.30\linewidth}
			\centering
			\includegraphics[width=4.80cm]{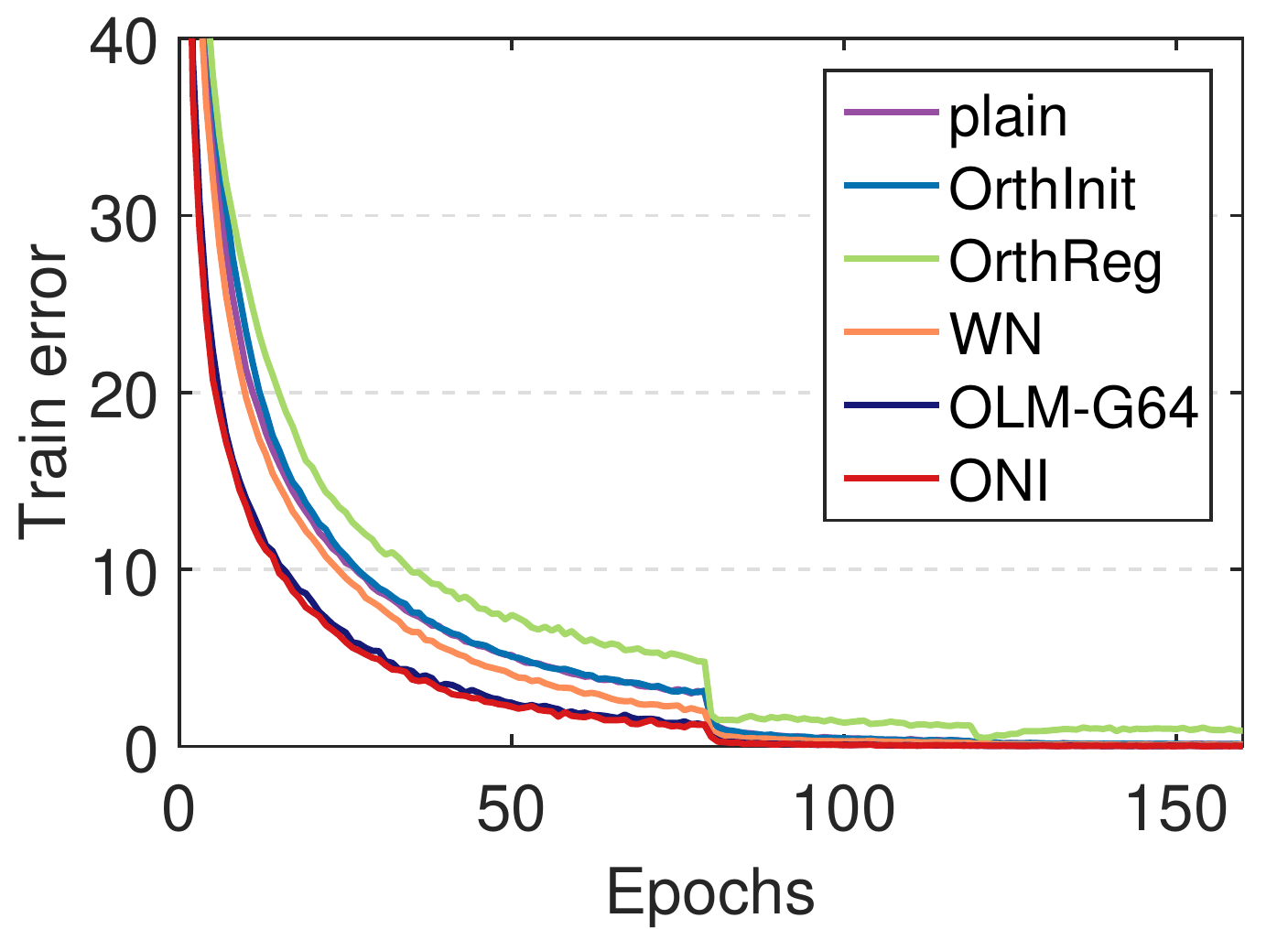}
		\end{minipage}
	}
	\subfigure[g=2, k=3]{
		\begin{minipage}[c]{.30\linewidth}
			\centering
			\includegraphics[width=4.80cm]{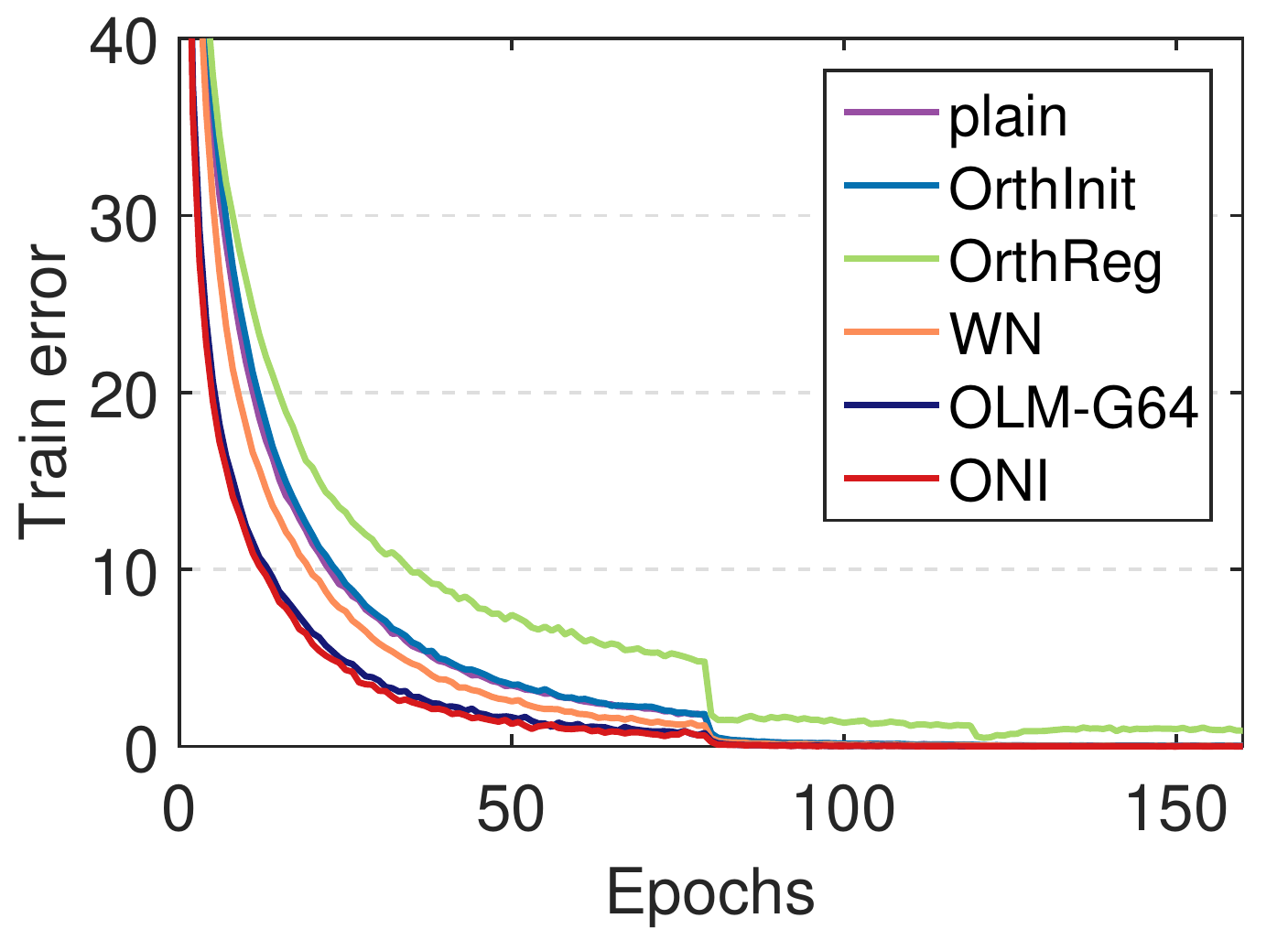}
		\end{minipage}
	}
	\\
	\vspace{-0.1in}
	\subfigure[g=3, k=1]{
		\begin{minipage}[c]{.30\linewidth}
			\centering
			\includegraphics[width=4.80cm]{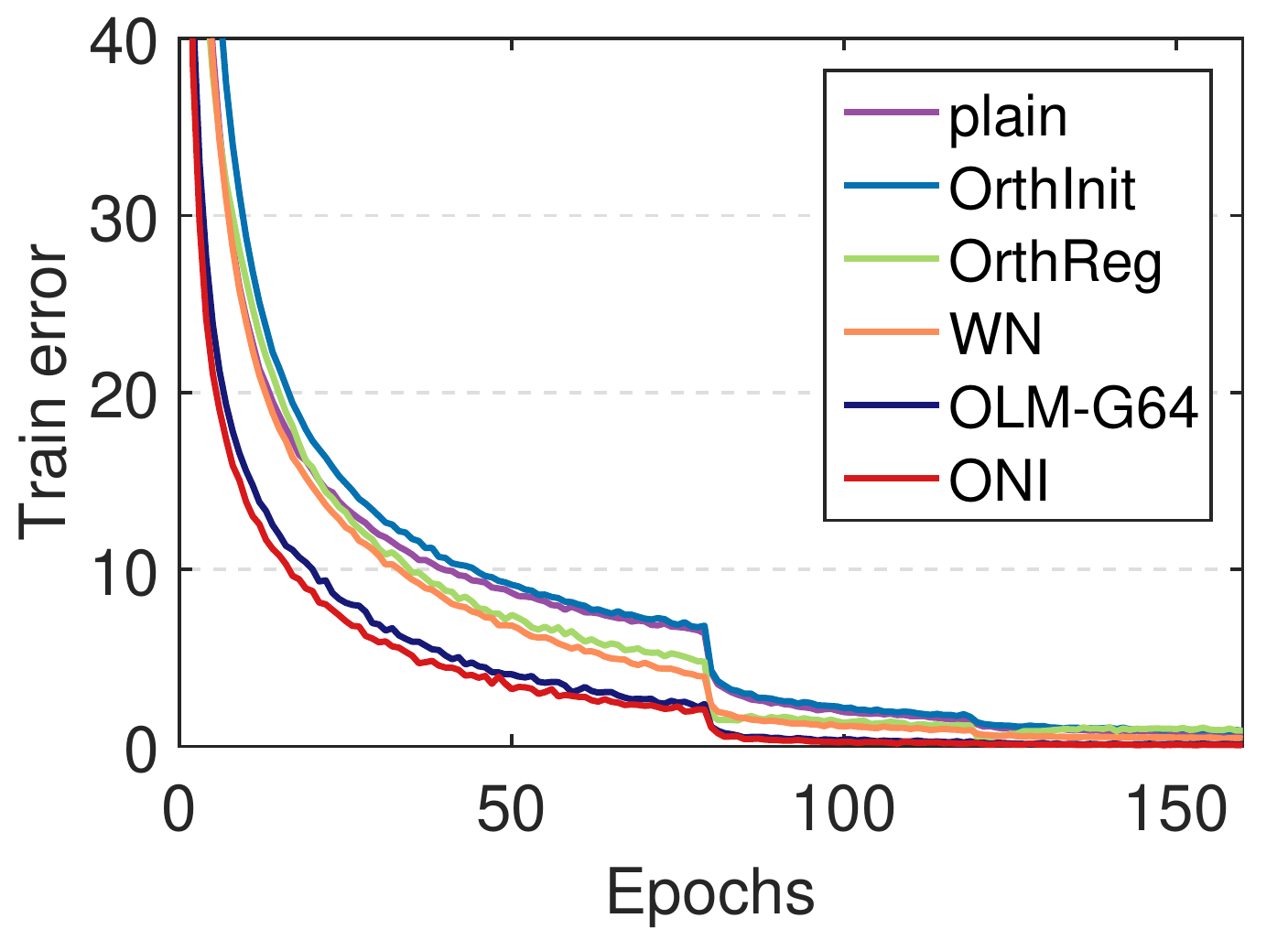}
		\end{minipage}
	}
	\subfigure[g=3, k=2]{
		\begin{minipage}[c]{.30\linewidth}
			\centering
			\includegraphics[width=4.80cm]{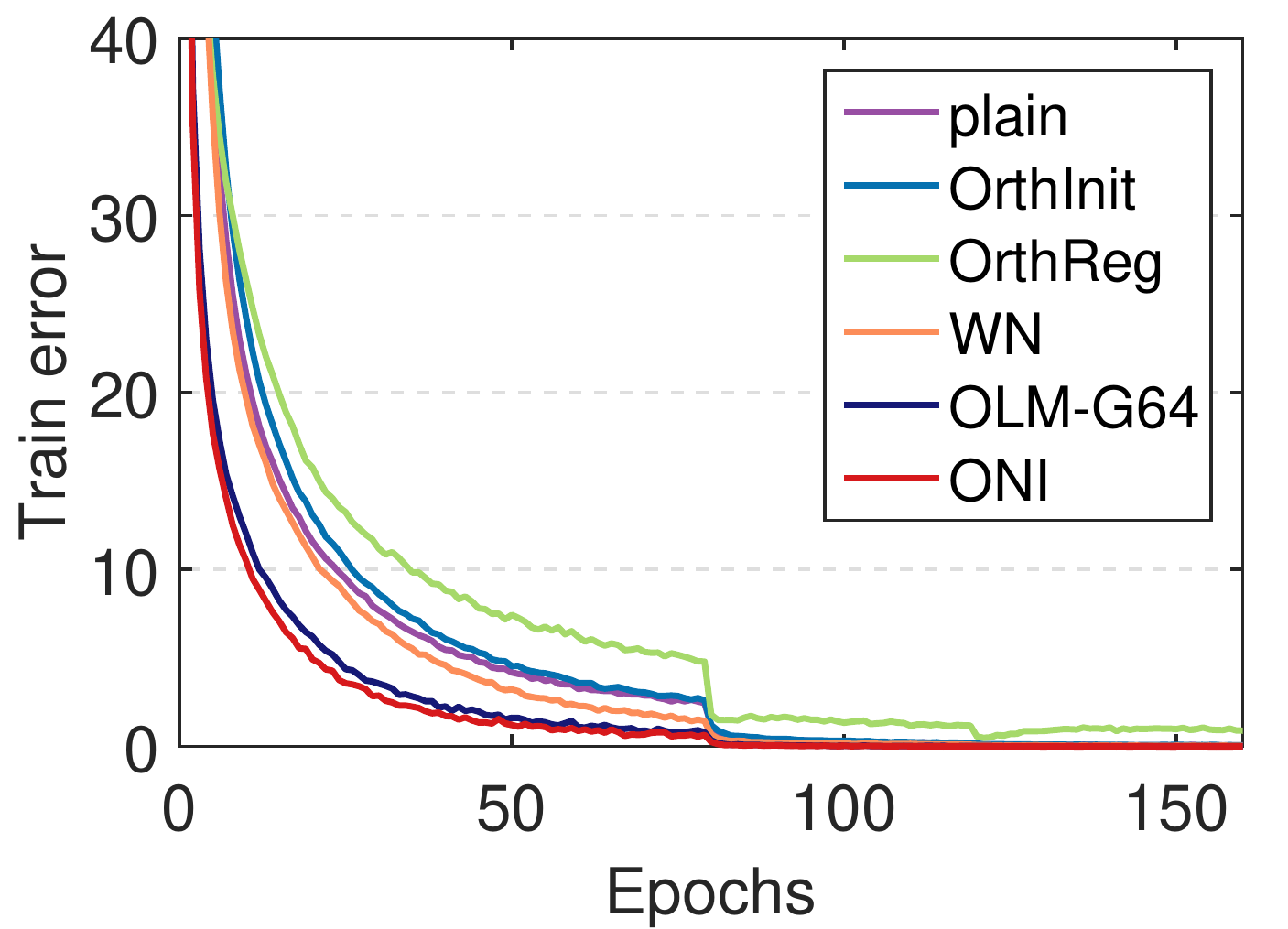}
		\end{minipage}
	}
	\subfigure[g=3, k=3]{
		\begin{minipage}[c]{.30\linewidth}
			\centering
			\includegraphics[width=4.80cm]{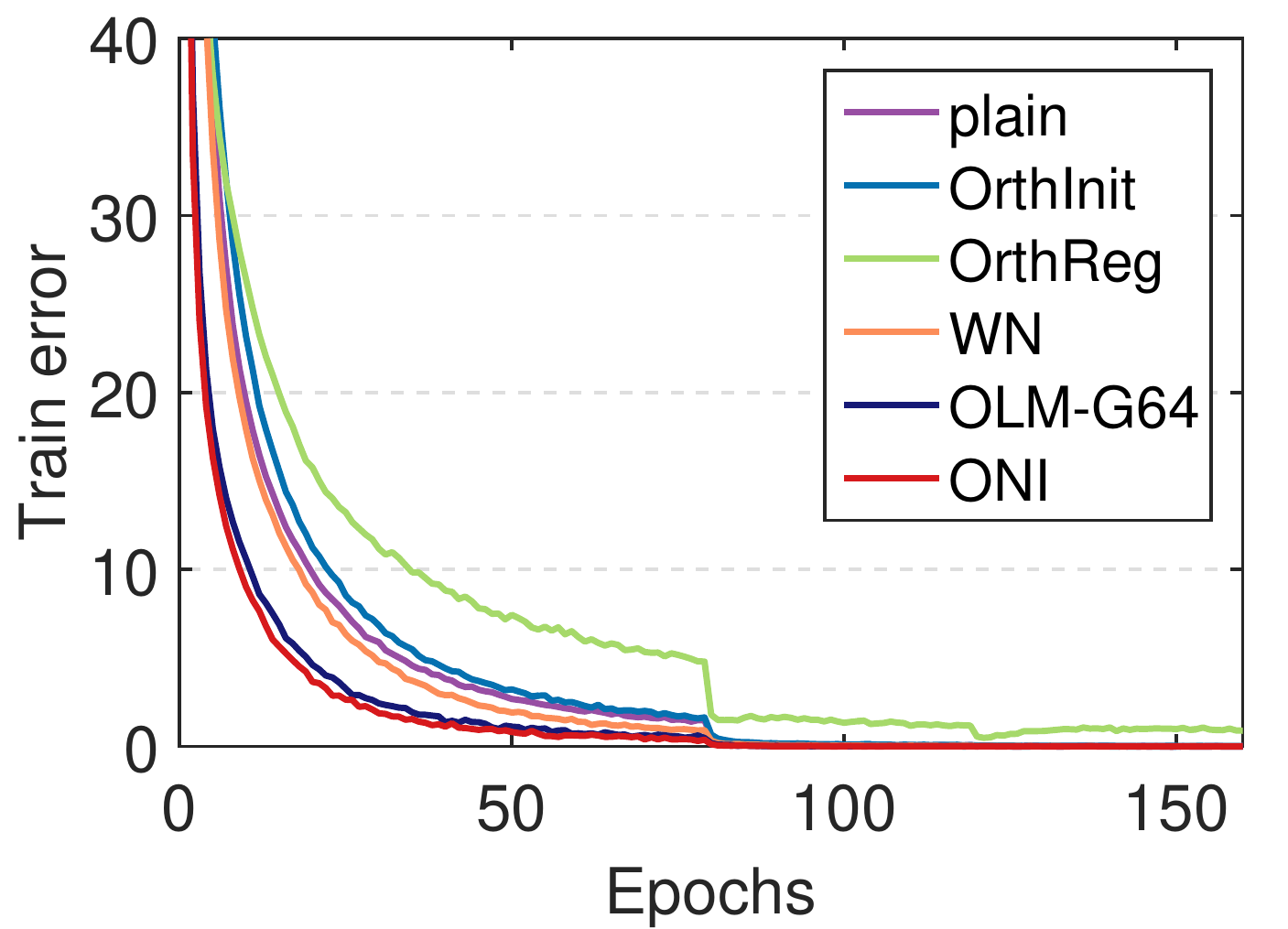}
		\end{minipage}
	}
	\\
	\vspace{-0.1in}
	\subfigure[g=4, k=1]{
		\begin{minipage}[c]{.30\linewidth}
			\centering
			\includegraphics[width=4.80cm]{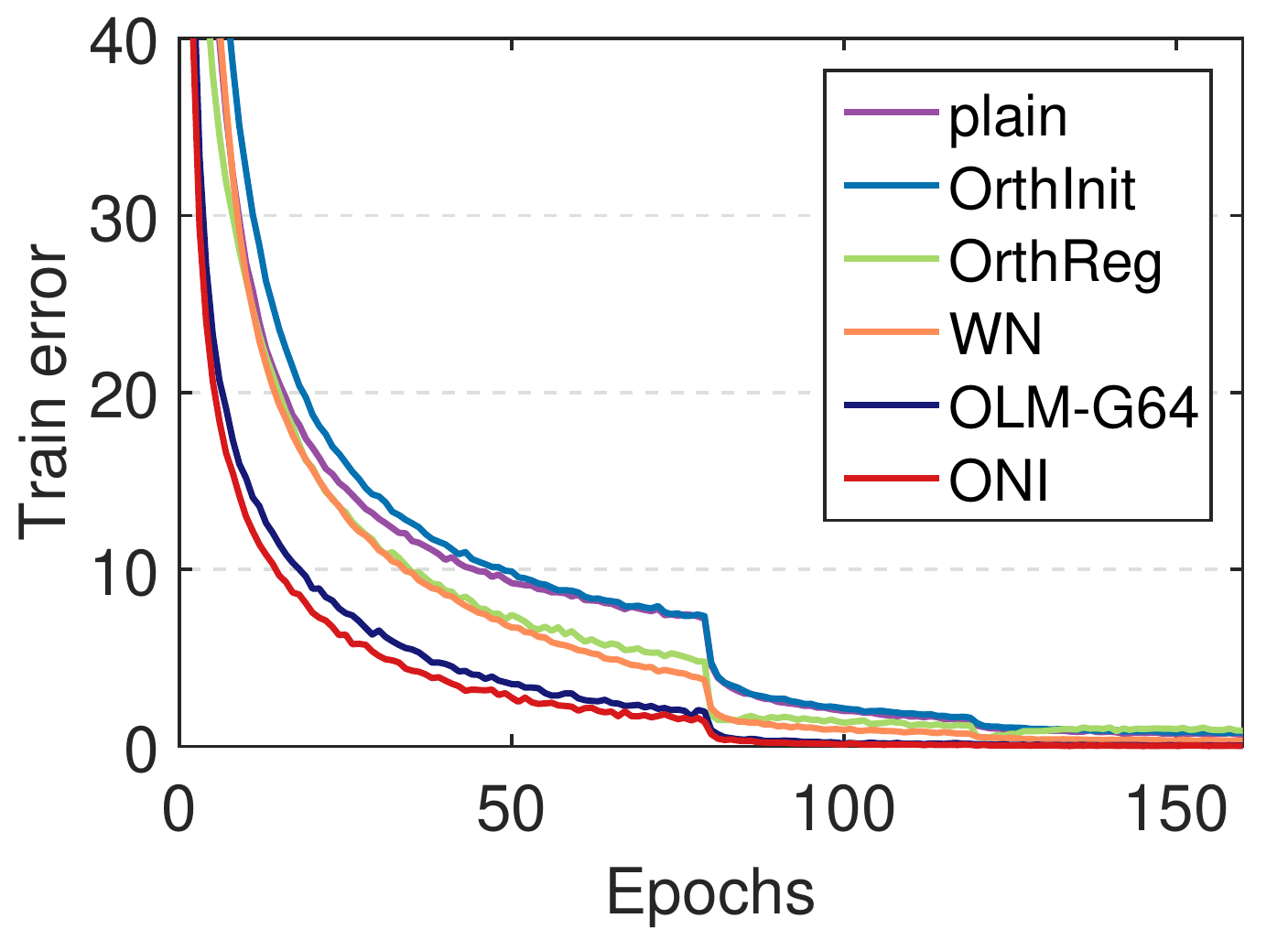}
		\end{minipage}
	}
	\subfigure[g=4, k=2]{
		\begin{minipage}[c]{.30\linewidth}
			\centering
			\includegraphics[width=4.80cm]{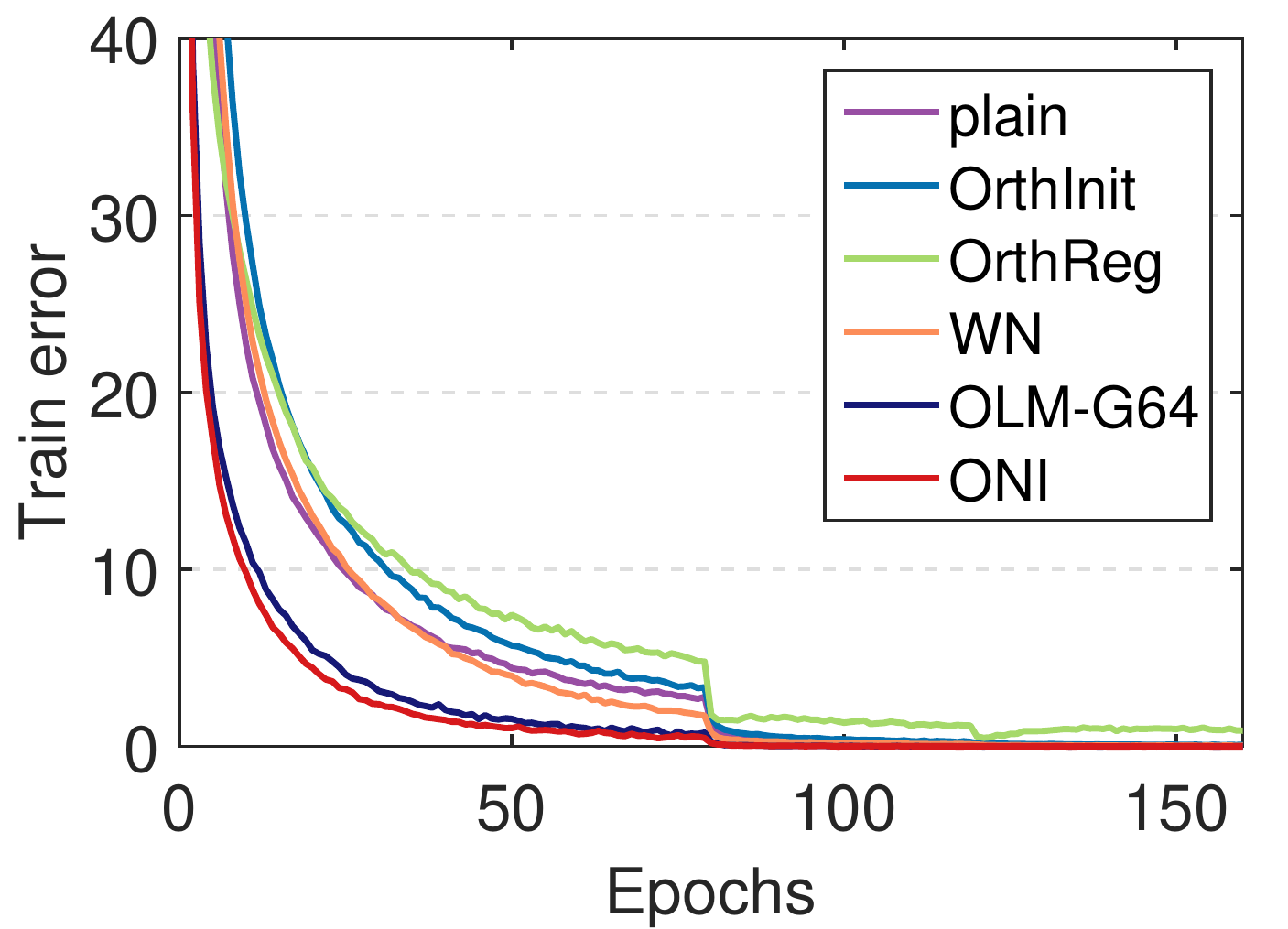}
		\end{minipage}
	}
	\subfigure[g=4, k=3]{
		\begin{minipage}[c]{.30\linewidth}
			\centering
			\includegraphics[width=4.80cm]{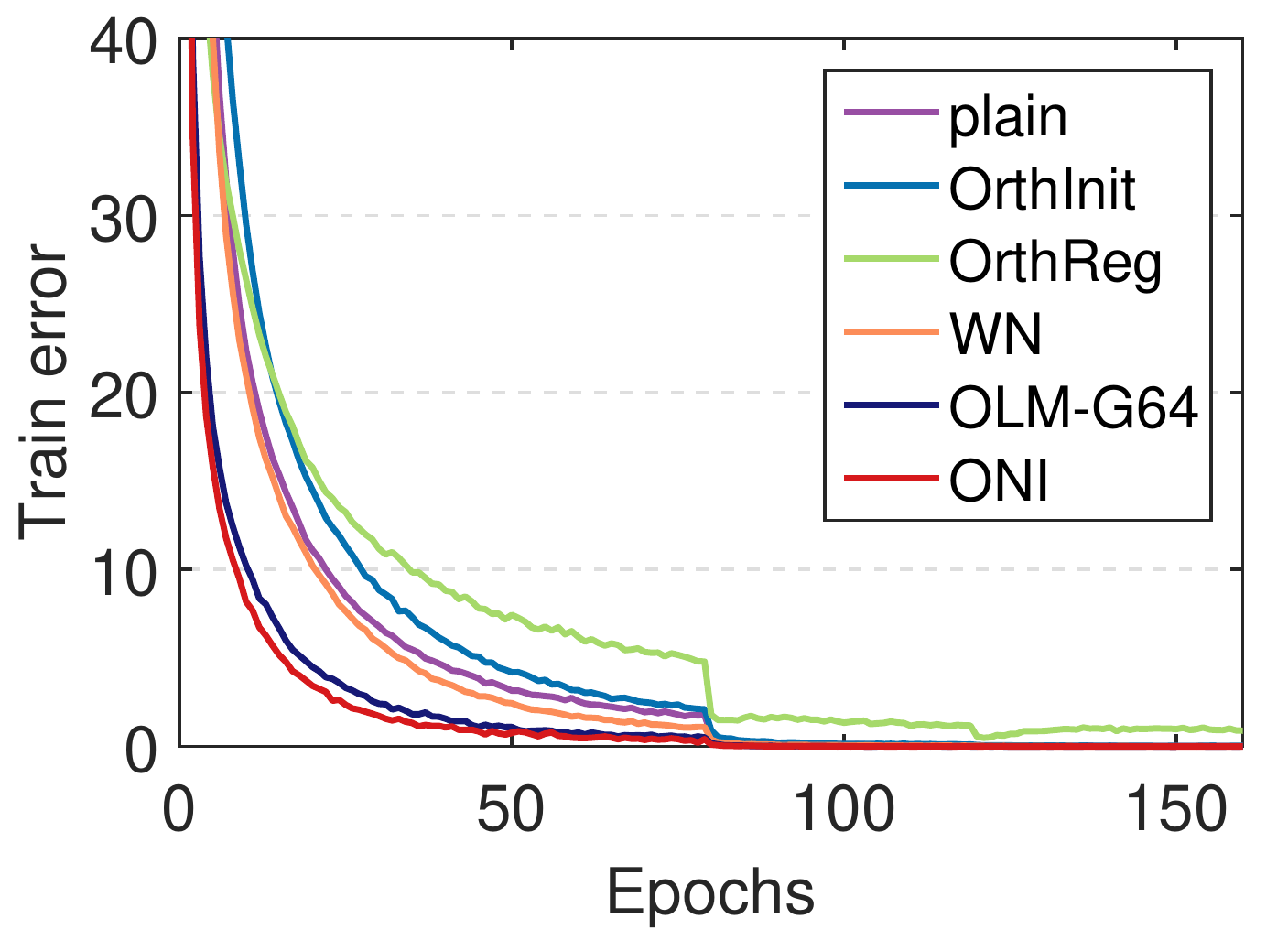}
		\end{minipage}
	}	
	\caption{Comparison of training errors on VGG-style networks for CIFAR-10 image classification.  From (a) to (i),  we vary the depth $3g+2$ and width $32k$, with $g \in \{2,3,4 \}$ and $k \in \{ 1,2,3 \}$. }
	\label{fig:expVGG}
\end{figure*}

\vspace{-0.1in}
\paragraph{Training Performance}
In Section~\ref{sec:CNN_CIFAR10} of the paper, we mention `ONI' and `OLM-$\sqrt{2}$' converge faster than other baselines, in terms of training epochs. 
Figure \ref{fig:expVGG} shows the training curves under different configurations (depth and width). It's clear that `ONI' and `OLM-$\sqrt{2}$' converge faster than other baselines under all network configurations, in terms of training epochs. The results support our conclusion that maintaining orthogonality can benefit optimization.  

\subsubsection{Residual Network without Batch Normalization}
Here we provide the details of the experimental setups and training performance of the experiments on a 110-layer residual network \cite{2015_CVPR_He} without batch normalization (BN) \cite{2015_ICML_Ioffe}, described in Section~\ref{sec:CNN_CIFAR10} of the paper.  
\vspace{-0.16in}
\paragraph{Experimental Setups} We run the experiments on one GPU. We apply SGD with a batch size of 128, a momentum of 0.9 and a weight decay of 0.0001. We set the initial learning rate to 0.1 by default, and divide it by 10 at 80 and 120 epochs, and terminate the training at 160 epochs. For Xavier Init \cite{2010_AISTATS_Glorot,2018_NIPS_Bjorck}, we search the initial learning rate from $\{0.1, 0.01, 0.001 \}$ and report the best result. For group normalization (GN) \cite{2018_ECCV_Wu}, we search the group size from $\{64, 32,16 \}$ and report the best result. For our ONI, we use the data-dependent initialization methods used in \cite{2016_CoRR_Salimans} to initial the learnable scale parameters.

\begin{figure}[t]
	\centering
	\hspace{-0.2in}	\subfigure[train error]{
		\begin{minipage}[c]{.46\linewidth}
			\centering
			\includegraphics[width=4.1cm]{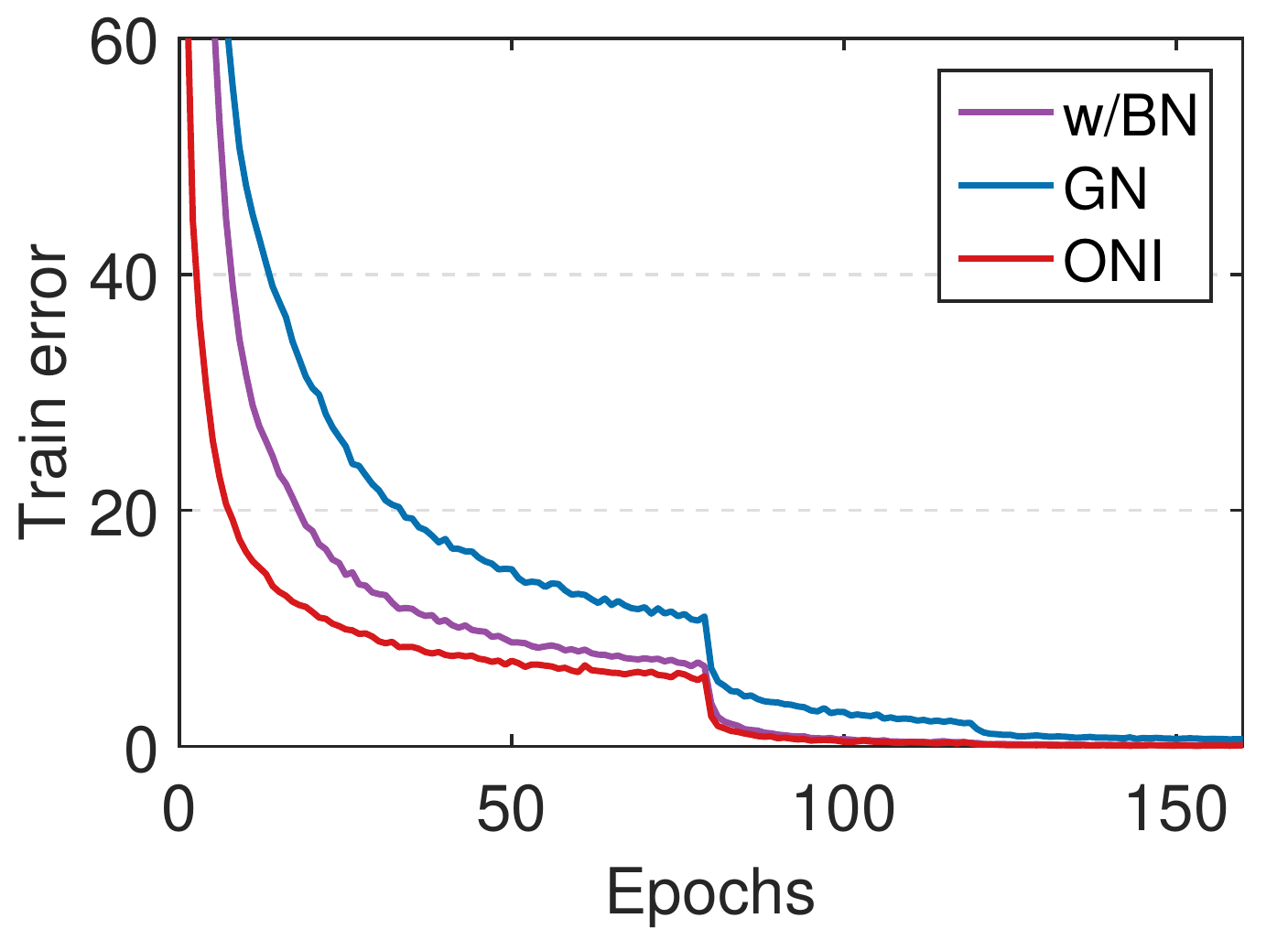}
		\end{minipage}
	}
	\hspace{0.05in}	\subfigure[test error]{
		\begin{minipage}[c]{.46\linewidth}
			\centering
			\includegraphics[width=4.1cm]{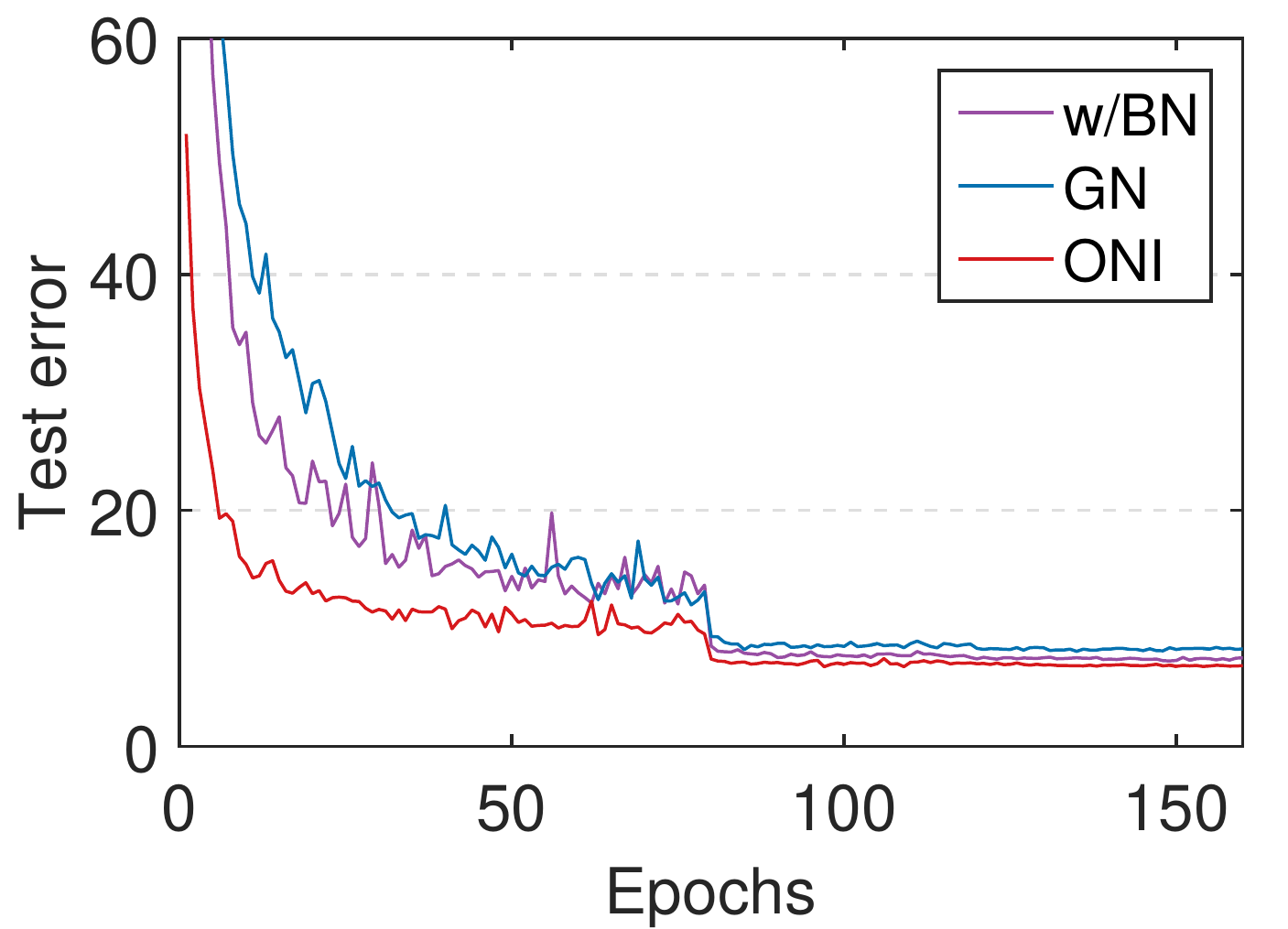}
		\end{minipage}
	}
	\caption{Training performance comparison on 110-layer residual network without batch normalization for CIFAR-10 dataset. `w/BN' indicates with BN. We show the (a) training error with respect to the epochs and (b) test error with respect to epochs.}
	\label{fig:ResNetwithoutBN}
\end{figure}

For small batch size experiments,  we train the network with an initial learning rate following  the linear learning rate scaling rule \cite{2017_Arxiv_Priya}, to adapt the batch size. 
\vspace{-0.16in}
\paragraph{Training Performance} Figure \ref{fig:ResNetwithoutBN} (a) and (b) show the training curve and test curve respectively. We observe that `ONI' converges significantly faster than `BN' and `GN', in terms of training epochs.

%

\subsection{Details of Experimental Setup on ImageNet}
\label{sup:sec:exp:ImageNet}
ImageNet-2012  consists of  1.28M images from 1,000 classes~\cite{2015_ImageNet}. We use the  official 1.28M labeled images provided for training and evaluate the top-1 and  top-5 test classification errors on the validation set, with 50k images. 

We  keep almost all the experimental settings the same as the publicly available PyTorch implementation \cite{2017_NIPS_pyTorch}:  we apply SGD with  a momentum of 0.9, and a weight decay of 0.0001; We train over 100 epochs in total and  set the initial learning rate to 0.1,  lowering it  by a factor of 10 at epochs 30, 60 and 90. For `WN' and `ONI', we don't us weight decay on the learnable scalar parameters.
\vspace{-0.1in}
\paragraph{VGG Network} We run the experiments on one GPU, with a batch size of 128. Apart from our 'ONI', all other methods (`plain', `WN', `OrthInit' and `OrthReg') suffer from difficulty in training with a large learning rate of 0.1. We thus run the experiments with initial learning rates of $\{0.01, 0.05\}$ for these, and report the best result.

\vspace{-0.1in}
\paragraph{Residual Network}  We run the experiments on one GPU for the 18- and 50-layer residual network, and two GPUs for the 101-layer residual network. We use a batch size of 256. Considering that `ONI' can improve the optimization efficiency, as shown in the ablation study on ResNet-18, we  run the 50- and 101-layer residual network with a weight decay of $\{0.0001, 0.0002\}$ and report the best result from these two configurations for each method, for a more fair comparison.

\subsection{Ablation Study on Iteration Number}
\label{sup:sec:exp:ablation}
\begin{figure}[t]
	\centering
	\hspace{-0.2in}	\subfigure[Train Error]{
		\begin{minipage}[c]{.46\linewidth}
			\centering
			\includegraphics[width=4.2cm]{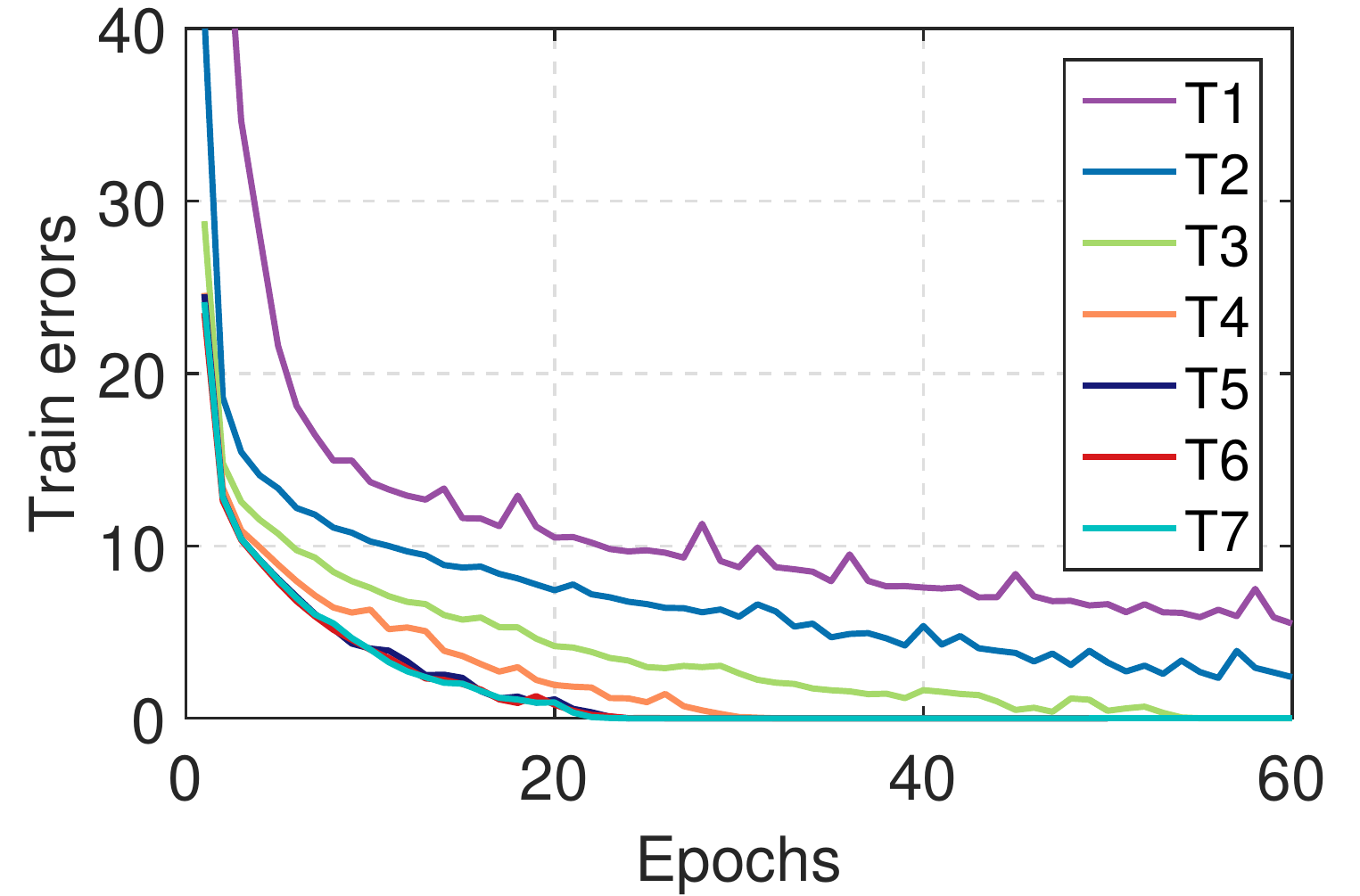}
		\end{minipage}
	}
	\hspace{0.1in}	\subfigure[Test Error]{
		\begin{minipage}[c]{.46\linewidth}
			\centering
			\includegraphics[width=4.2cm]{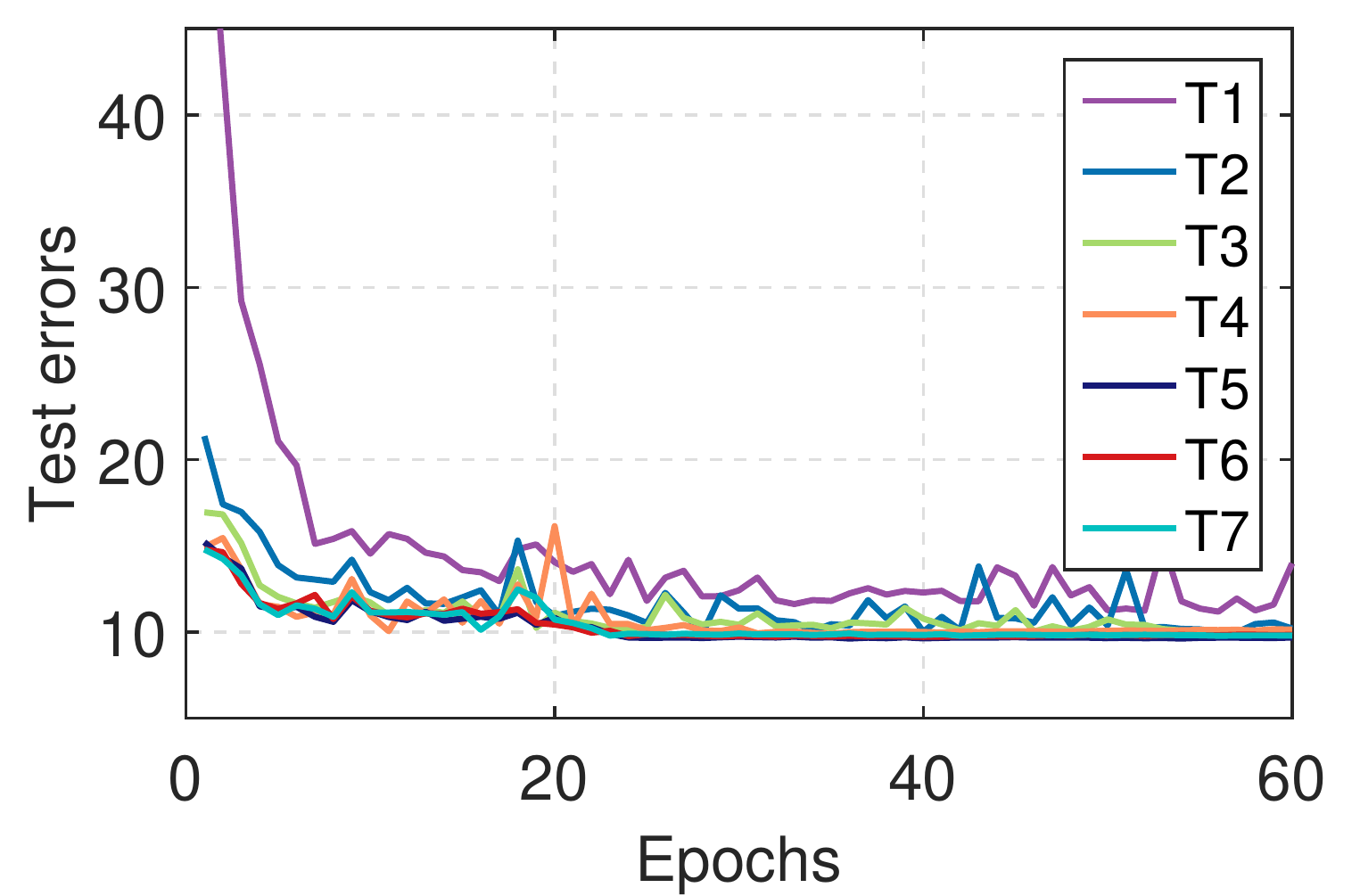}
		\end{minipage}
	}\\
	\subfigure[Distribution of Singular Values]{
		\begin{minipage}[c]{.54\linewidth}
			\centering
			\includegraphics[width=5.0cm]{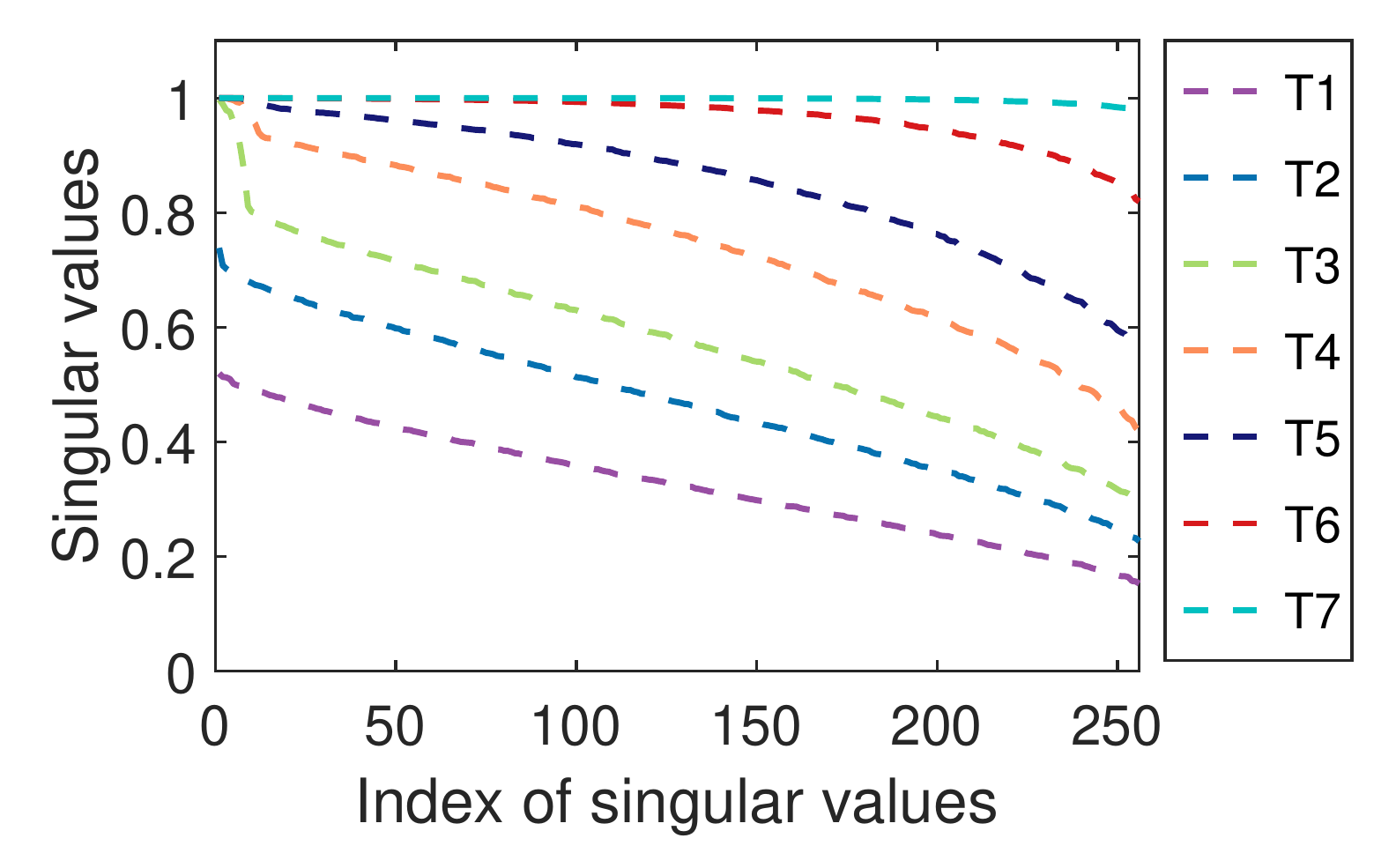}
		\end{minipage}
	}
	\caption{Effects of the iteration number $T$ in proposed `ONI'. We evaluate the training errors on a 10-layer MLP. We show (a) the training errors; (b)  the test errors and (c) the  distribution of the singular values of the orthogonalized weight matrix $\mathbf{W}$ from the first layer, at the 200th iteration.} 
	\label{fig:MLP-iteration}
	\vspace{-0.16in}
\end{figure}

We provide the details of the training performance for ONI on Fashion MNIST in Figure~\ref{fig:MLP-iteration}. We vary $T$, for a range of 1 to 7, and show the training (Figure~\ref{fig:MLP-iteration} (a)) and test  (Figure~\ref{fig:MLP-iteration} (b)) errors with respect to the training epochs. We also provide the   distribution of the singular values of the orthogonalized weight matrix $\mathbf{W}$, using our ONI with different iteration numbers $T$.  Figure \ref{fig:MLP-iteration} (c) shows the results from the first layer, at the 200th iteration.   We also obtain  similar observations for other layers.

%
%
%

\section{Details on Training GANs}
\label{sup:sec:GAN}
For completeness, we provide descriptions of the main concepts used in the paper, as follows. 
\vspace{-0.1in}
\paragraph{Inception score (IS)}
Inception score (IS)  (the higher the better) was introduced by Salimans \etal \cite{2016_NIPS_Salimans}:
\begin{small}
	\begin{eqnarray}
	\label{eqn:IS}
	I_{\mathbf{D}}= exp(E_{\mathbf{D}}[KL (p(y| \mathbf{x}) \rVert p(y)) ]),
	\end{eqnarray}
\end{small}
where $KL(\cdot \rVert \cdot)$ denotes the Kullback-Leibler Divergence, $p(y)$ is approximated by 
$\frac{1}{N}\sum_{i=1}^{N} p(y| \mathbf{x}_i)$ and $p(y|\mathbf{x}_i)$ is the trained Inception model \cite{2014_CoRR_Szegedy}. Salimans \etal \cite{2016_NIPS_Salimans} reported that this score is highly correlated with subjective human judgment of image quality. Following \cite{2016_NIPS_Salimans} and \cite{2018_ICLR_Miyato}, we calculate the score for 5000 randomly generated  examples from each trained generator to evaluate IS. We repeat 10 times and report the average and the standard deviation of IS.

\vspace{-0.1in}
\paragraph{Fr\'{e}chet Inception Distance (FID) \cite{2017_NIPS_Heusel}}
Fr\'{e}chet inception distance (FID) \cite{2017_NIPS_Heusel}  (the lower the better) is another measure for the quality of the generated examples that uses second-order information from the final layer of the inception model applied to the examples. The Fr\'{e}chet distance itself is a 2-Wasserstein distance between two distributions $p_1$ and $p_2$, assuming they are both multivariate Gaussian distributions:
\begin{small}
	\begin{eqnarray}
	\label{eqn:FID}
	F(p_1, p_2) = \|\mathbf{\mu}_{p_1} - \mathbf{\mu}_{p_2}\|_2^2
	+ tr(C_{p_1} + C_{p_2} -2 (C_{p_1}C_{p_2})^{\frac{1}{2}}),
	\end{eqnarray}
\end{small}
\hspace{-0.05in}where $\{\mathbf{\mu}_{p_1}, C_{p_1} \}$, $\{\mathbf{\mu}_{p_2}, C_{p_2} \}$ are the mean and covariance of samples from generated $p_1$ and $p_2$, respectively, and $tr(\cdot)$ indicates the trace operation. We calculate the FID between the  10K test examples (true distribution) and the 5K randomly generated   samples (generated distribution). 
\vspace{-0.1in}
\paragraph{GAN with Non-saturating Loss}
The standard non-saturating function for the adversarial loss is:
\begin{small}
	\begin{eqnarray}
	\label{eqn:gan_bce}
	\mathcal{L}(G, D)=\mathbb{E}_{\mathbf{x} \sim q(\mathbf{x})} [log D(\mathbf{x})]
	+ \mathbb{E}_{\mathbf{z} \sim p(\mathbf{z})} [1- log D(G(\mathbf{z}))],
	\end{eqnarray}
\end{small}
\hspace{-0.05in}where $q(\mathbf{x})$ is the distribution of the real data, $\mathbf{z} \in \mathbb{R}^{d_z}$ is a latent variable, $p(\mathbf{z})$ is the standard normal distribution $N(0, I)$, and $G$ is a deterministic generator function. $d_z$ is set to 128 for all experiments. Based on the suggestion in \cite{2014_NIPS_goodfellow,2018_ICLR_Miyato}, we use the alternate cost $-\mathbb{E}_{\mathbf{z} \sim p(\mathbf{z})} [log D(G(\mathbf{z}))]$ to update G, while using the original cost defined in Eqn.~\ref{eqn:gan_bce} for updating D.

\vspace{-0.1in}
\paragraph{GAN with Hinge Loss}
The hinge loss for adversarial learning is: 
\begin{small}
	\begin{align}
	\label{eqn:hinge}
	\mathcal{L}_D(\hat{G}, D)&=\mathbb{E}_{\mathbf{x} \sim q(\mathbf{x})} [max(0, 1 - D(\mathbf{x}))] \nonumber \\
	&~~~~~~~+ \mathbb{E}_{\mathbf{z} \sim p(\mathbf{z})} [max(0, 1 + D(G(\mathbf{z})))]  \\
	\mathcal{L}_G(G, \hat{D})&= - \mathbb{E}_{\mathbf{z} \sim p(\mathbf{z})} \hat{D}(G(\mathbf{z}))
	\end{align}
\end{small}
\hspace{-0.05in}for the discriminator and the generator, respectively. This type of loss has  already been used in \cite{2017_Corr_GGAN,2018_ICLR_Miyato,2019_ICML_Zhang,2019_ICLR_Brock}. 

Our code is implemented in PyTorch \cite{2017_NIPS_pyTorch} and the  trained Inception model is from the official models in PyTorch \cite{2017_NIPS_pyTorch}. The IS and FID for the real training data are $10.20 \pm 0.13$ and $3.07$ respectively. 
Note that  we do not use the learnable scalar in any the GAN experiment, and set $\sigma=1$ in ONI, for more consistent comparisons with SN. 
\begin{table}[t]
	\centering
	\begin{small}
		\begin{tabular}{lcccc}
			\toprule[1pt]
			\textbf{Setting}&  $\alpha$   &  $\beta_1$  &  $\beta_2$ & $n_{dis}$\\ 
			\hline
			A   & 0.0001 &0.5 & 0.9&5\\
			B & 0.0001  & 0.5 & 0.999&1\\
			C   & 0.0002  & 0.5 & 0.999&1\\
			D  & 0.001 & 0.5& 0.9&5\\
			E  &0.001 & 0.5& 0.999&5\\
			F  & 0.001  &  0.9& 0.999&5\\ 
			\toprule[1pt]
		\end{tabular}
	\end{small}
	\caption{Hyper-parameter settings in stability experiments on DCGAN, following \cite{2018_ICLR_Miyato}.}
	\label{tab:Configure-DCGAN}
	\vspace{-0.1in}
\end{table}

\begin{table*}[t]
	\small
	\centering
	\vspace{0.1cm}
	\subtable[Generator]{
		\begin{tabular}{c} 
			\hline 
			\hline
			$z \in \mathbb{R}^{128} \sim \mathcal{N}(0, I)$ \\
			\hline
			$4\times 4$, stride=1 deconv. BN 512 ReLU $\rightarrow$ $4\times 4\times 512$\\
			\hline
			$4\times 4$, stride=2 deconv. BN 256 ReLU\\
			\hline
			$4\times 4$, stride=2 deconv. BN 128 ReLU\\
			\hline
			$4\times 4$, stride=2 deconv. BN 64 ReLU\\
			\hline
			$3\times 3$, stride=1 conv. 3 Tanh\\
			\hline
			\hline
		\end{tabular}
	}
	\hspace{0.4in}	\subtable[Discriminator]{
		\renewcommand\arraystretch{1.1}
		\begin{tabular}{c} 
			\hline 
			\hline
			RGB image $x \in \mathbb{R}^{32 \times 32 \times 3}$ \\
			\hline
			$3\times 3$, stride=1 conv 64 lReLU\\
			$4\times 4$, stride=2 conv 64 lReLU\\
			\hline			
			$3\times 3$, stride=1 conv 128 lReLU\\
			$4\times 4$, stride=2 conv 128 lReLU\\
			\hline		
			$3\times 3$, stride=1 conv 256 lReLU\\
			$4\times 4$, stride=2 conv 256 lReLU\\
			\hline
			$3\times 3$, stride=1 conv 512 lReLU\\
			\hline
			dense $\rightarrow$ 1 \\
			\hline
			\hline
		\end{tabular}
	}	
	\caption{DCGAN architectures for CIFAR10 dataset in our experiments.  `lReLU` indicates the leaky ReLU \cite{2013_ICMLW_Maas}  and its slope is set to 0.1. }
	\label{tab:arc-dcgan}
\end{table*}
\begin{table*}[t]
	\small
	\begin{minipage}[t]{0.9\textwidth}
		\centering
		\vspace{0.1cm}
		\subtable[Generator]{
			\centering
			\renewcommand\arraystretch{1.1}
			\begin{tabular}{c} 
				\hline 
				\hline
				$z \in \mathbb{R}^{128} \sim \mathcal{N}(0, I)$ \\
				\hline
				dense, $4\times 4\times 128$\\
				\hline
				ResBlock up 128\\
				\hline
				ResBlock up 128\\
				\hline
				ResBlock up 128\\
				\hline
				BN, ReLU, $3\times 3$ conv, 3 Tanh\\
				\hline
				\hline
			\end{tabular}
		}
		\hspace{0.4in}	\subtable[Discriminator]{
			\centering
			\renewcommand\arraystretch{1.1}
			\begin{tabular}{c} 
				\hline 
				\hline
				RGB image $x \in \mathbb{R}^{32 \times 32 \times 3}$ \\
				\hline
				ResBlock down 128\\
				\hline
				ResBlock down 128\\
				\hline
				ResBlock 128\\
				\hline
				ResBlock 128\\
				\hline
				ReLU\\
				\hline
				Global sum pooling\\
				\hline
				dense $\rightarrow$ 1 \\
				\hline
				\hline
			\end{tabular}
		}	
		\caption{ResNet architectures for CIFAR10 dataset in our experiments. We use the same ResBlock as the SN paper \cite{2018_ICLR_Miyato}.}
		\label{tab:arc-Resnet}
	\end{minipage}
	\vspace{-0.1in}
\end{table*}

\subsection{Experiments on  DCGAN}
\label{sup:sec:DCGAN}
The DCGAN architecture follows the configuration in \cite{2018_ICLR_Miyato}, and we provide the details in Table~\ref{tab:arc-dcgan} for completeness. 
The spectral normalizaiton (SN) and our ONI are only applied on the discriminator, following the experimental setup in the SN paper \cite{2018_ICLR_Miyato}. 

\begin{figure}[t]
	\centering
	\hspace{-0.2in}	\subfigure[]{
		\begin{minipage}[c]{.46\linewidth}
			\centering
			\includegraphics[width=4.2cm]{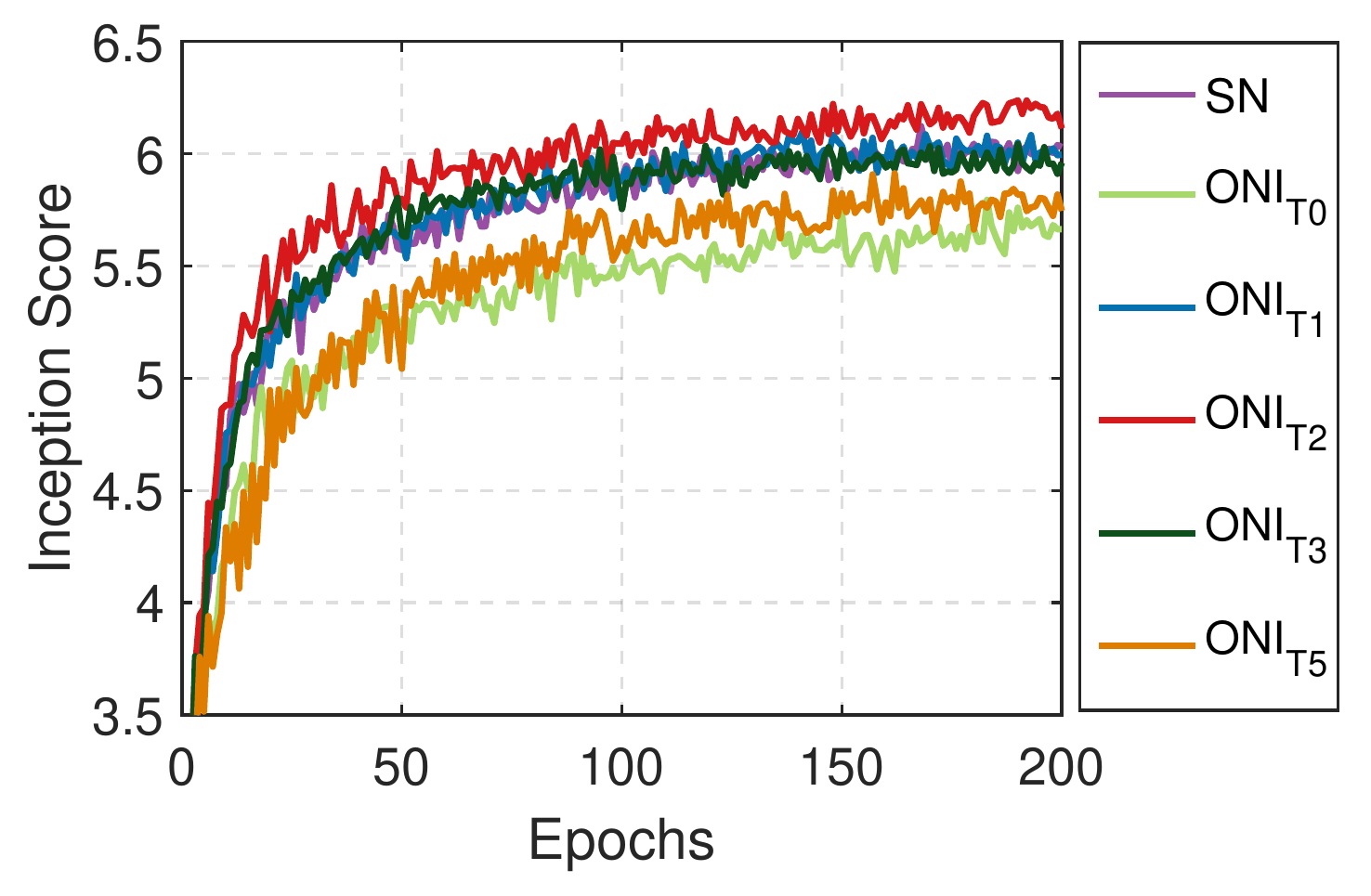}
		\end{minipage}
	}
	\hspace{0.2in}	\subfigure[]{
		\begin{minipage}[c]{.46\linewidth}
			\centering
			\includegraphics[width=4.0cm]{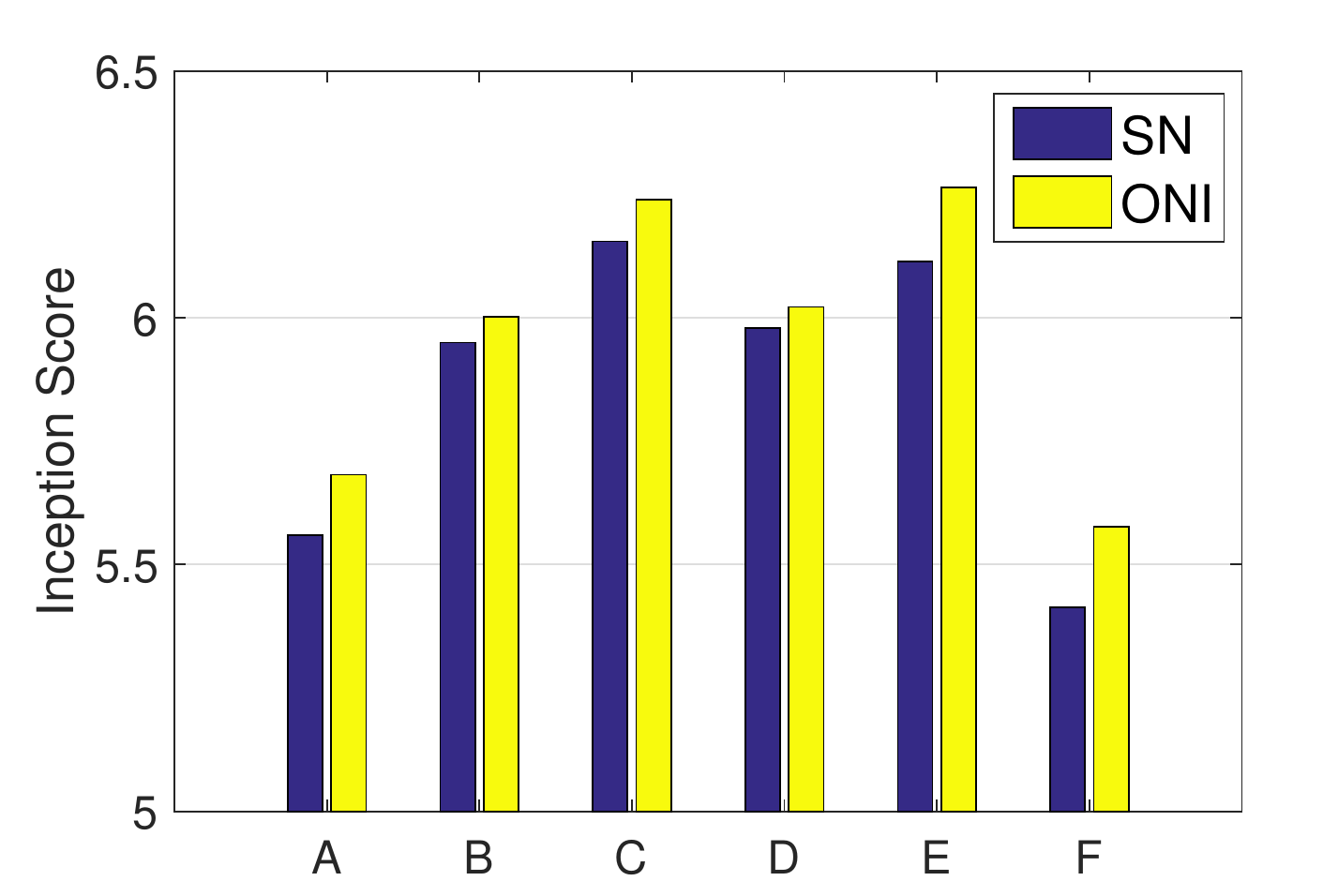}
		\end{minipage}
	}
	\caption{Comparison of SN and ONI on DCGAN. (a) The IS with respect to training epochs. (b)The stability experiments on the six configurations described in \cite{2018_ICLR_Miyato}.} 
	\label{sup:fig:GAN-dcagan}
	\vspace{-0.1in}
\end{figure}

Figure \ref{sup:fig:GAN-dcagan} (a) shows the IS of SN and ONI when varying Newton's iteration number $T$ from 0 to 5. We obtain the same observation as the FID evaluation, shown in Section~\ref{sec:GAN} of the paper.

As discussed in Section~\ref{sec:GAN} of the paper, we conduct experiments to validate the stability of our proposed ONI under different experimental configurations, following \cite{2018_ICLR_Miyato}. 
Table~\ref{tab:Configure-DCGAN} shows the corresponding configurations (denoted by A-F) when  varying the learning rate $\alpha$, first momentum $\beta_1$, second momentum  $\beta_2$, and the number of updates of the discriminator per update of the generator $n_{dis}$. The results evaluated by IS are shown in Figure~\ref{sup:fig:GAN-dcagan} (b). We observe that our ONI is consistently better than SN under the IS evaluation.

\begin{figure}[t]
	\centering
	\vspace{-0.1in}
	\hspace{-0.1in}	\subfigure[]{
		\begin{minipage}[c]{.46\linewidth}
			\centering
			\includegraphics[width=4.3cm]{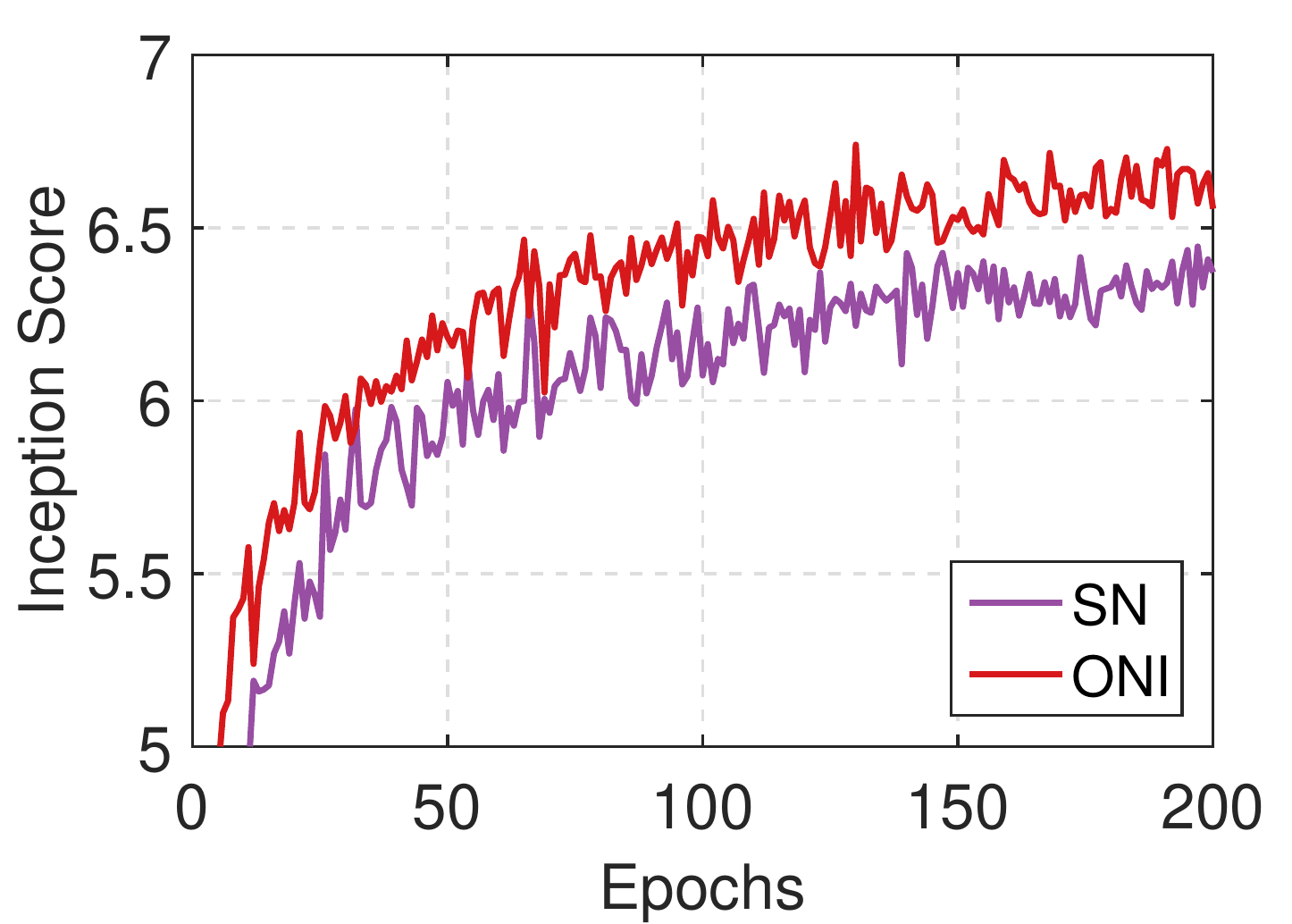}
		\end{minipage}
	}
	\hspace{0.1in}	\subfigure[]{
		\begin{minipage}[c]{.46\linewidth}
			\centering
			\includegraphics[width=4.3cm]{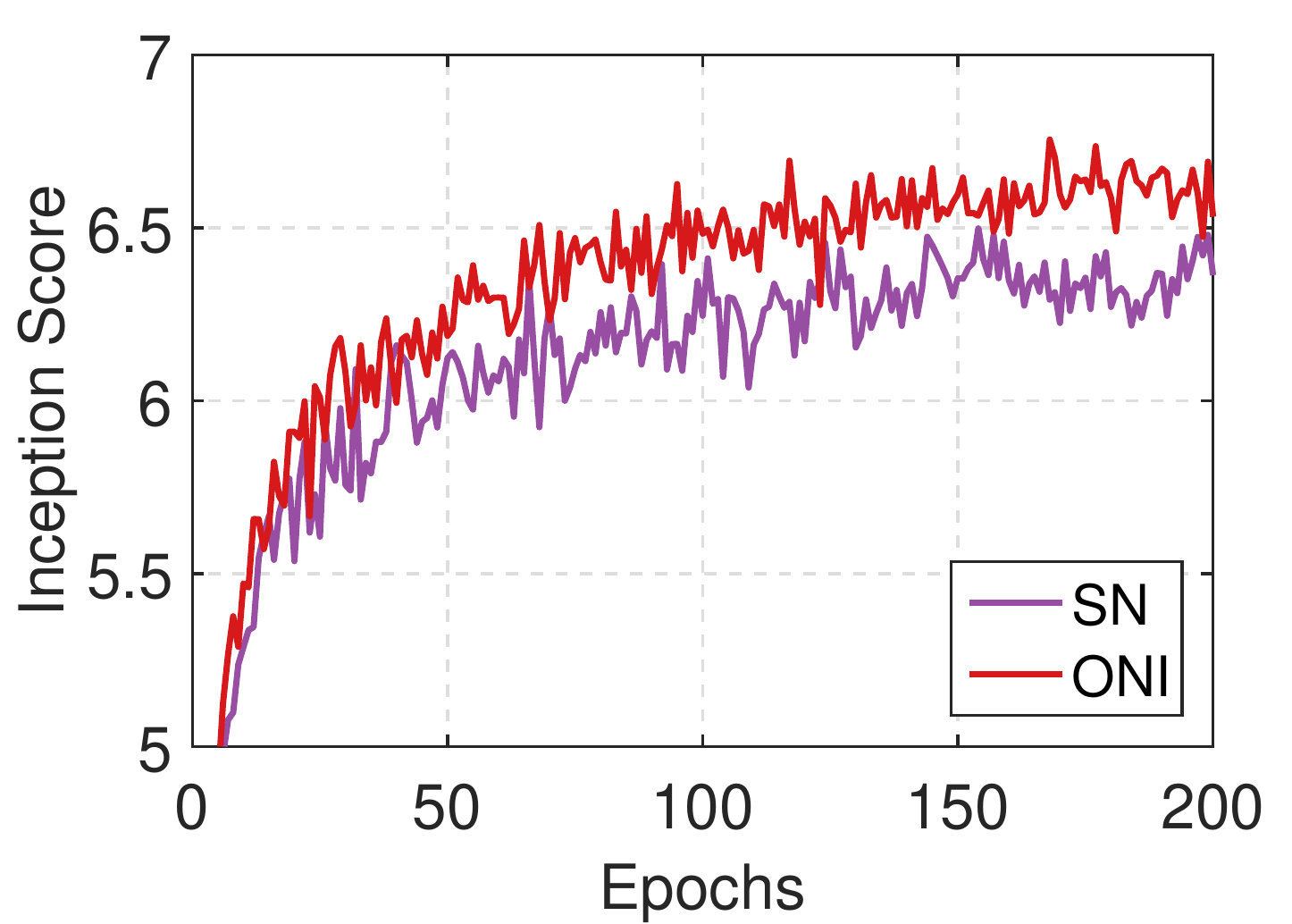}
		\end{minipage}
	}
	\caption{Comparison of SN and ONI on ResNet GAN. We show the  IS with respect to training epochs  using (a) the non-saturating loss and (b) the hinge loss.} 
	\label{sup:fig:GAN-resnet}
	\vspace{-0.1in}
\end{figure}


\subsection{Implementation Details of ResNet-GAN}
The ResNet architecture also follows the configuration in \cite{2018_ICLR_Miyato}, and we provide the details in Table \ref{tab:arc-Resnet} for completeness. 
The SN and our ONI are only applied on the discriminator, following the experimental setup in the SN paper \cite{2018_ICLR_Miyato}.

We provide the results of  SN and ONI in Figure \ref{sup:fig:GAN-resnet}, evaluated by IS.

\subsection{Qualitative Results of GAN}
\begin{figure*}[t]
	\centering
	\subfigure[$T=0$]{
		\begin{minipage}[c]{.32\linewidth}
			\centering
			\includegraphics[width=4.6cm]{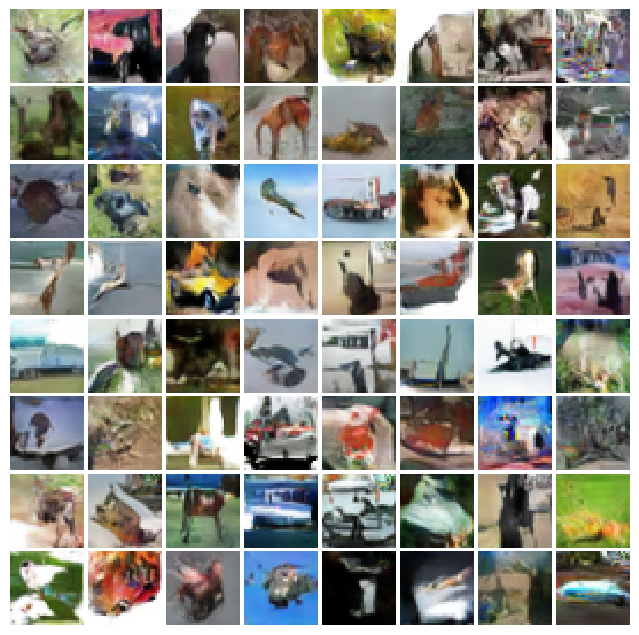}
		\end{minipage}
	}
	\subfigure[$T=1$]{
		\begin{minipage}[c]{.32\linewidth}
			\centering
			\includegraphics[width=4.6cm]{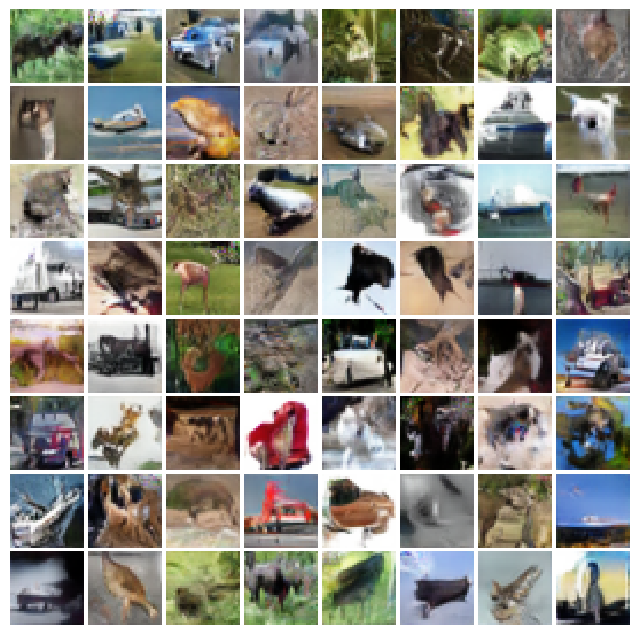}
		\end{minipage}
	}
	\subfigure[$T=2$]{
		\begin{minipage}[c]{.32\linewidth}
			\centering
			\includegraphics[width=4.6cm]{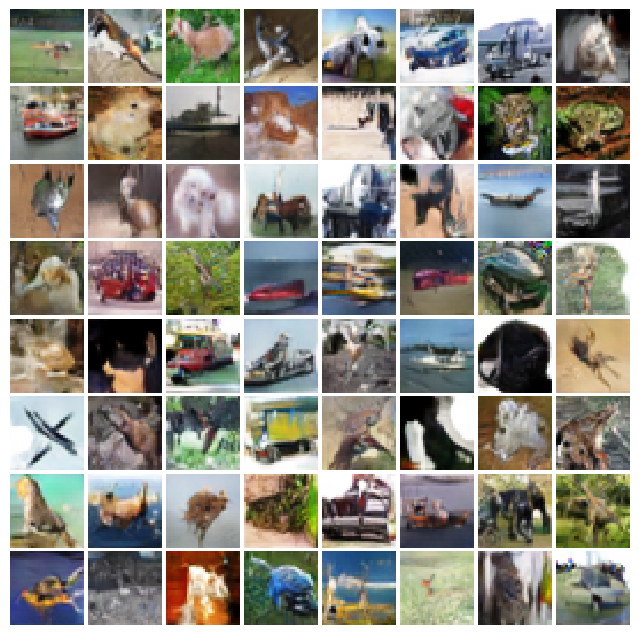}
		\end{minipage}
	}\\
	\subfigure[$T=3$]{
		\begin{minipage}[c]{.32\linewidth}
			\centering
			\includegraphics[width=4.6cm]{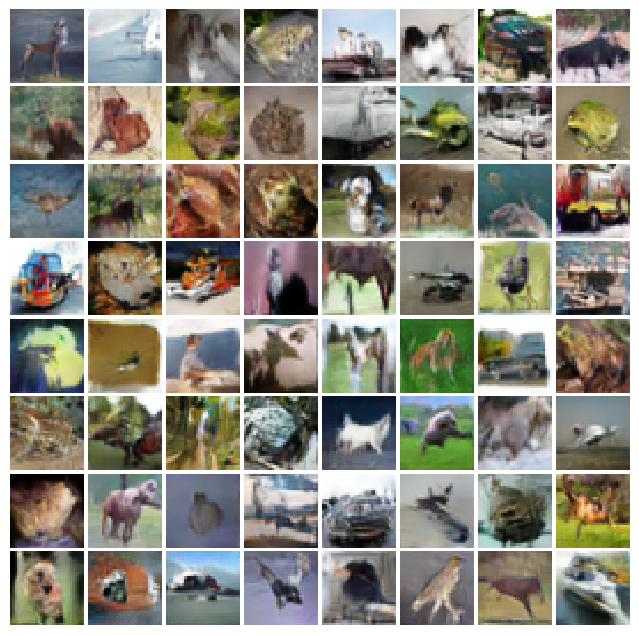}
		\end{minipage}
	}
	\subfigure[$T=4$]{
		\begin{minipage}[c]{.32\linewidth}
			\centering
			\includegraphics[width=4.6cm]{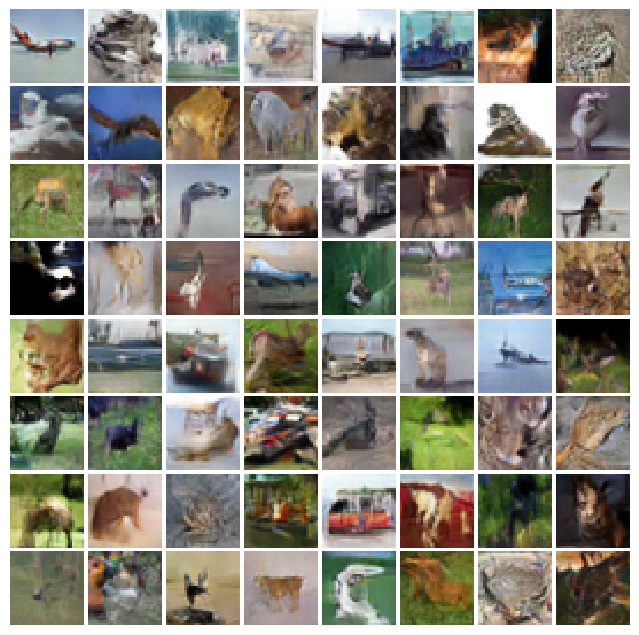}
		\end{minipage}
	}
	\subfigure[$T=5$]{
		\begin{minipage}[c]{.32\linewidth}
			\centering
			\includegraphics[width=4.6cm]{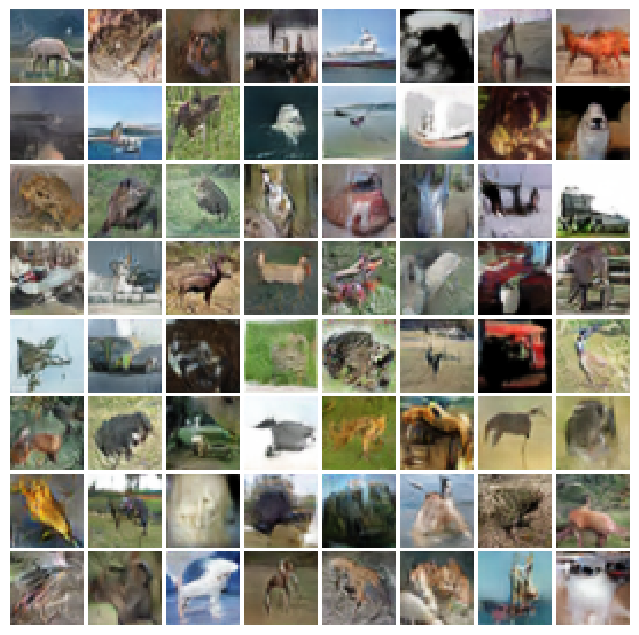}
		\end{minipage}
	}
	\caption{Generated images for CIFAR-10 by our ONI with different iterations,  using DCGAN \cite{2018_ICLR_Miyato}.  } 
	\label{fig:GAN-gImage-Iteration}
	\vspace{-0.16in}
\end{figure*}

\begin{figure*}[t]
	\centering
	\subfigure[SN-A]{
		\begin{minipage}[c]{.32\linewidth}
			\centering
			\includegraphics[width=4.6cm]{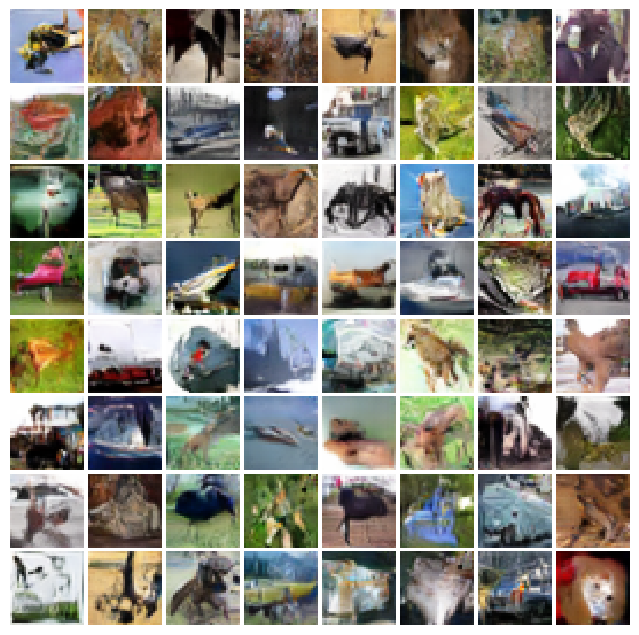}
		\end{minipage}
	}
	\subfigure[SN-B]{
		\begin{minipage}[c]{.32\linewidth}
			\centering
			\includegraphics[width=4.6cm]{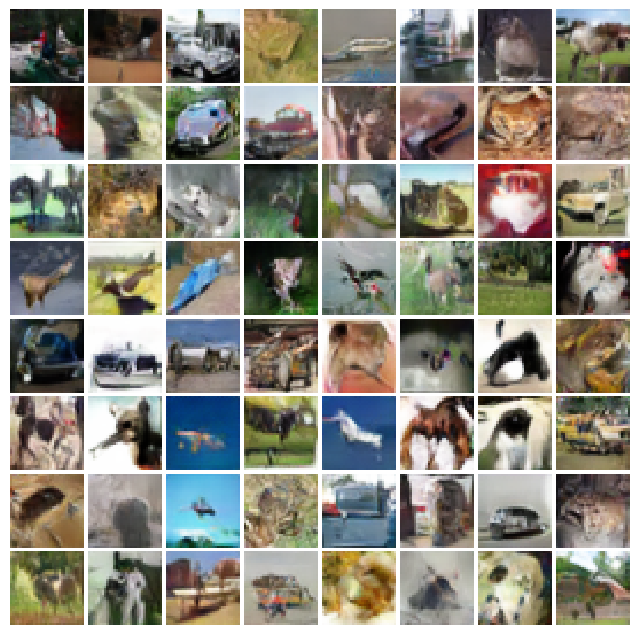}
		\end{minipage}
	}
	\subfigure[SN-C]{
		\begin{minipage}[c]{.32\linewidth}
			\centering
			\includegraphics[width=4.6cm]{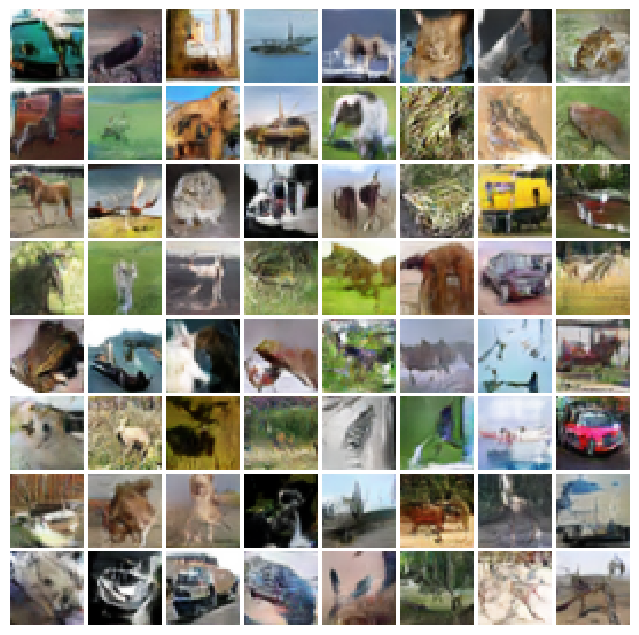}
		\end{minipage}
	}\\
	\subfigure[ONI-A]{
		\begin{minipage}[c]{.32\linewidth}
			\centering
			\includegraphics[width=4.6cm]{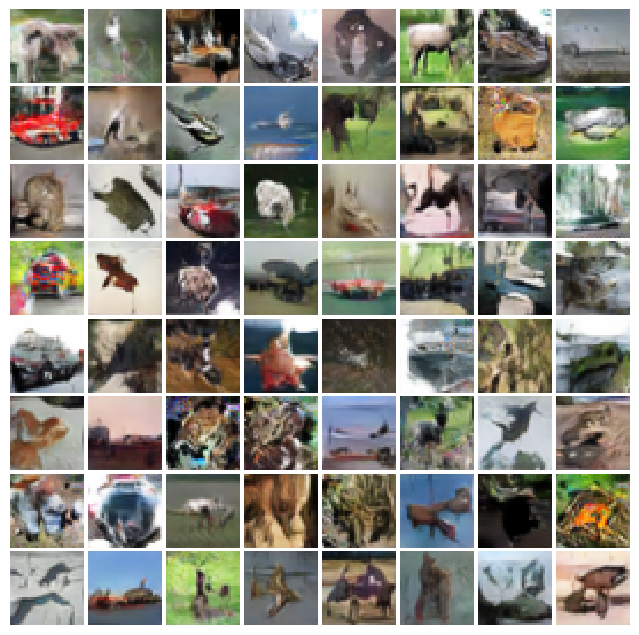}
		\end{minipage}
	}
	\subfigure[ONI-B]{
		\begin{minipage}[c]{.32\linewidth}
			\centering
			\includegraphics[width=4.6cm]{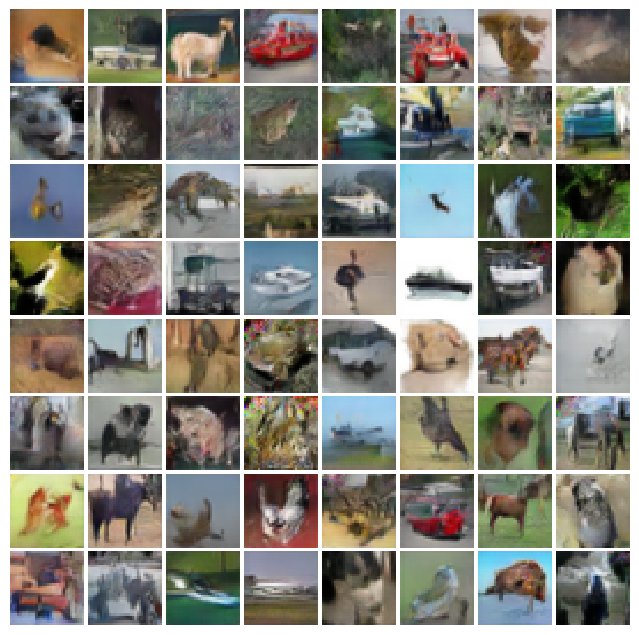}
		\end{minipage}
	}
	\subfigure[ONI-C]{
		\begin{minipage}[c]{.32\linewidth}
			\centering
			\includegraphics[width=4.6cm]{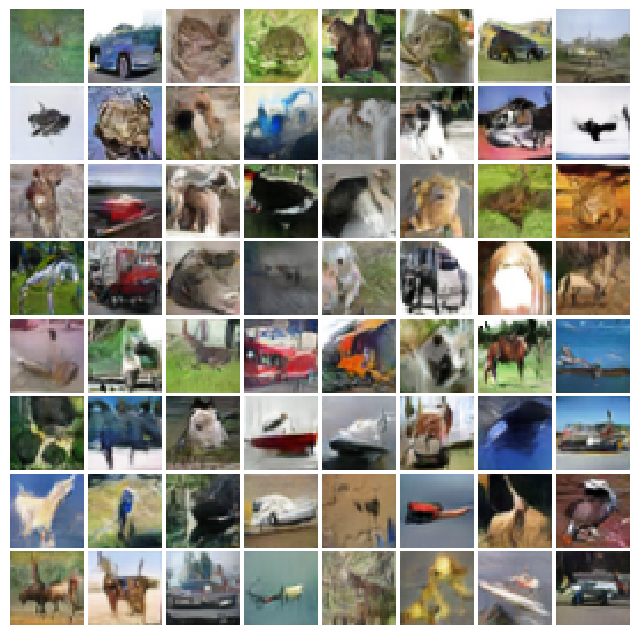}
		\end{minipage}
	}
	\caption{Generated images for CIFAR-10 by SN and ONI,  using DCGAN \cite{2018_ICLR_Miyato}. We show the results of SN and ONI, with configuration A, B and C.} 
	\label{fig:GAN-gImage-DCGAN}
	\vspace{-0.16in}
\end{figure*}

\begin{figure*}[t]
	\centering
	\subfigure[SN with non-satruating loss]{
		\begin{minipage}[c]{.46\linewidth}
			\centering
			\includegraphics[width=6.4cm]{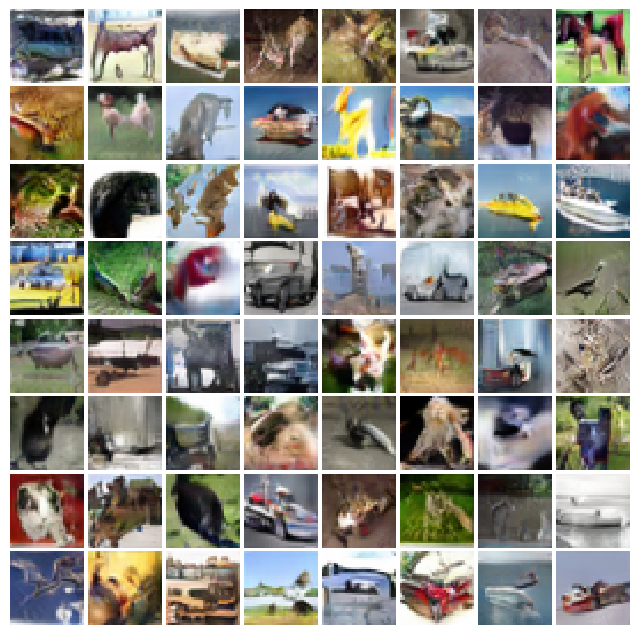}
		\end{minipage}
	}
	\subfigure[SN with hinge loss]{
		\begin{minipage}[c]{.46\linewidth}
			\centering
			\includegraphics[width=6.4cm]{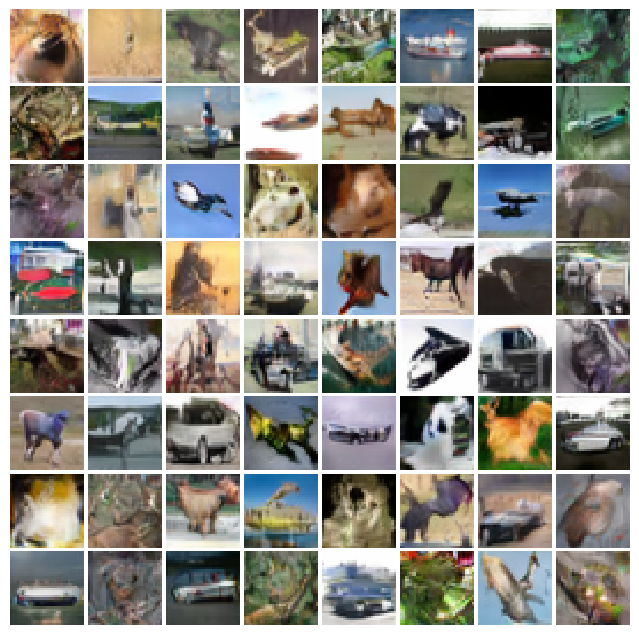}
		\end{minipage}
	}
	\\
	\subfigure[ONI with non-satruating loss]{
		\begin{minipage}[c]{.46\linewidth}
			\centering
			\includegraphics[width=6.4cm]{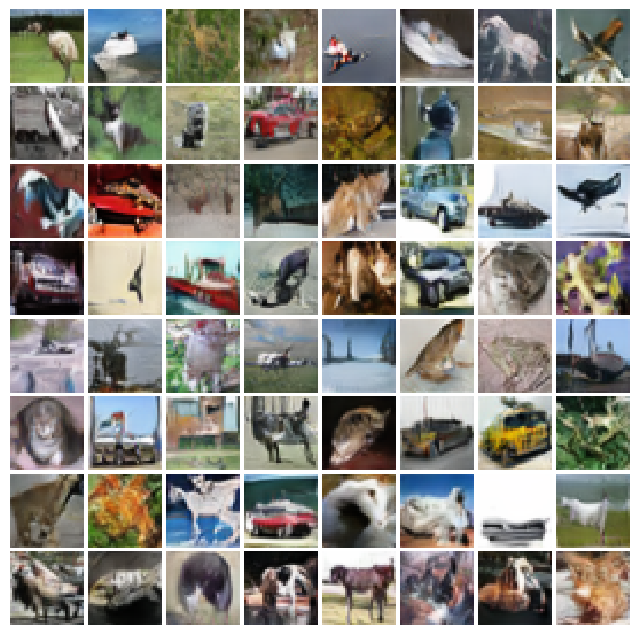}
		\end{minipage}
	}
	\subfigure[ONI with hinge loss]{
		\begin{minipage}[c]{.46\linewidth}
			\centering
			\includegraphics[width=6.4cm]{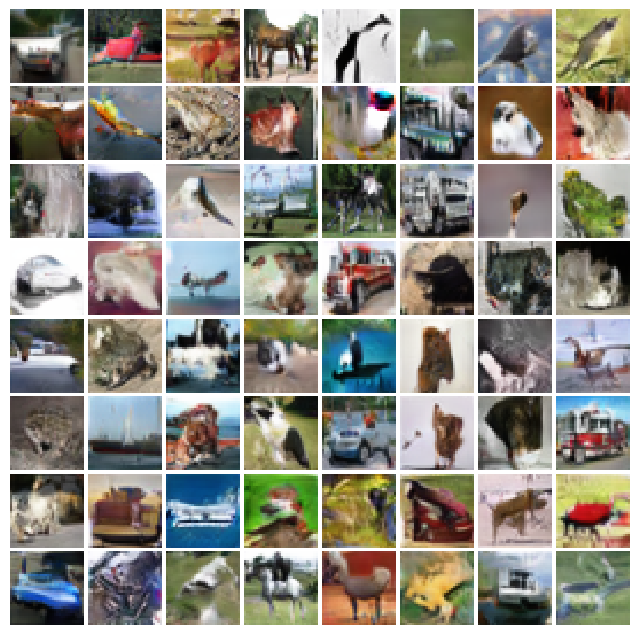}
		\end{minipage}
	}
	\caption{ Generated images for CIFAR-10 by SN and ONI,  using ResNet \cite{2018_ICLR_Miyato}. We show the results of SN and ONI, with the non-satruating and hinge loss.} 
	\label{fig:GAN-gImage-Resnet}
	\vspace{-0.16in}
\end{figure*}

We provide the generated images in Figure~\ref{fig:GAN-gImage-Iteration}, \ref{fig:GAN-gImage-DCGAN} and ~\ref{fig:GAN-gImage-Resnet}. Note that we don't hand-pick the images, and show all the results at the end of the training.

\end{document}